\documentclass[runningheads]{llncs}

\usepackage{eccv}

\usepackage{eccvabbrv}
\usepackage{wrapfig}

\usepackage{graphicx}
\usepackage{booktabs}

\usepackage[accsupp]{axessibility}  

\usepackage{indentfirst}
\usepackage{algorithm}

\usepackage[breaklinks,colorlinks,citecolor=cvprblue]{hyperref}
\usepackage{subfloat}
\usepackage{xcolor}
\usepackage{graphicx}
\usepackage{amsmath}
\usepackage{amssymb}
\usepackage{booktabs}
\usepackage{enumitem}
\usepackage{times}
\usepackage{epsfig}
\usepackage{enumitem}
\usepackage{subcaption}
\usepackage{algpseudocode}

\newtheorem{observation}[theorem]{Observation}

\usepackage{multirow}

\usepackage{multirow}

\usepackage{subfloat}
\usepackage{xcolor}
\usepackage{graphicx}
\usepackage{amsmath}
\usepackage{amssymb}
\usepackage{booktabs}
\usepackage{enumitem}
\usepackage{times}
\usepackage{epsfig}
\usepackage{enumitem}

\usepackage{orcidlink}

\begin{document}

\title{MaxFusion: Plug\&Play  Multi-Modal Generation in  Text-to-Image Diffusion Models} 

\titlerunning{MaxFusion}

\author{Nithin Gopalakrishnan Nair\inst{1}\orcidlink{0000-1111-2222-3333} \and
Jeya Maria Jose Valanarasu\inst{2} \orcidlink{0000-0001-6424-7529} \and
Vishal M Patel\inst{1}\orcidlink{0000-0002-5239-692X}}

\authorrunning{Nair et al.}

\institute{Johns Hopkins University$^{1}$, Stanford University$^{2}$ \\
\email{\{ngopala2, vpatel36\}@jhu.edu, jmjose@stanford.edu}\\
\url{https://nithin-gk.github.io/maxfusion.github.io/}}

\definecolor{cvprblue}{rgb}{0.21,0.49,0.74}

\maketitle

\begin{center}
\centering
\setlength{\tabcolsep}{0.5pt}
\captionsetup{type=figure}
{\footnotesize
\renewcommand{\arraystretch}{0.5}

\begin{tabular}{c c c c c c c c c c c c}

\tabularnewline
\begin{minipage}{0.088\linewidth}
\centering
\vspace{-1.92\linewidth}
\includegraphics[width=\linewidth]{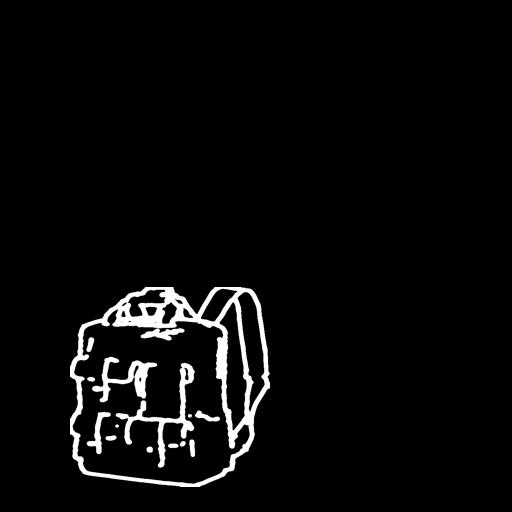}\
\vfill
\includegraphics[width=\linewidth]{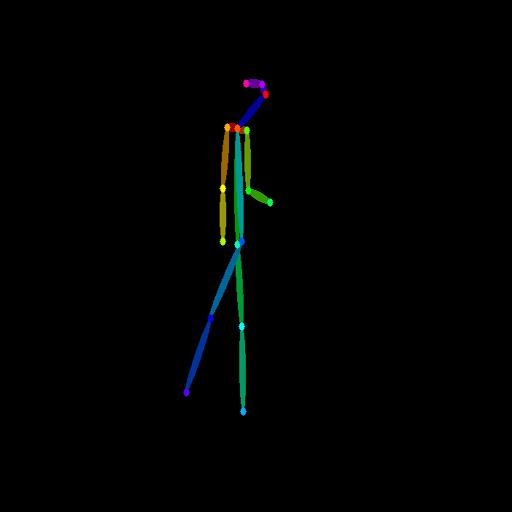}
\end{minipage}
\includegraphics[width=0.18\linewidth]{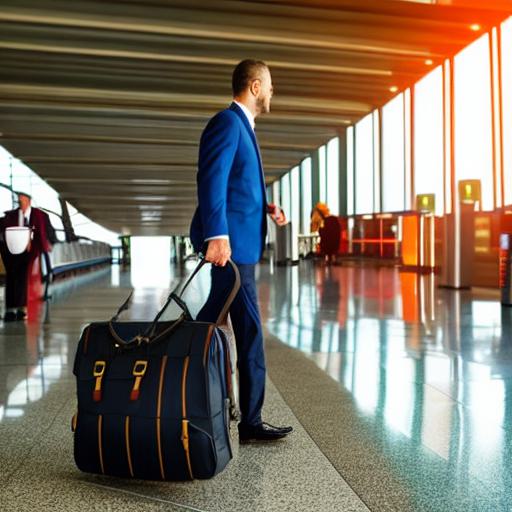}
\includegraphics[width=0.18\linewidth]{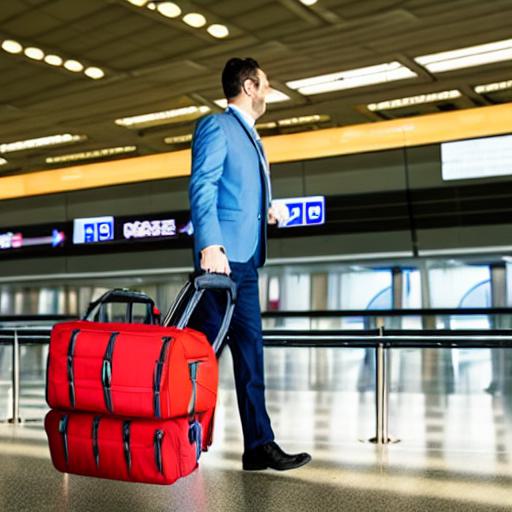}

\hspace{2mm}
\begin{minipage}{0.088\linewidth}
\centering
\vspace{-1.92\linewidth}
\includegraphics[width=\linewidth]{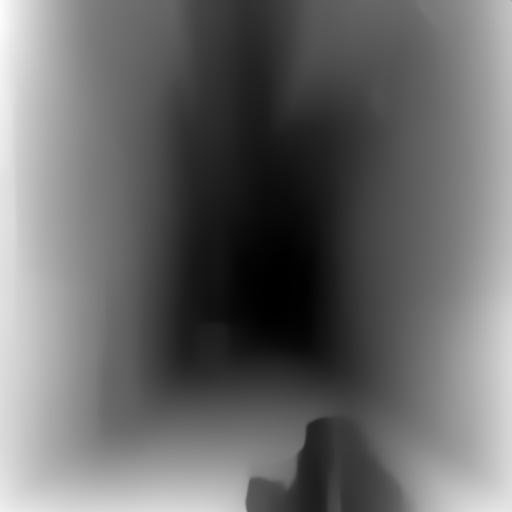}\
\vfill
\includegraphics[width=\linewidth]{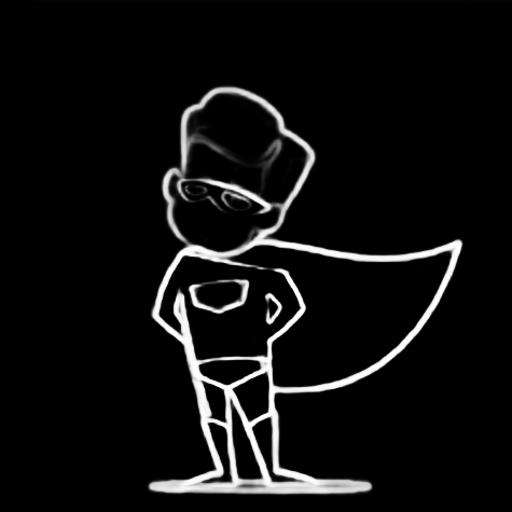}
\end{minipage}
\includegraphics[width=0.18\linewidth]{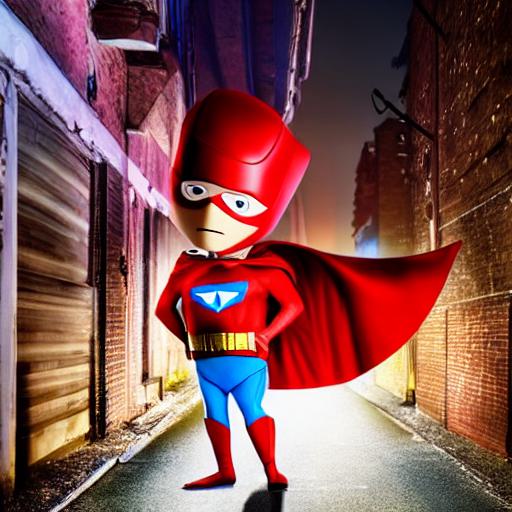}
\includegraphics[width=0.18\linewidth]{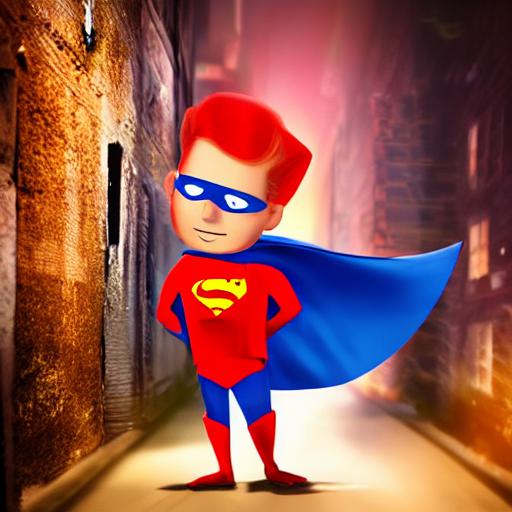}
\tabularnewline
\begin{minipage}{0.088\linewidth}
\centering
\vspace{-1.92\linewidth}
\includegraphics[width=\linewidth]{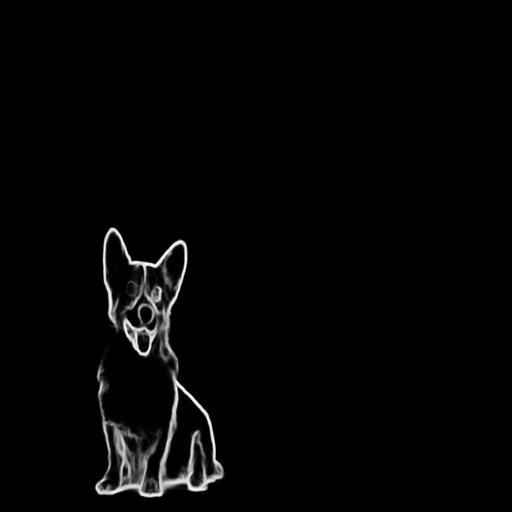}\
\vfill
\includegraphics[width=\linewidth]{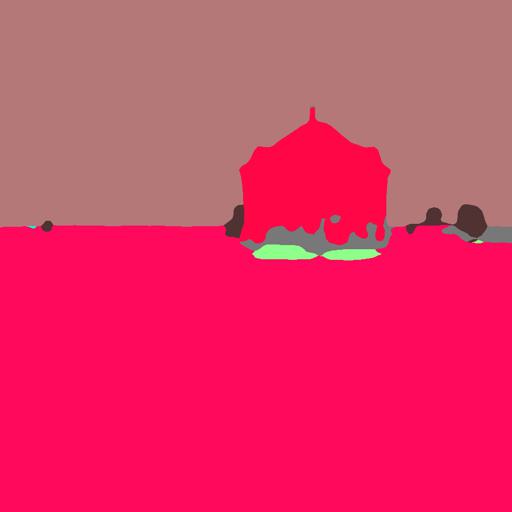}
\end{minipage}
\includegraphics[width=0.18\linewidth]{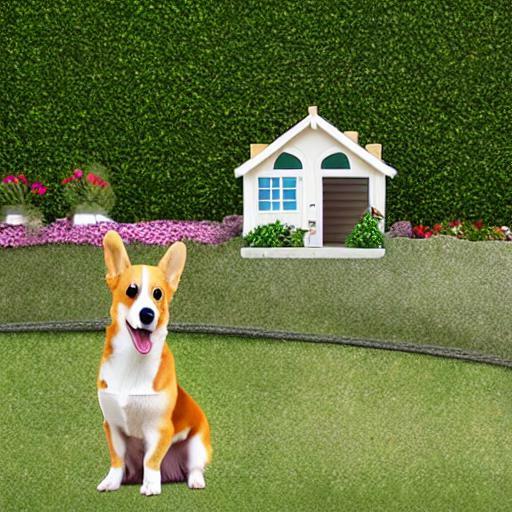}
\includegraphics[width=0.18\linewidth]{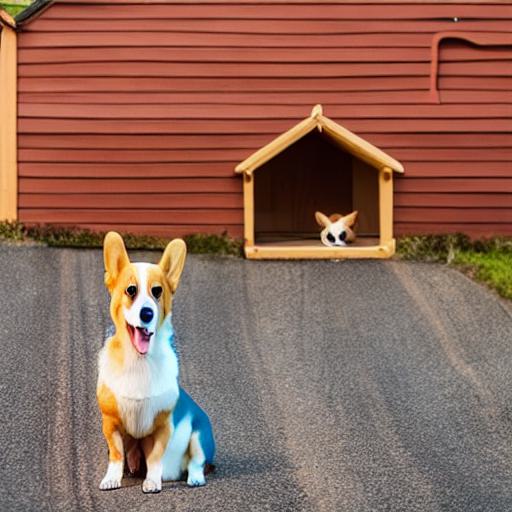}
\hspace{2mm}
\begin{minipage}{0.088\linewidth}
\centering
\vspace{-1.92\linewidth}
\includegraphics[width=\linewidth]{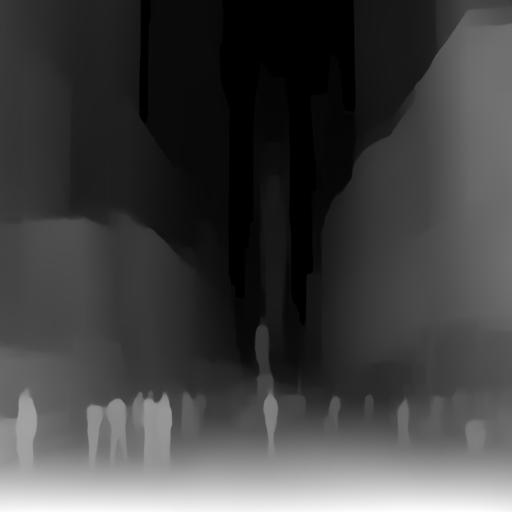}\
\vfill
\includegraphics[width=\linewidth]{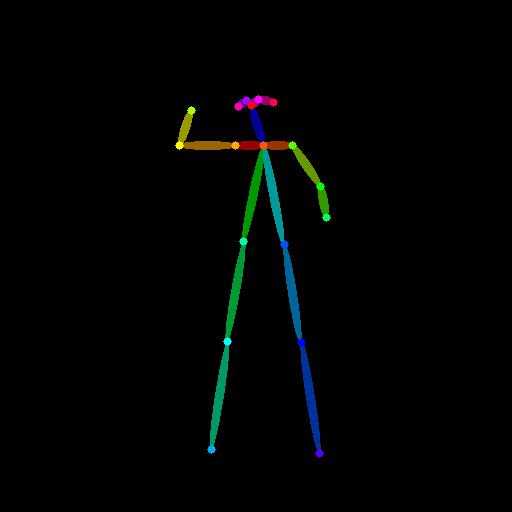}
\end{minipage}
\includegraphics[width=0.18\linewidth]{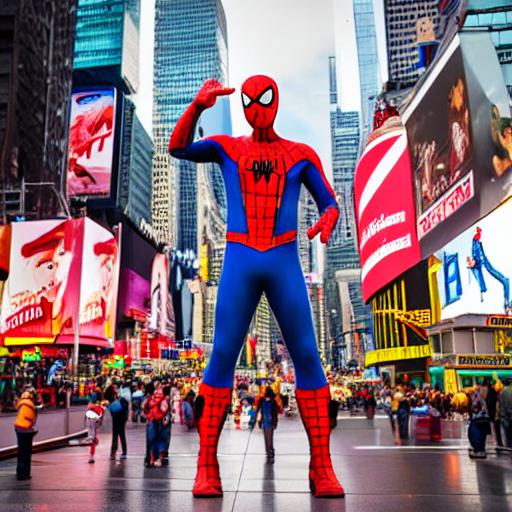}
\includegraphics[width=0.18\linewidth]{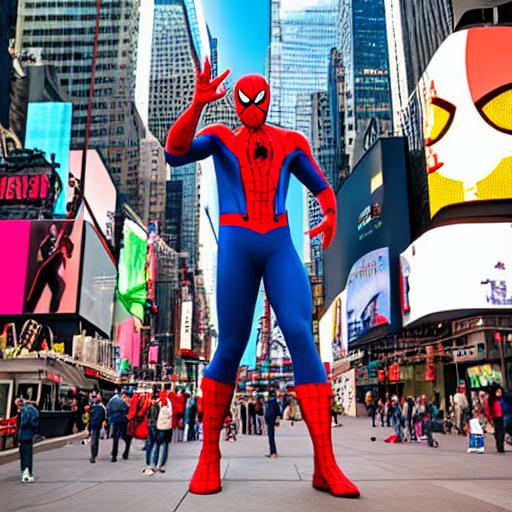}
\tabularnewline
\textit{(a) Contradictory conditions \{Task1 (Object 1) + Task2 (Object 2)}\} $\rightarrow$ Composite Task\
\tabularnewline
\tabularnewline
\begin{minipage}{0.088\linewidth}
\centering
\vspace{-1.92\linewidth}
\includegraphics[width=\linewidth]{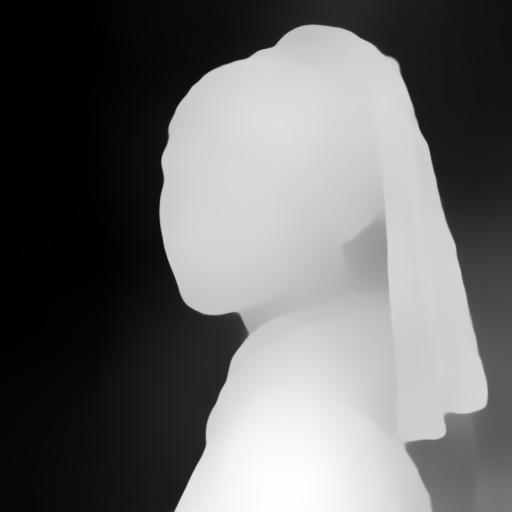}\
\vfill
\includegraphics[width=\linewidth]{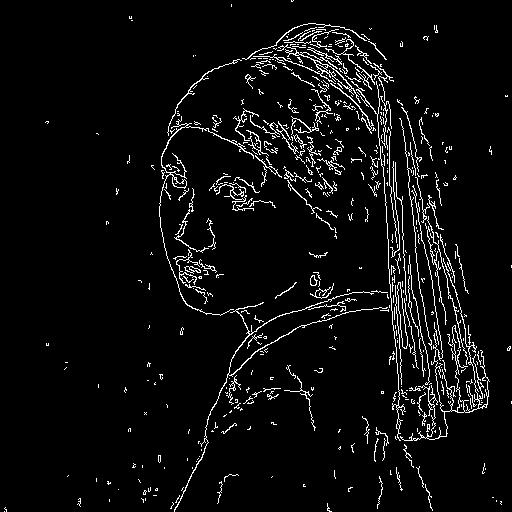}
\end{minipage}
\includegraphics[width=0.18\linewidth]{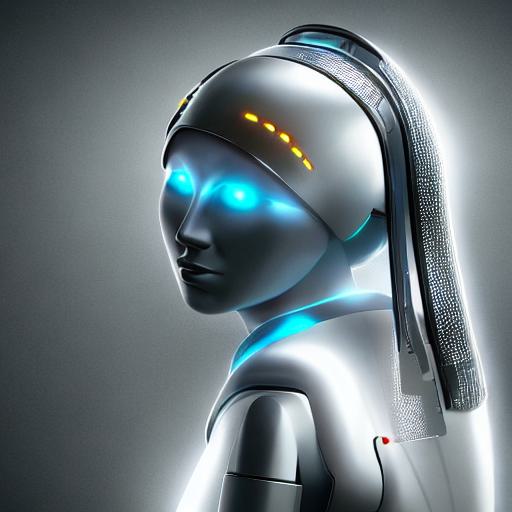}
\includegraphics[width=0.18\linewidth]{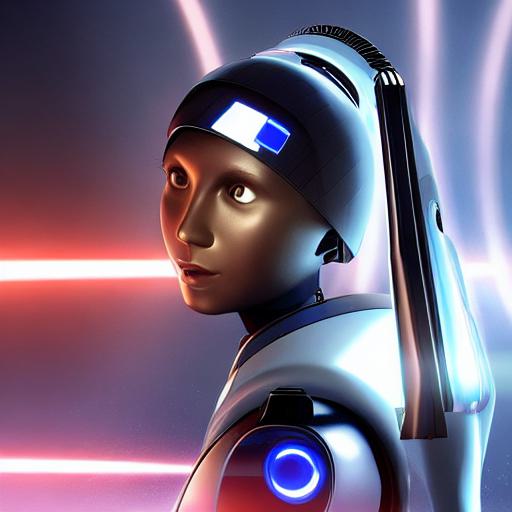}
\begin{minipage}{0.088\linewidth}
\centering
\vspace{-1.92\linewidth}
\includegraphics[width=\linewidth]{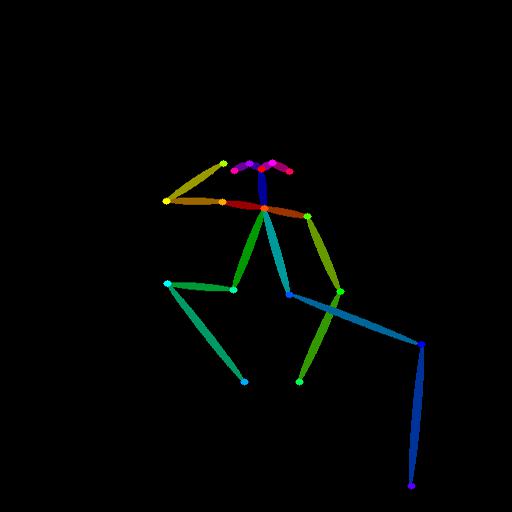}\
\vfill
\includegraphics[width=\linewidth]{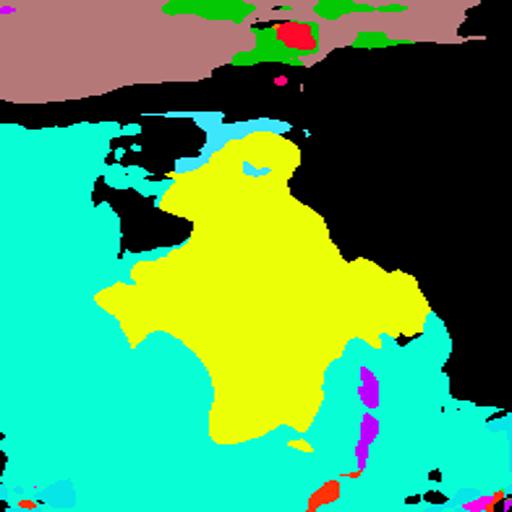}
\end{minipage}
\includegraphics[width=0.18\linewidth]{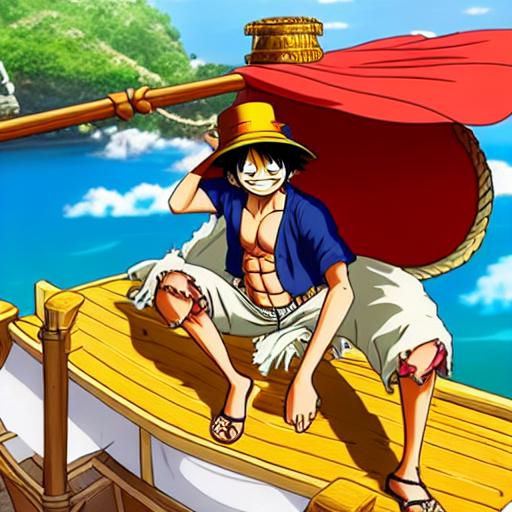}
\includegraphics[width=0.18\linewidth]{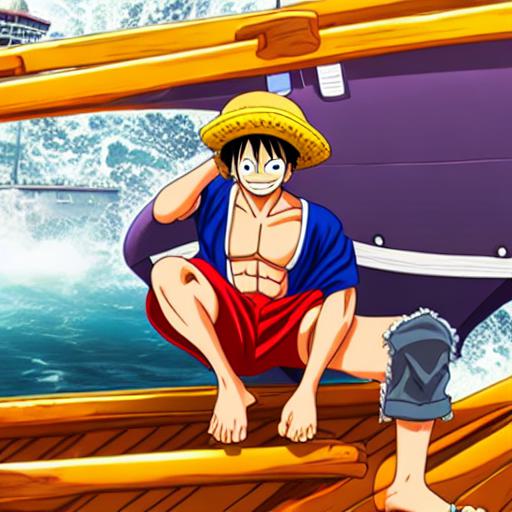}

\tabularnewline
    \textit{(b) Complementary conditions \{Task1 (Object 1) + Task2 (Object 1)\} $\rightarrow$ Composite Task}\\

%

\tabularnewline

\end{tabular}}
\vspace{-0.5\baselineskip}
\hspace{20pt}\captionof{figure}{Applications of MaxFusion: \textbf{(a)} Contradictory condition composition: We show cases where the generated images can easily bring in effects from multiple conditions that otherwise need extremely descriptive conditioning. \textbf{(b)} Complementary condition composition: Multi-modal generation combining details from different modalities but from the same image.
}
\label{fig:introfig}
\vspace{-2mm}
\end{center}%

\begin{abstract}
\vspace{-2mm}
Large diffusion-based Text-to-Image (T2I) models have shown impressive generative powers for text-to-image generation as well as spatially conditioned image generation.
For most applications, we can train the model end-to-end with paired data to obtain photorealistic generation quality. However, to add an additional task, one often needs to retrain the model from scratch using paired data across all modalities to retain good generation performance. In this paper, we tackle this issue and propose a novel strategy to scale a generative model across new tasks with minimal compute. During our experiments, we discovered that the variance maps of intermediate feature maps of diffusion models capture the intensity of conditioning. Utilizing this prior information, we propose MaxFusion, an efficient strategy to scale up text-to-image generation models to accommodate new modality conditions. Specifically, we combine aligned features of multiple models, hence bringing a compositional effect. Our fusion strategy can be integrated into off-the-shelf models to enhance their generative prowess.
\end{abstract}

\section{Introduction}
Denoising Diffusion Probabilistic Models (DDPMs)\cite{ho2020denoising} have revolutionized the content creation industry by delivering photo-realistic quality in image\cite{ho2022imagen,saharia2022photorealistic, rombach2022high,balaji2022ediffi}, video~\cite{ho2022video,ge2023preserve}, and 3D generation tasks~\cite{liu2023zero,poole2022dreamfusion}. These models excel in translating textual descriptions or class labels into precise visual representations, thanks to training that conditions them on text or labels~\cite{nichol2021glide, nichol2021improved}. The remarkable capabilities of recent text-to-image (T2I) diffusion models~\cite{balaji2022ediffi,rombach2022high} have expanded further with the integration of additional control signals, such as instructions~\cite{brooks2023instructpix2pix} or style-based conditioning~\cite{zhang2023inversion,lu2023specialist,ruiz2022dreambooth}. This allows for the creation of images that more closely align with user preferences by enabling more nuanced control through multi-tasking capabilities, enabling multiple conditions to be given to the model simultaneously.

However, training a generative model to accommodate multiple conditions presents significant challenges, primarily due to the need for paired multi-modal data~\cite{zeng2023scenecomposer, lingenfelter2022quantitative}. Moreover, achieving satisfactory performance from multi-modal training often necessitates prolonged training periods. Transfer learning has emerged as a popular solution for multi-modal generation of T2I Diffusion models~\cite{weinbach2022m,xu2023versatile}, allowing large text-to-image models trained on extensive datasets like LAION-2B~\cite{schuhmann2022laion} to undergo fine-tuning for downstream tasks, including image semantics~\cite{zeng2023scenecomposer, zhang2023magicbrush} and instruction-based synthesis~\cite{brooks2023instructpix2pix}, in a comparatively shorter time using smaller datasets (around 10M images)\cite{qin2023unicontrol}. Consequently, with sufficient multi-modal paired data and robust computing resources, fine-tuning a pre-trained text-to-image model to incorporate a broad range of conditions becomes feasible\cite{ruiz2022dreambooth,kumari2023multi}. However, this approach hinges on a fixed set of initial conditions and a strategic selection of training data. A notable drawback is that fine-tuning with smaller datasets can lead to catastrophic forgetting~\cite{kirkpatrick2017overcoming,zeng2023scenecomposer}, where the model loses previously acquired knowledge from the original T2I task.

Recent works such as ControlNet~\cite{zhang2023adding} and T2I Adapter~\cite{mou2023t2i} tackled this problem by introducing task-specific trainable parameters to include additional spatial conditioning for diffusion models. Both these approaches showed impressive image synthesis quality with good coherence to the structural conditioning. However, adapting ControlNet and T2I Adapters to multiple tasks requires either training from scratch or manual parameter tuning across the intermediate outputs of multiple adapters, which accounts for a non-trivial increase in compute time and memory. Hence, synthesizing images/videos satisfying multiple conditions remains a challenge~\cite{ceylan2023pix2video}.
Progressing towards a solution for multi-conditioning, a recent work, Unicontrol~\cite{qin2023unicontrol}, trained a generative model with paired data across diverse conditional modalities~\cite{huang2023collaborative}. Through the multitasking-based training paradigm, Unicontrol achieved better performance for individual spatial conditioning tasks than Controlnet. Unicontrol has been trained with large-scale curated data~\cite{schuhmann2022laion} across various modalities to achieve good visual synthesis quality. However, the challenge associated with such an approach is that in order to add a new conditioning task, the models must undergo retraining, incorporating data not only from the new task but also from all preceding tasks~\cite{qin2023unicontrol,nair2022unite}. We refer to this as the \textit{"scaling issue"} signifying a rise in training complexity when integrating an extra task into a pre-trained model. A training-free solution for multi-conditioning that exists in the literature is compositional generation~\cite{liu2022compositional, feng2022training, du2023reduce}. However, compositional algorithms impose a linear overhead on inference memory. To this end, we try to address the question.
\begin{figure}[tb!]
    \centering
    \setlength{\tabcolsep}{0.5pt}
    {\small
    \renewcommand{\arraystretch}{0.5} 
    \begin{tabular}{c c c c c c  c c c c c }
    \captionsetup{type=figure, font=scriptsize}
  \includegraphics[width=0.105\linewidth]{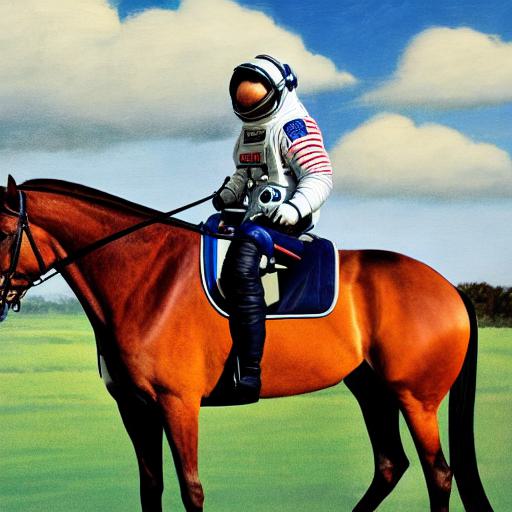}
  \includegraphics[width=0.105\linewidth]{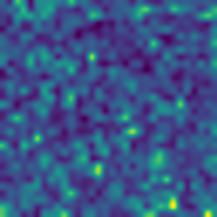}
  \includegraphics[width=0.105\linewidth]{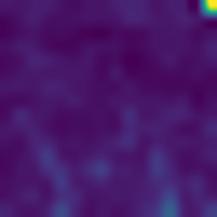}
  \includegraphics[width=0.105\linewidth]{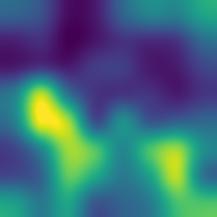}
  \includegraphics[width=0.105\linewidth]{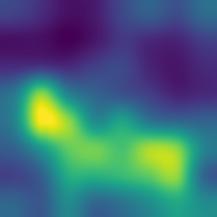}
  \includegraphics[width=0.105\linewidth]{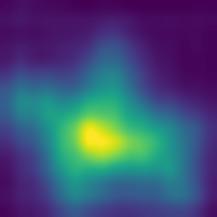}
  \includegraphics[width=0.105\linewidth]{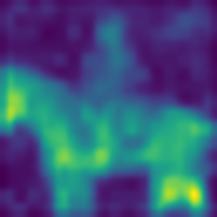}
  \includegraphics[width=0.105\linewidth]{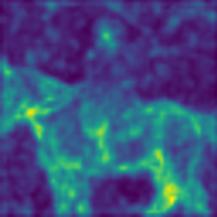}
  \includegraphics[width=0.105\linewidth]{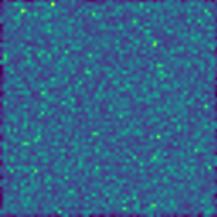}
 \hspace{0.5mm}
\end{tabular}}
\hspace{20pt}\captionof{figure}{Figure illustrating variance maps of intermediate features of encoder and decoder for the text prompt "An astronaut riding a horse" for the $5^{th}$ timestep of diffusion.}\label{fig:astrohorse}
\vspace{-0.5cm}
\label{fig:text_feature}
\end{figure}%
\textbf{\textit{``Can we enable a single diffusion model to scale efficiently across multiple diverse tasks without retraining?"}}.

In this paper, we propose a training-free solution for multi-modal generation. We achieve this by utilizing a novel feature fusion scheme that leverages distinct single task models for compositional generation. Our feature fusion criteria is motivated by an observation regarding intensity of conditioning feature and the variance maps of intermediate layers of a text-to-image diffusion model. An example is shown in \Cref{fig:text_feature}. We establish a criteria derived from feature variance maps across different layers of a diffusion model. We propose MaxFusion - a simple and efficient feature fusion algorithm that allows scaling up T2I models to multiple tasks simultaneously, hence enabling zero-shot multi-modal generation powers to T2I models. \Cref{fig:introfig} presents different applications of our method. In \Cref{fig:introfig}(a), the top row illustrates the case of zero-shot multi-modal generation where two models trained for individual tasks are combined only during inference time to obtain multi-modal generation. These cases illustrate cases where we can combine different modalities of different objects to create a compositional scene. For example, in the first case, we add a pose of a person to a sketch of an object and create a scenario where it can interact. Other examples also show similar multi-modal generations with contradictory condition composition of pose, sketch, depth-maps and semantic maps. Our method can also be utilized for multi-modal conditioning of the same object to add complementary information and this can be seen in \Cref{fig:introfig}(b). For example, we can see that ${Depth, Edges}$ conditions the same image with complementary information.
Our proposed merging mechanism is versatile and can be applied atop any off-the-shelf diffusion models like ControlNet and T2I-Adapter, enabling compositional conditioning tasks and extending their generative capabilities. We perform extensive experiments to show the effectiveness of the proposed method. We present results on:
(i) Multi-modal generation using multiple conditions describing different levels of semantics on the same spatial locations in an image.
(ii) Zero-shot generation where we utilize individual models trained for different tasks and show the combined task without any retraining.

In summary, the following are the main contributions of this work:

\begin{itemize}
\item We tackle the need for training with paired data for multi-task conditioning using diffusion models.

\item We propose a novel variance-based feature merging strategy for diffusion models.

\item Our method allows us to use combined information to influence the output, unlike individual models that are limited to a single condition.
\item Unlike previous solutions, our approach is easily scalable and can be added on top of off-the-shelf models.

\end{itemize}

\section{Related Works}

\subsection{Scaling Diffusion Models to Additional Conditions}

Denoising Diffusion Probabilistic Models (DDPMs)\cite{ho2020denoising} have achieved remarkable results in a variety of tasks, including text-conditioned image generation\cite{rombach2022high,nichol2021glide,saharia2022photorealistic}, video generation~\cite{ho2022imagen,ho2022video}, and multi-modal generation~\cite{rombach2022high,ruiz2022dreambooth}. Methods like ControlNet~\cite{zhang2023adding} and T2I Adapter~\cite{mou2023t2i} have been developed to extend the capabilities of these large, pre-trained diffusion models, enabling them to accommodate additional input conditions. These conditions influence the spatial semantics of the generated images by using paired ${image, text, conditioning}$ triplets, synthetically created for this purpose. Various semantic details, including edge maps, poses, depth, and segmentation maps, serve as conditioning inputs for the model.

To add a new task, a task-specific encoder, carrying the original encoder's weights, is seamlessly integrated into the network. This encoder then connects to convolutional layers initialized at zero, aiming to maintain consistent intermediate latent outputs early in the training, thus ensuring a stable foundation for effective learning. Only the parameters of this additional encoder undergo training to introduce the new condition, avoiding the retraining of pre-existing model weights. This approach is crucial as retraining pre-trained weights could lead to catastrophic forgetting~\cite{kirkpatrick2017overcoming}, a significant risk that ControlNet and T2I-Adapter skillfully navigate to preserve the original model's visual synthesis quality without any performance degradation.

\subsection{Model Merging}

Recent advancements in Large Language Model (LLM) research have introduced the innovative concept of model merging~\cite{yadav2023ties,yang2023adamerging, sung2023empirical, matena2022merging, feng2022training}, a process where weights from multiple pre-trained models for the same task are combined. This technique aims to create a superior model that outperforms the individual contributors. RegMean~\cite{jin2022dataless} pioneered a method for merging models trained on diverse datasets for identical tasks, enhancing overall performance. This method relies on a validation set to derive a closed-form solution to a least squares regression problem, maintaining the merged model's weights in close alignment with that of the original models. Fisher merging~\cite{matena2022merging} assigns varying degrees of importance to different model weights by utilizing the Fisher Information Matrix. Task Arithmetic~\cite{ilharco2022editing} views the discrepancy between pre-trained and fine-tuned weights as task vectors, averaging these weights across tasks to forge a combined model.

While these techniques show promise for token-based input and output scenarios in language models, they encounter limitations when input modalities shift. Direct application of these methods becomes problematic with changes in input modalities, explaining their rare utilization in multi-task conditioning for vision tasks. Specifically, in structural conditioning tasks where features vary significantly across modalities, simplistic averaging mechanisms like task merging prove inadequate, often resulting in the generation of multiple undesired artifacts in multi-task conditioned images. Recent efforts, such as GitRebasin~\cite{ainsworth2022git}, have explored fusing models trained on distinct tasks by aligning their weights. However, subsequent analyses, including work by Repair~\cite{jordan2022repair}, have highlighted the emergence of a 'feature forgetting' issue, where signals from certain tasks weaken in the model's deeper layers. This challenge renders such strategies ineffective for integrating diffusion model weights across modalities for multi-modal generation, as it risks neglecting some conditions.

Therefore, there exists a pressing need for a merging strategy capable of facilitating out-of-domain fusion for diffusion models. A potential solution is the framework proposed by Zipit~\cite{stoica2023zipit}, which maintains two parallel sets of model weights and merges them based on input features. Inspired by this approach, our work seeks to implement a similar strategy for effective model integration.

\section{Proposed Method}
\label{section:method}
\subsection{Unlocking Incremental Task Addition}
ControlNet~\cite{zhang2023adding} performs structural conditioning by using dedicated layers designed to handle different types of inputs, enabling it to process diverse input conditions. Once the inputs are processed by these layers, they are fused with the features from the first layer of the diffusion-UNet model. Therefore, we can intuitively conclude that, during the training process of ControlNet, the weights of the first layer of different control module input layers, like controllers designed for poses and depth, align the modality to the latent space of a natural image. Hence, the signal input to the blocks is converted to a similar domain as that of the input of the diffusion UNet. Moreover, different modality ControlNets, like depth, sketch, and segmentation mask, handle distinct input modalities. Therefore, it is a must that these input processing blocks remain separate. We walk the readers through the strategy of how different modalities can be combined in the rest of the section. Before proposing a technique that can combine features of two different models, we first address two crucial questions: (a) When utilizing two different input modalities and the same backbone architecture, how do we align the feature locations across the models processing these inputs for fusion? (b) Given the presence of two different aligned features, which is the most relevant feature vector required to enforce the condition at that particular feature location?

\subsection{Which Features are Aligned in Conditioned Diffusion Models?}
\begin{figure*}[tb]
\centering
\includegraphics[width=\linewidth]{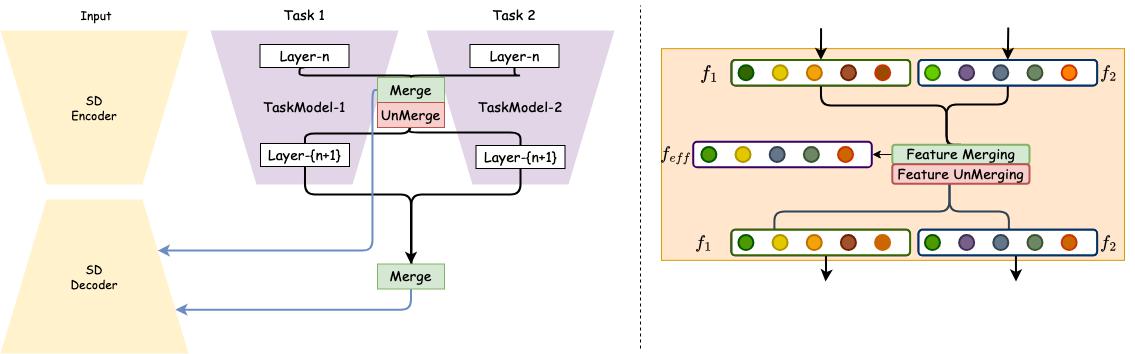}
\caption{An illustration of MaxFusion. During the progression through Stable Diffusion, intermediate features are consolidated at each location, and the resulting merged features are subsequently transmitted to the diffusion module. Please note the colors during the merging and unmerging operation. Similar features are not unmerged and are passed as such to the next network layer. Non-similar intermediate outputs are passed without any change.}
\label{fig:steered}
\vspace{-5mm}
\end{figure*}

To address the problem of feature fusion-based schemes for multi-modal generation, a naive solution would be simply averaging the features of every layer of diffusion. However, when the features are not aligned, such a fusion scheme leads to ineffective conditioning. We point out such an example in \Cref{section:method}.
To overcome this, we leverage the model structure of controlled diffusion models. As seen in \Cref{fig:steered}, intermediate layer outputs get added to specific layers of the T2I model. Hence, we make a proposition:
\begin{observation}
Features from different modules that get added to the same spatial location in Stable Diffusion are aligned.
\end{observation}

\begin{figure*}[tb!]
    \centering
    \setlength{\tabcolsep}{0.5pt}
    {\small
    \renewcommand{\arraystretch}{0.5} 
    \begin{tabular}{c c c c c c c c c c c c c c }
    \captionsetup{type=figure, font=scriptsize}
\includegraphics[width=0.105\linewidth]{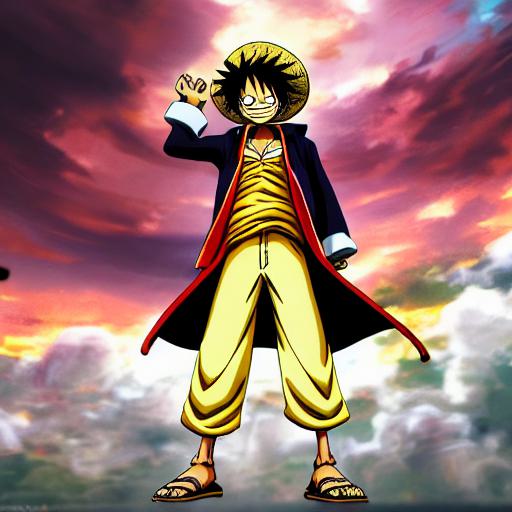}
  \includegraphics[width=0.105\linewidth]{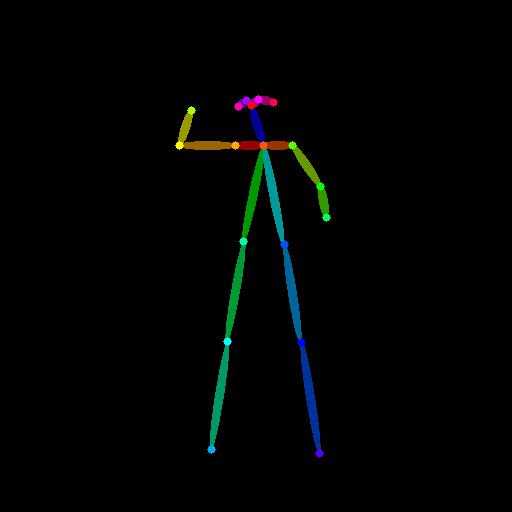}
  \includegraphics[width=0.105\linewidth]{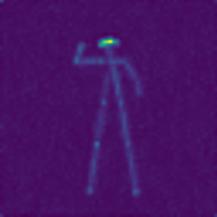}
  \includegraphics[width=0.105\linewidth]{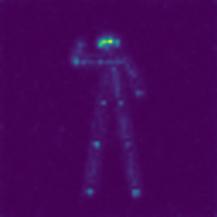}
  \includegraphics[width=0.105\linewidth]{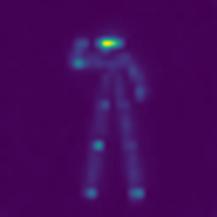}
  \includegraphics[width=0.105\linewidth]{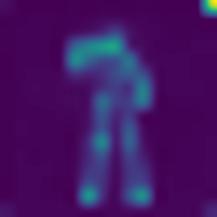}
  \includegraphics[width=0.105\linewidth]{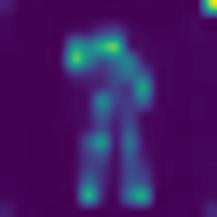}
  \includegraphics[width=0.105\linewidth]{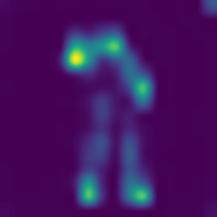}
  \includegraphics[width=0.105\linewidth]{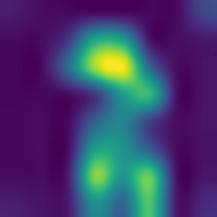}
  \tabularnewline
  \includegraphics[width=0.105\linewidth]{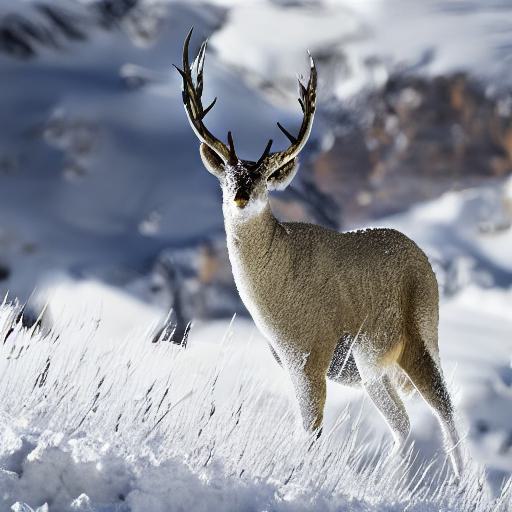}
  \includegraphics[width=0.105\linewidth]{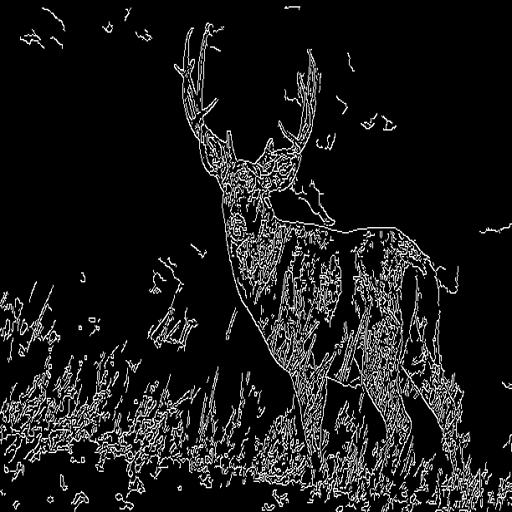}
  \includegraphics[width=0.105\linewidth]{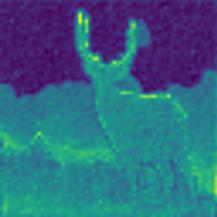}
  \includegraphics[width=0.105\linewidth]{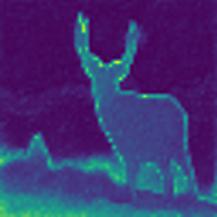}
  \includegraphics[width=0.105\linewidth]{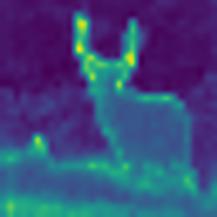}
  \includegraphics[width=0.105\linewidth]{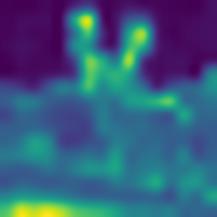}
  \includegraphics[width=0.105\linewidth]{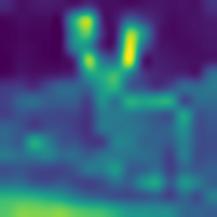}
  \includegraphics[width=0.105\linewidth]{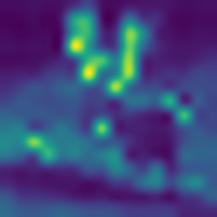}
  \includegraphics[width=0.105\linewidth]{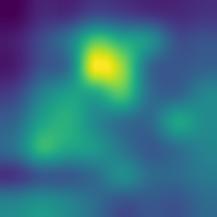}
  \tabularnewline
  \includegraphics[width=0.105\linewidth]{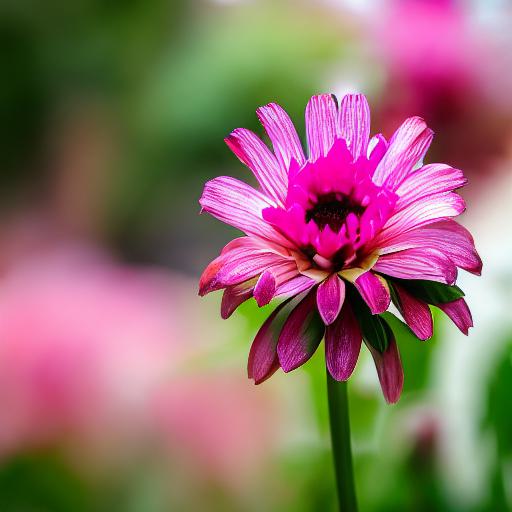}
  \includegraphics[width=0.105\linewidth]{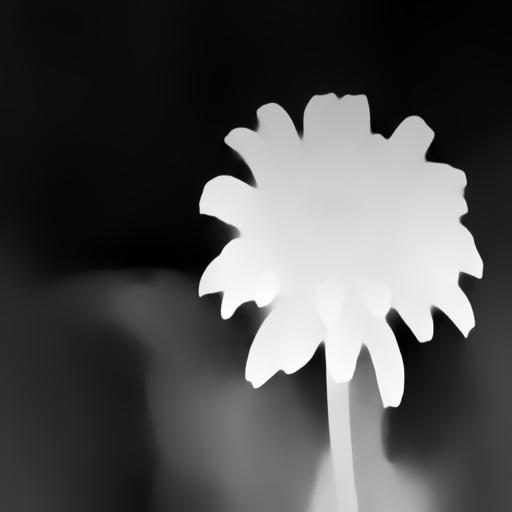}
  \includegraphics[width=0.105\linewidth]{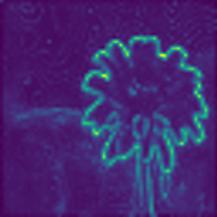}
  \includegraphics[width=0.105\linewidth]{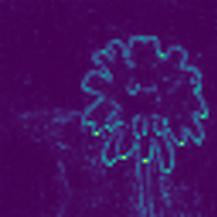}
  \includegraphics[width=0.105\linewidth]{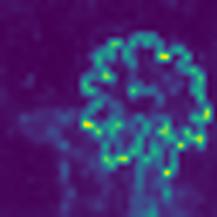}
  \includegraphics[width=0.105\linewidth]{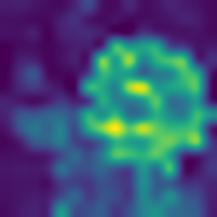}
  \includegraphics[width=0.105\linewidth]{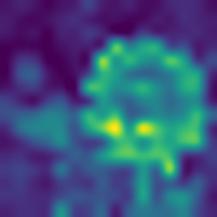}
  \includegraphics[width=0.105\linewidth]{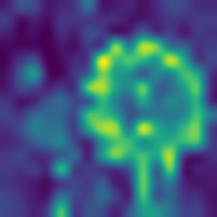}
  \includegraphics[width=0.105\linewidth]{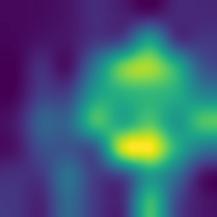}
\vspace{-2mm}
\vspace{-2\baselineskip}
\end{tabular}}
\hspace{20pt}\captionof{figure}{Variance Maps across channels for intermediate features of ControlNet for different modalities. As we can see the variance map  has high values where the condition is present and has low values for locations where the condition is absent.}\label{fig:sup1b}
\end{figure*}%
\subsection{Expressiveness of Control Modules}

The task of choosing the most suitable feature from two at the same spatial location for fusion poses a significant challenge. To effectively merge intermediate features from multiple models, it becomes essential to employ a metric that quantifies the strength or expressiveness of the condition at that specific location. For instance, as illustrated in \Cref{fig:introfig}, where multiple conditions are showcased, the network encounters the challenge of prioritizing the most expressive condition at each spatial point. Consider a scenario with a sketch map and a segmentation map, as shown in \Cref{fig:introfig}(b); the network needs to identify and prioritize the condition that provides the most significant information for that specific location. Our investigations led us to a straightforward yet effective observation.

\begin{observation}
The expressiveness of a condition provided to the diffusion model can be quantified using the variance maps from the model's intermediate layers.
\end{observation}

To demonstrate this, we examine variance maps of different control signals originating from an identical image, noise, and text prompt, as depicted in \Cref{fig:sup1b}. These maps reveal higher activation in areas pertinent to the specific condition, suggesting that variance maps can serve as a measure of conditioning strength needed for each spatial area. Therefore, when comparing two models that process the same input noise but under different conditions, the significance of each condition at every spatial point can be approximated by the activation level of the variance map at that location. This approach provides a reliable means to gauge the relative importance of each condition, enabling more informed feature selection for fusion.

\subsection{MaxFusion}

This section delves into the fusion mechanism among the intermediate outputs from various task models passed to the T2I model and the subsequent layer. The InterModel fusion occurs through a merging process, detailed in the following paragraphs.

Let's begin with a straightforward scenario: fusing two independent modalities. Let the ControlNet blocks handle different modalities ${M_1,M_2}$. A basic approach for combining these modalities is naive averaging, represented as:
\begin{equation}
f_{avg} = \frac{f_1+f_2}{2}
\end{equation}
where $f_1$ and $f_2$ are intermediate feature outputs across the models. However, such a simplistic fusion method poses a challenge. Consider an image where the left half contains only depth details while the right half features an edge. A naive averaging scheme would dilute the significance of each feature by averaging them equally. An ideal scheme would adjust spatially, giving more weight to the modality input value at a given location $(j,k)$. In cases of overlapping conditions, such as when conditions denote the same object and spatial location, as illustrated in the robot example in \Cref{fig:introfig}, it is desirable to incorporate characteristics from both modalities during merging. To achieve this, we assess the correlation of features at each spatial location. For highly correlated features, we adopt a weighted summation approach, formulated as:
\begin{equation}
f_{eff}^{(j,k)} = \frac{f_1^{(j,k)}+f_2^{(j,k)}}{2}, \text{if } \rho^{(j,k)} > \delta
\end{equation}
where $\delta$ is a predefined threshold and $\rho^{(j,k)}$ represents the correlation value between $f_1^{(j,k)}$ and $f_2^{(j,k)}$ at each spatial location and is defined as:
\begin{equation}
\rho^{(j,k)} = \frac{f_1^{(j,k)} \cdot f_2^{(j,k)}}{|f_1^{(j,k)}| \cdot |f_2^{(j,k)}|}
\end{equation}
If the correlation falls below the threshold, we prioritize the modality with the most spatial information for Stable Diffusion. This prioritization is determined by the feature variance at the location. To ensure fairness across modalities with differing absolute variance values, we propose a relative standard deviation measure by normalizing standard deviation values across spatial locations:
\begin{equation}
\hat{\sigma}_i^{(j,k)}= \frac{\sigma_i^{(j,k)}}{\sum{^{(j,k)}}\sigma_i^{(j,k)}}
\end{equation}

Hence, when the correlation is below the threshold, we define the aggregated feature as the one with the highest channel standard deviation:
\begin{equation}
f_{eff}^{(j,k)} = f_i^{(j,k)}, \text{max}_i(\hat{\sigma}_i^{(j,k)}), \text{if } \rho^{(j,k)} < \delta
\end{equation}
This formulation ensures automatic selection of the more relevant spatial condition, even in conflicting cases, providing effective conditioning.

\subsection{InterModel Unmerging}

In this section, we discuss the inter-model fusion process that proceeds to the next layer of the model. During the InterModel merging stage, we evaluate whether the correlation value exceeds the predefined threshold. For each spatial point not meeting this criterion, we verify if the feature being passed at that spatial location corresponds to the same index it belonged to and has the maximum variance. If the element itself possesses the maximum variance, we pass it as such; otherwise, we rescale the feature vector to have the same standard deviation before merging.
\begin{equation}
f_{i}^{(j,k)}=
\begin{cases}
f_i^{(j,k)}             & \text{$i =\text{max}_i(\hat{\sigma}_i^{(j,k)})$}\\
\frac{\sigma_i^{(j,k)}}{\sigma_{max}^{(j,k)}}f_{max}^{(j,k)}& \text{$otherwise$}\
\end{cases}
\label{eq:var}
\end{equation}
where,
\begin{equation}
f_{max}^{(j,k)} =f_i^{(j,k)};  \text{$i =\text{max}_i(\sigma_i^{(j,k)})$}
\end{equation}
 Please note that renormalizaiton of variance in \Cref{eq:var} reactivates regions that are otherwise diminished during the estimation of $f_{eff}$. This effect of vanishing feature strength is negated and rescaled to improve the performance. We detail the overall algorithm in \Cref{test algo}, where SD denotes Stable Diffusion and $l1,l2$ are model layers and we name the input modalities as $c1,c2$.

\begin{algorithm}
\caption{MaxFusion for scaling two modalities}
\label{ref:algo2}
\begin{algorithmic}[1]
\renewcommand{\algorithmicrequire}{\textbf{Input:}}
\renewcommand{\algorithmicensure}{\textbf{Initialize:}}
\Require Model 1 layers $l_1$, Model 2 layers $l_2$ , threshold $\delta$, \
input condition $c_1$, input condition $c_2$
\For{$l = {1}, \ldots, L$}
\If{$l=1$}
\State $f_1= l_1(c_1)$
\State $f_2= l_2(c_2)$
\Else
\If{$\rho(f_1^{(j,k)},f_2^{(j,k)})\geq\delta$}
\State $f_{eff}^{(j,k)}=\frac{f_1^{(j,k)}+f_2^{(j,k)}}{2}$
\Else
\State $f_{eff}^{(j,k)} = f_i^{(j,k)}, \text{max}_i(\hat{\sigma}_i^{(j,k)})$
\EndIf
\State  SD decoder $\leftarrow f{eff}$
\If{$\rho(f_1^{(j,k)},f_2^{(j,k)})\geq\delta$}
\State $fi^{(j,k)}= f_{eff}^{(j,k)}; i=1,2$
\Else
\State    $f_{i}^{(j,k)}=
\begin{cases}
f_i^{(j,k)}             & \text{$i =\text{max}_i(\hat{\sigma}_i^{(j,k)})$}\\
\frac{\sigma_i^{(j,k)}}{\sigma_{max}^{(j,k)}}f_{max}^{(j,k)}& \text{$otherwise$}\\

\end{cases}$
\EndIf
\EndIf

\EndFor
\
\Return $f_{eff}$
\end{algorithmic}
\label{test algo}
\vspace{-0.1cm}
\end{algorithm}

\section{Experiments and Results}

\subsection{Implementation Details}

All experiments are conducted on a single NVIDIA A5000 GPU utilizing off-the-shelf models. We employ Stable Diffusion-v1.5~\cite{rombach2022high} as the T2I model. For task models, we use ControlNet~\cite{zhang2023adding} and T2IAdapter~\cite{mou2023t2i} modules trained separately for different tasks. Throughout our experiments, we maintain a consistent diffusion duration of 50 steps. Sampling is facilitated using the UniPC scheduler~\cite{zhao2023unipc}. In our experiments, we set the correlation threshold at a fixed value of $0.7$. Due to the absence of public datasets tailored for evaluating multi-modal generation with multiple modality inputs, we curate a synthetic dataset leveraging the COCO dataset~\cite{lin2014microsoft}. This dataset comprises modalities including depth maps, segmentation masks, HED, and canny edge. Similar to the approach in ControlNet~\cite{zhang2023adding}, we utilize MiDAS~\cite{kopf2021robust} for depth map estimation, PiDiNet~\cite{su2021pixel}, the canny edge detector, and UPERNET~\cite{xiao2018unified} for generating maps of individual modalities. We curate a subset of 2000 images from the COCO validation set for experimentation purposes. We utilize BLIP~\cite{li2022blip} to acquire corresponding captions and perform evaluations under dual sets of conditions. For evaluations, we leverage variants of existing networks as well as publicly available single-modality networks.

\subsection{Qualitative Evaluations}
\begin{figure*}[t!]
    \centering
    \begin{subfigure}[t]{0.115\linewidth}
      \captionsetup{justification=centering, labelformat=empty, font=scriptsize}
      \includegraphics[width=1\linewidth]{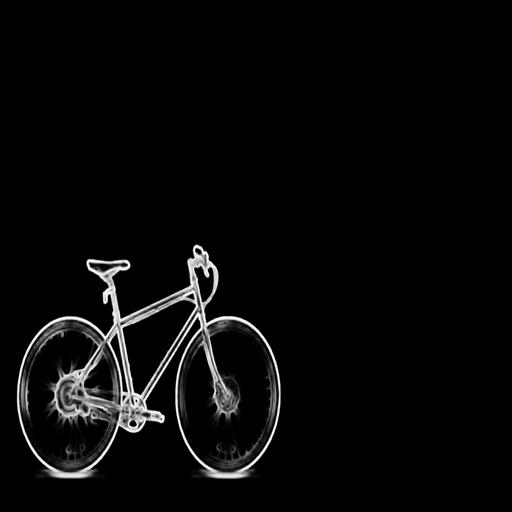}

      \includegraphics[width=1\linewidth, height=\linewidth]{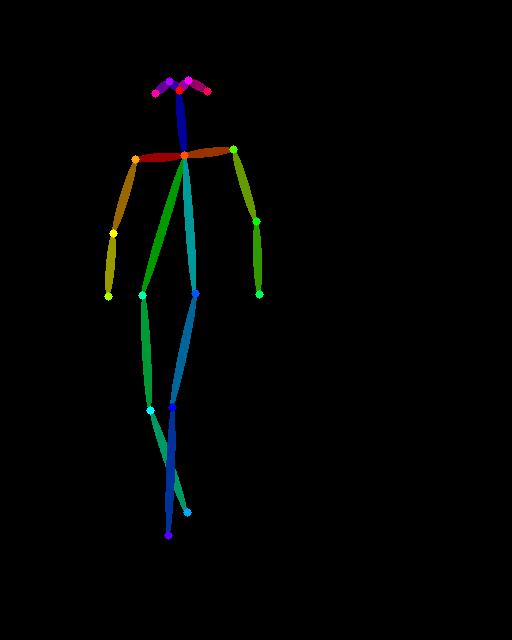}

      \includegraphics[width=1\linewidth]{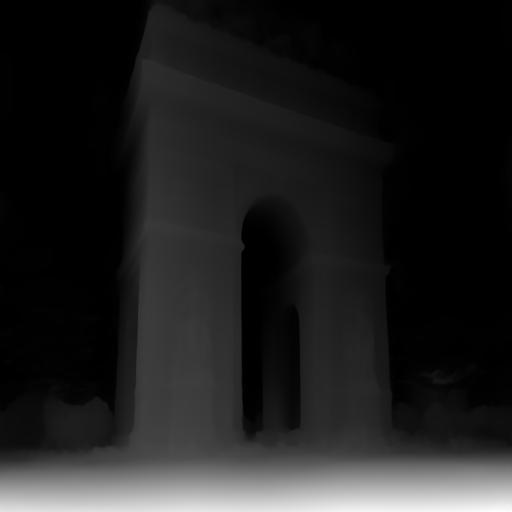}

      \caption{TASK-1}
    \end{subfigure}
    \begin{subfigure}[t]{0.115\linewidth}
      \captionsetup{justification=centering, labelformat=empty, font=scriptsize}
      \includegraphics[width=1\linewidth]{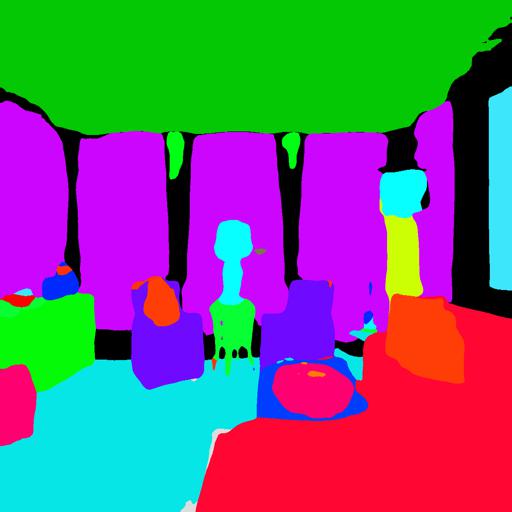}

      \includegraphics[width=1\linewidth]{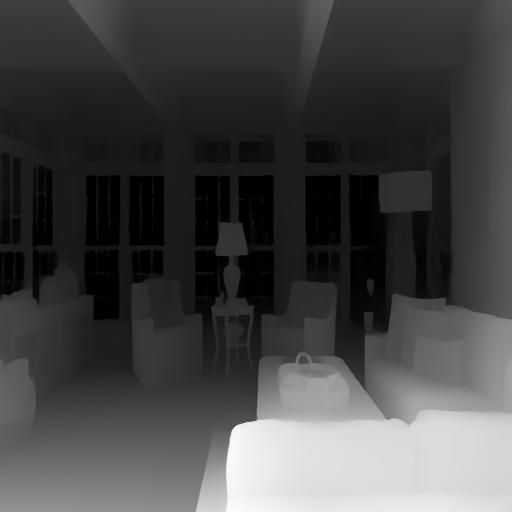}
      \includegraphics[width=1\linewidth,height=1\linewidth]{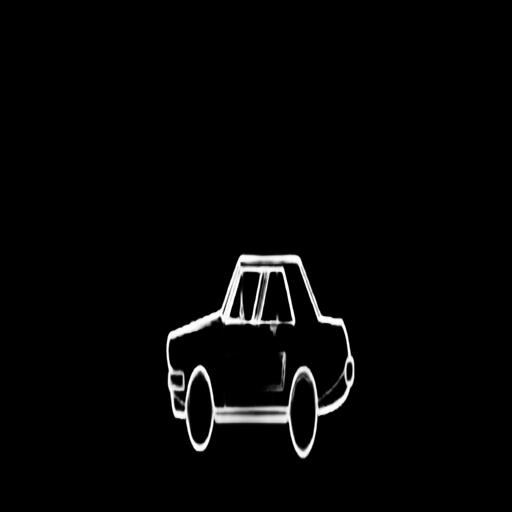}

      \caption{TASK-2}
    \end{subfigure}
    \begin{subfigure}[t]{0.115\linewidth}
      \captionsetup{justification=centering, labelformat=empty, font=scriptsize}
      \includegraphics[width=1\linewidth]{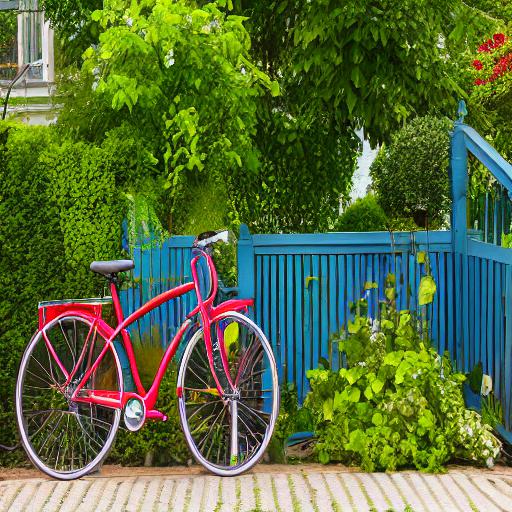}
      \includegraphics[width=1\linewidth]{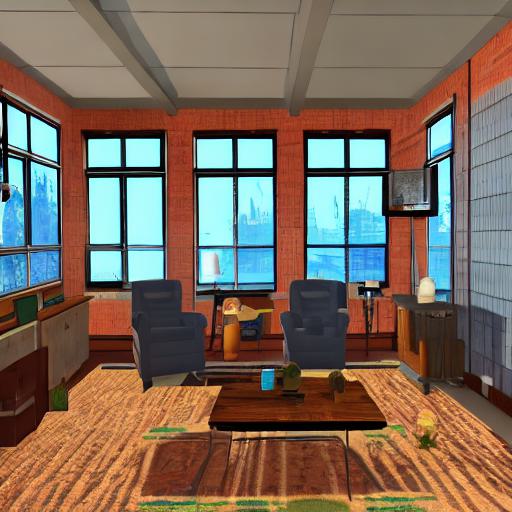}
        \includegraphics[width=1\linewidth]{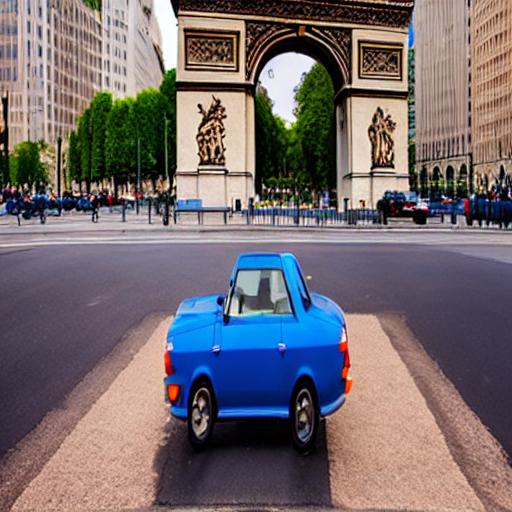}

      \caption{Multi-T2I Adapter}
    \end{subfigure}
    \begin{subfigure}[t]{0.115\linewidth}
      \captionsetup{justification=centering, labelformat=empty, font=scriptsize}

      \includegraphics[width=1\linewidth]{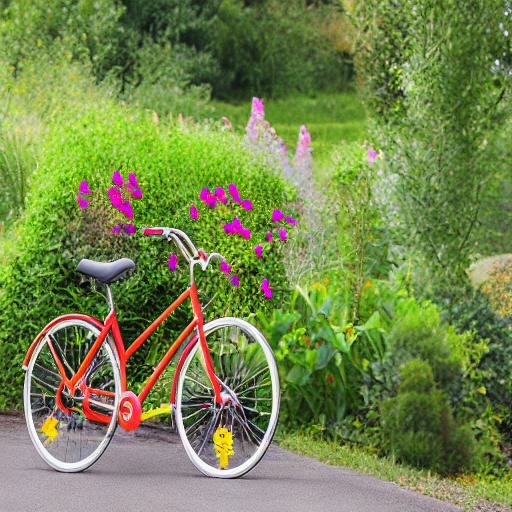}
      \includegraphics[width=1\linewidth]{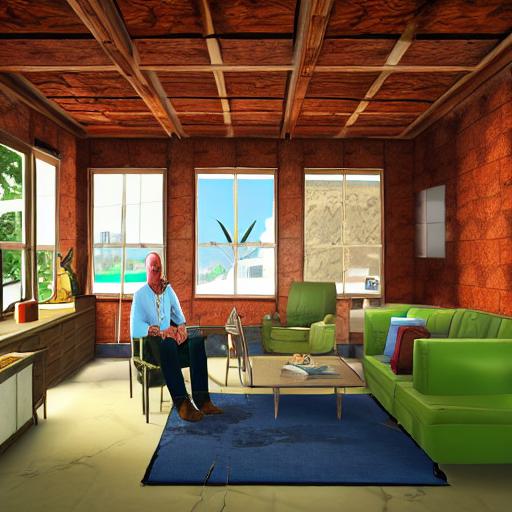}
      \includegraphics[width=1\linewidth]{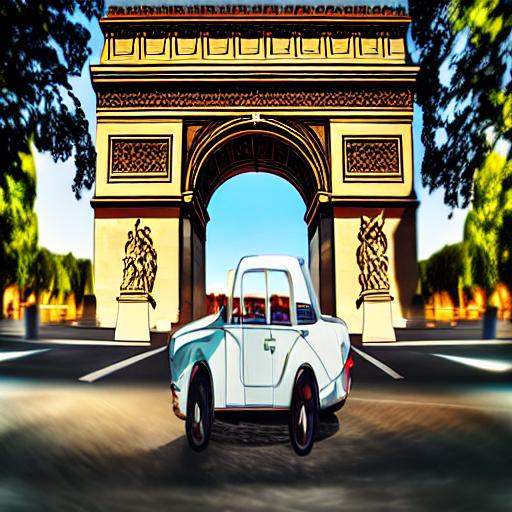}

      \caption{Multi-T2I Adapter}
    \end{subfigure}
    \begin{subfigure}[t]{0.115\linewidth}
      \captionsetup{justification=centering, labelformat=empty, font=scriptsize}
      \includegraphics[width=1\linewidth]{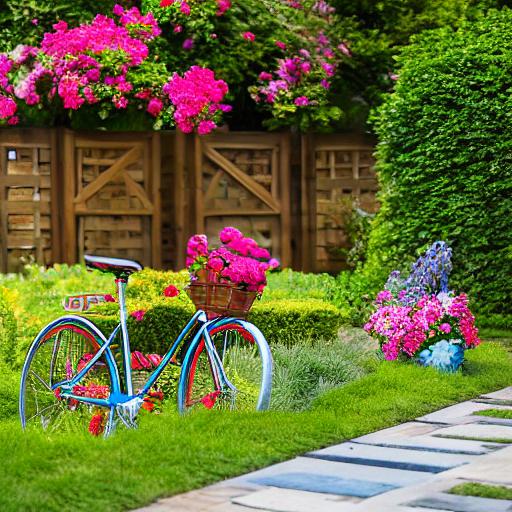}
       \includegraphics[width=1\linewidth]{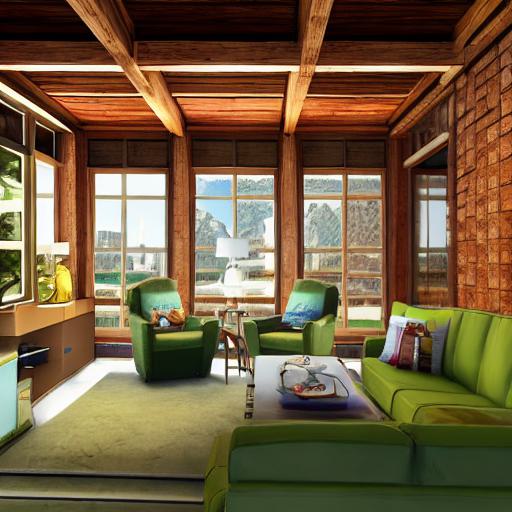}
       \includegraphics[width=1\linewidth]{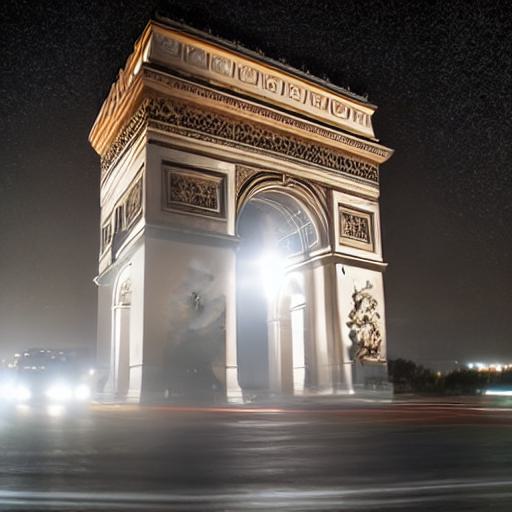}

      \caption{Multi-ControlNet}
    \end{subfigure}
    \begin{subfigure}[t]{0.115\linewidth}
      \captionsetup{justification=centering, labelformat=empty, font=scriptsize}
      \includegraphics[width=1\linewidth]{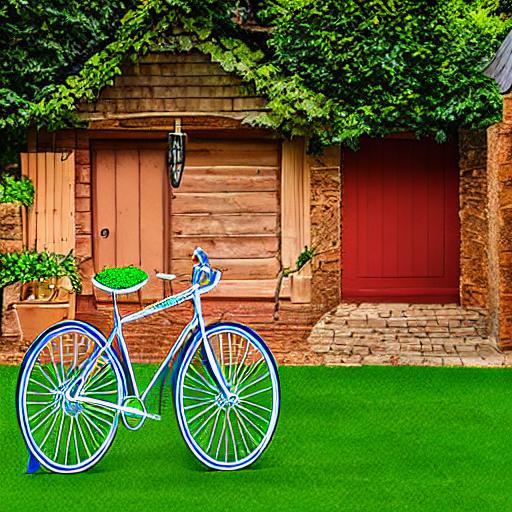}
       \includegraphics[width=1\linewidth]{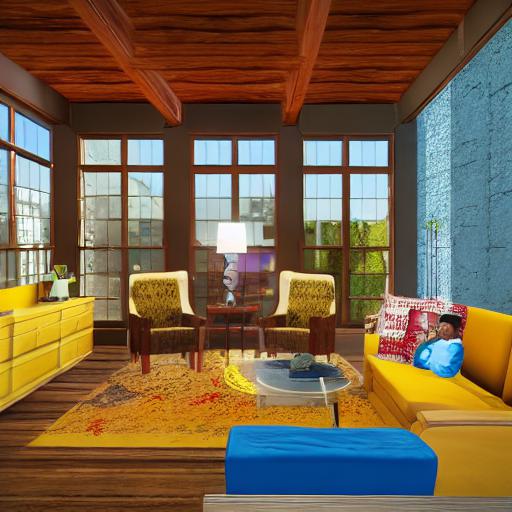}
       \includegraphics[width=1\linewidth]{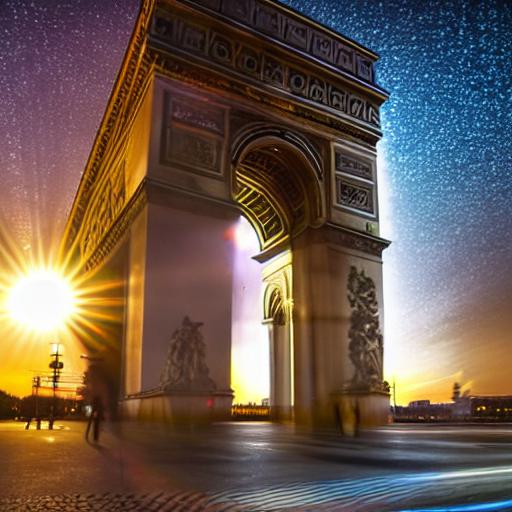}

      \caption{Multi-ControlNet}
    \end{subfigure}
    \begin{subfigure}[t]{0.115\linewidth}
      \captionsetup{justification=centering, labelformat=empty, font=scriptsize}

      \includegraphics[width=1\linewidth]{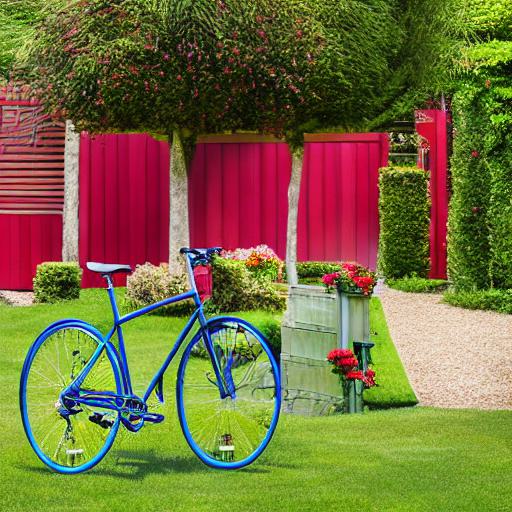}

        \includegraphics[width=1\linewidth]{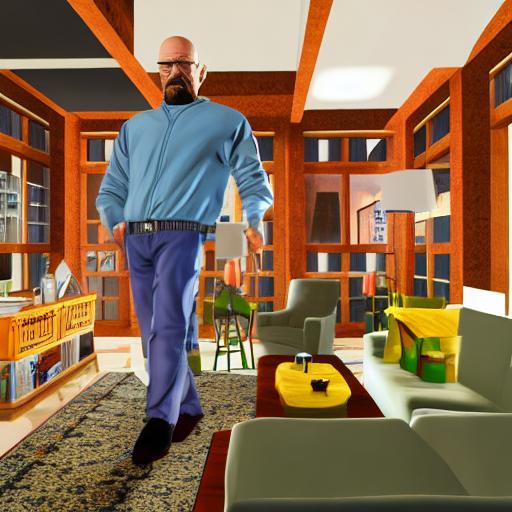}

      \includegraphics[width=1\linewidth]{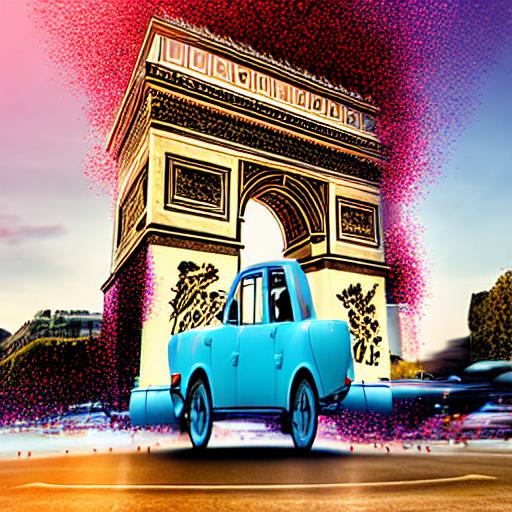}

      \caption{OURS}
    \end{subfigure}
    \begin{subfigure}[t]{0.115\linewidth}
      \captionsetup{justification=centering, labelformat=empty, font=scriptsize}
      \includegraphics[width=1\linewidth]{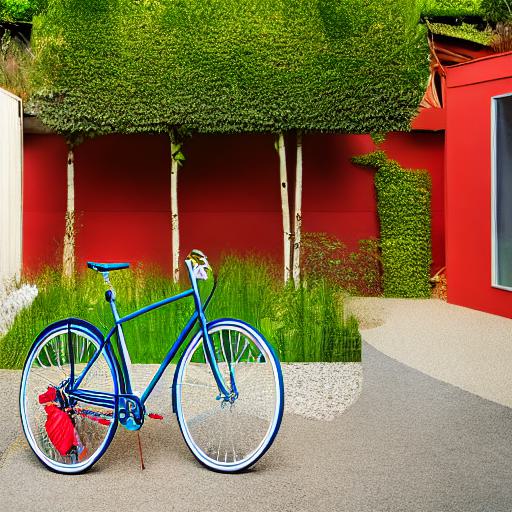}
        \includegraphics[width=1\linewidth]{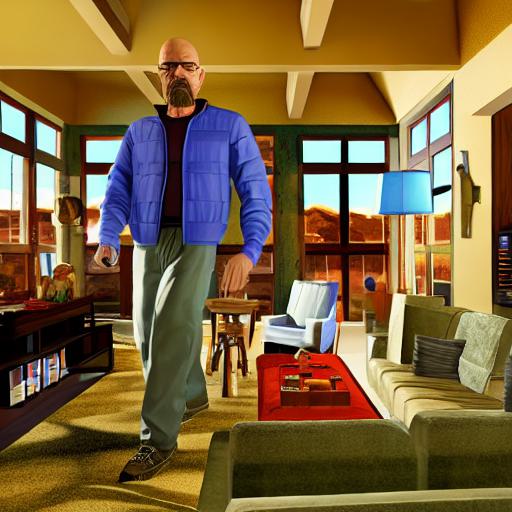}
      \includegraphics[width=1\linewidth]{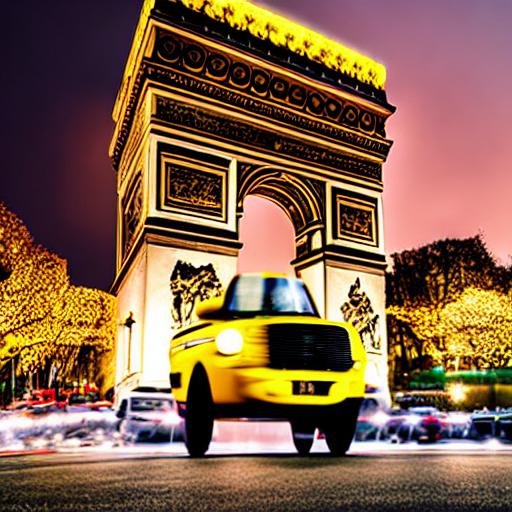}
      \caption{OURS}
    \end{subfigure}
    \vspace{-3mm}    \caption{\textbf{Qualitative comparisons for contradictory conditions from different modalities}. Figure illustrates the case of different conditions where out method shows a clear advantage over existing approaches. The text prompts used are   (1): \textit{"Walter White in a living room, GTA style." (2) "A bicycle in a garden." (3) "A car in front of arc de triomphe"}}
    \label{fig:facesematic}
  \end{figure*}

\begin{figure*}[t!]
    \centering
    \begin{subfigure}[t]{0.137\linewidth}
      \captionsetup{justification=centering, labelformat=empty, font=scriptsize}
      \includegraphics[width=1\linewidth]{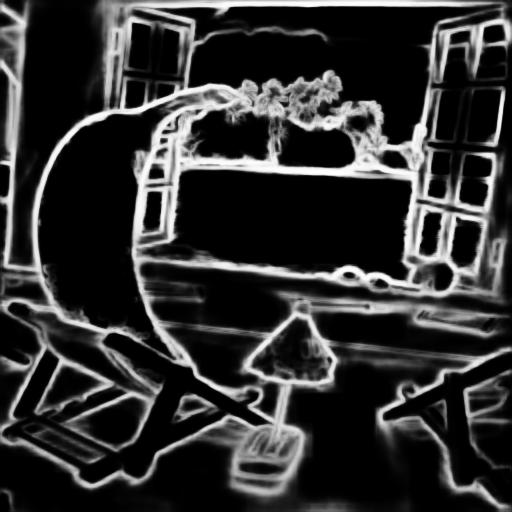}

      \includegraphics[width=1\linewidth]{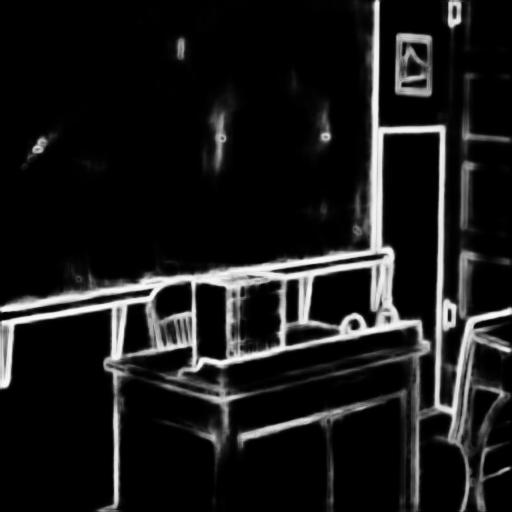}

      \includegraphics[width=1\linewidth]{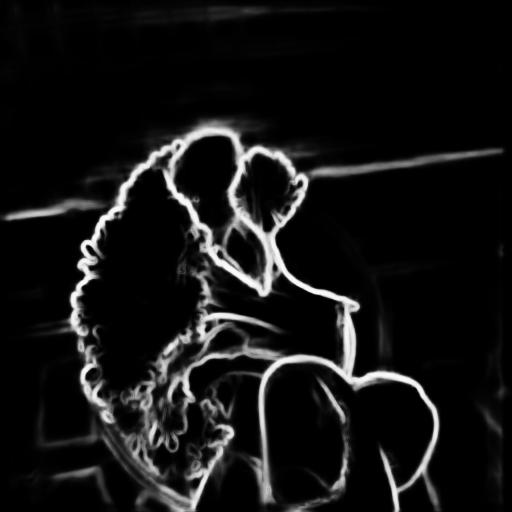}

      \includegraphics[width=1\linewidth]{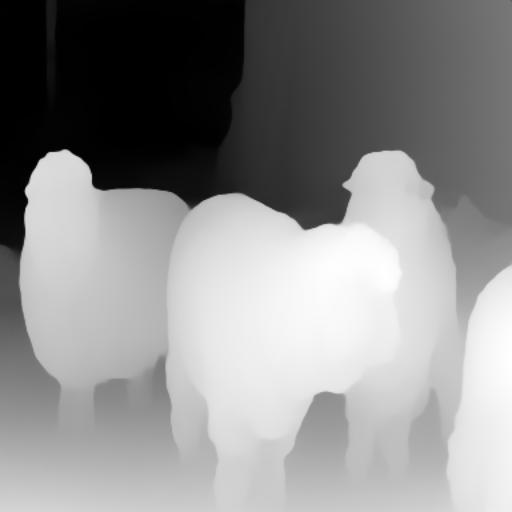}

      \caption{TASK-1}
    \end{subfigure}
    \begin{subfigure}[t]{0.137\linewidth}
      \captionsetup{justification=centering, labelformat=empty, font=scriptsize}
      \includegraphics[width=1\linewidth]{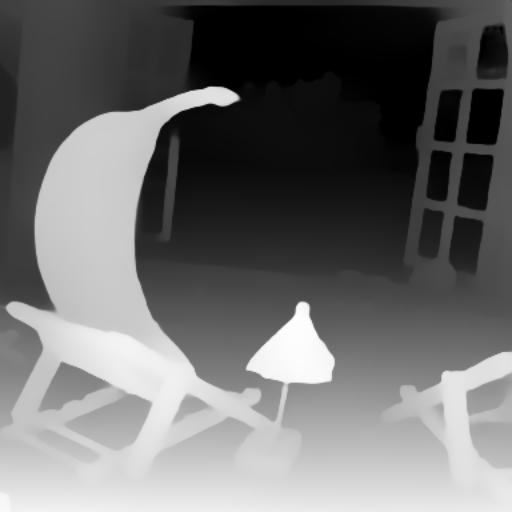}
      \includegraphics[width=1\linewidth]{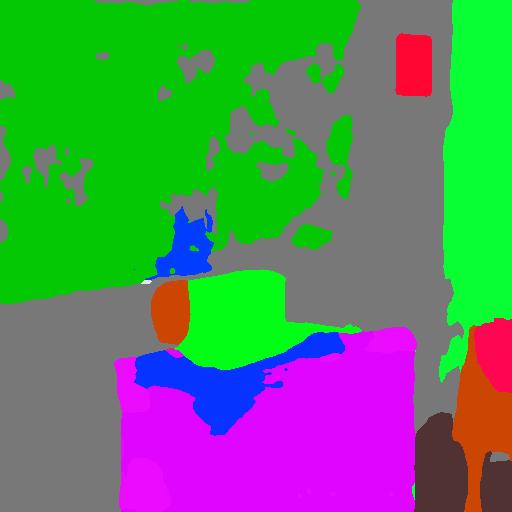}

      \includegraphics[width=1\linewidth]{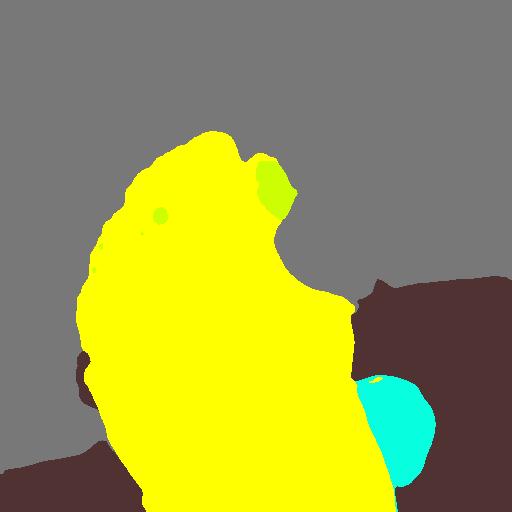}

      \includegraphics[width=1\linewidth]{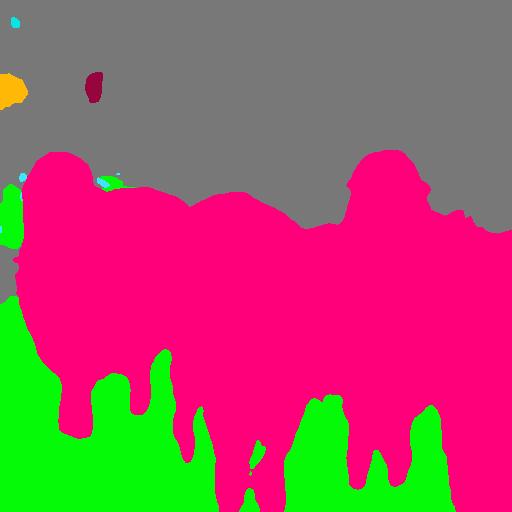}

      \caption{TASK-2}
    \end{subfigure}
    \begin{subfigure}[t]{0.137\linewidth}
      \captionsetup{justification=centering, labelformat=empty, font=scriptsize}
      \includegraphics[width=1\linewidth]{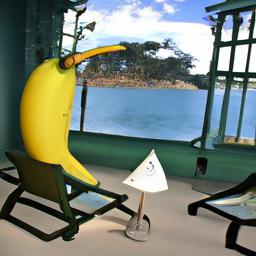}
      \includegraphics[width=1\linewidth]{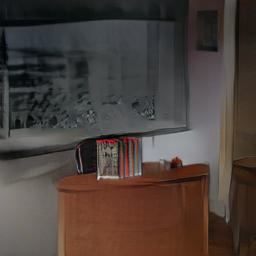}

      \includegraphics[width=1\linewidth]{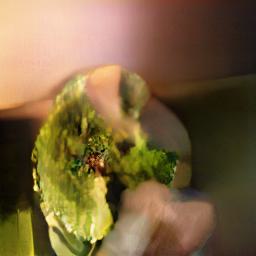}

      \includegraphics[width=1\linewidth]{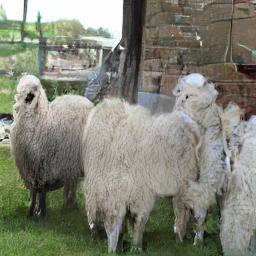}

      \caption{SPADE}
    \end{subfigure}
    \begin{subfigure}[t]{0.137\linewidth}
      \captionsetup{justification=centering, labelformat=empty, font=scriptsize}

      \includegraphics[width=1\linewidth]{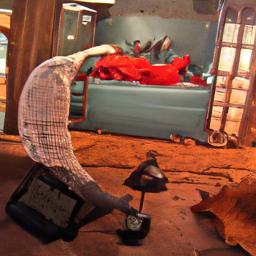}
      \includegraphics[width=1\linewidth]{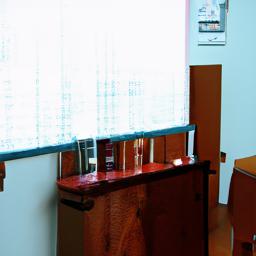}
      \includegraphics[width=1\linewidth]{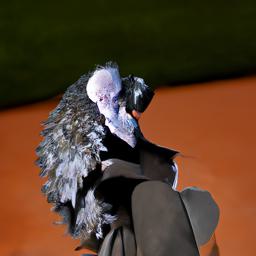}

      \includegraphics[width=1\linewidth]{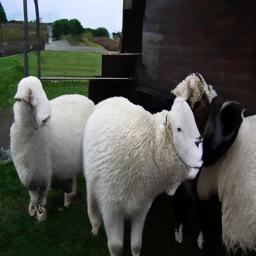}

      \caption{PITI}
    \end{subfigure}
    \begin{subfigure}[t]{0.137\linewidth}
      \captionsetup{justification=centering, labelformat=empty, font=scriptsize}
      \includegraphics[width=1\linewidth]{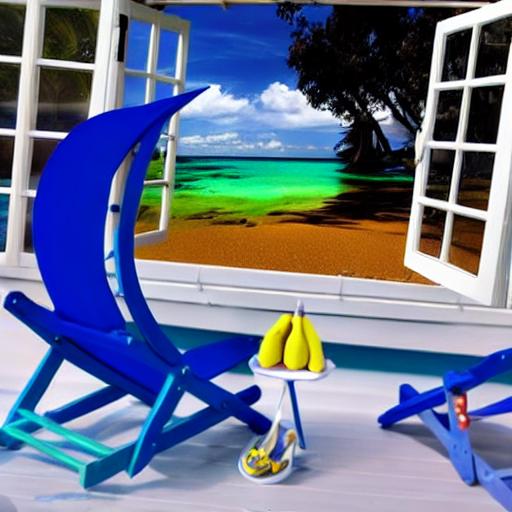}
      \includegraphics[width=1\linewidth]{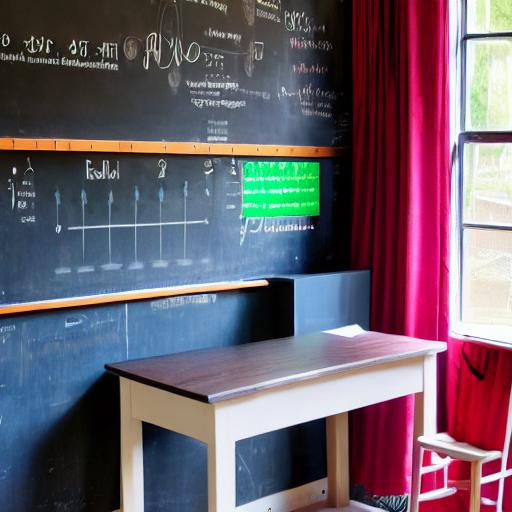}
      \includegraphics[width=1\linewidth]{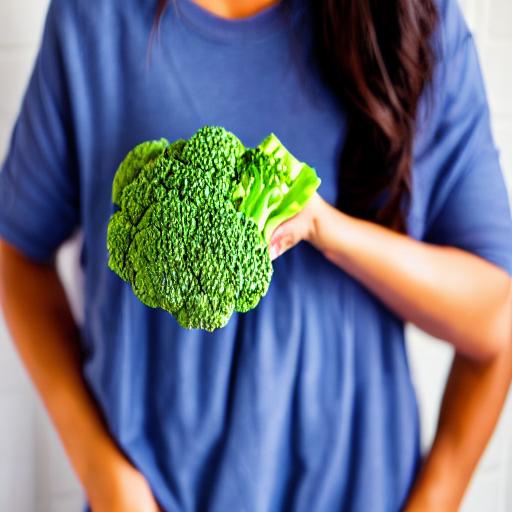}

      \includegraphics[width=1\linewidth]{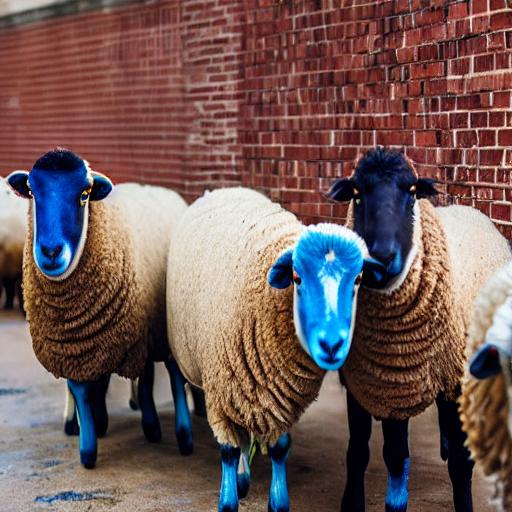}

      \caption{Multi-T2I Adapter}
    \end{subfigure}
    \begin{subfigure}[t]{0.137\linewidth}
      \captionsetup{justification=centering, labelformat=empty, font=scriptsize}
      \includegraphics[width=1\linewidth]{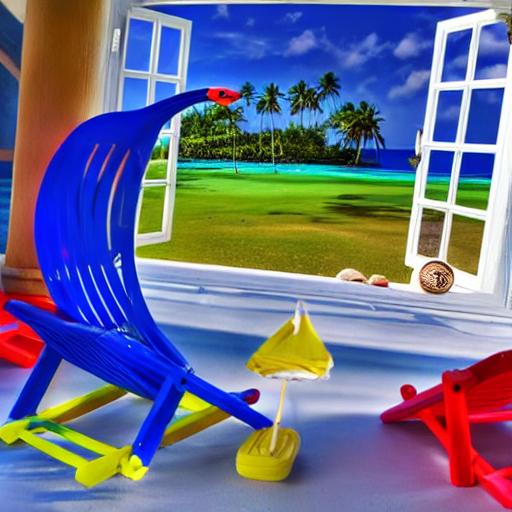}
      \includegraphics[width=1\linewidth]{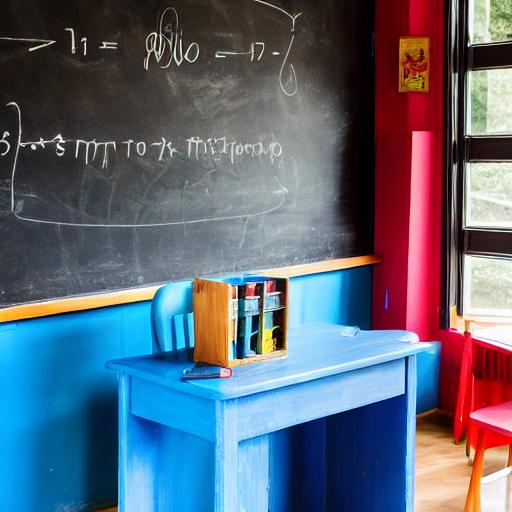}
      \includegraphics[width=1\linewidth]{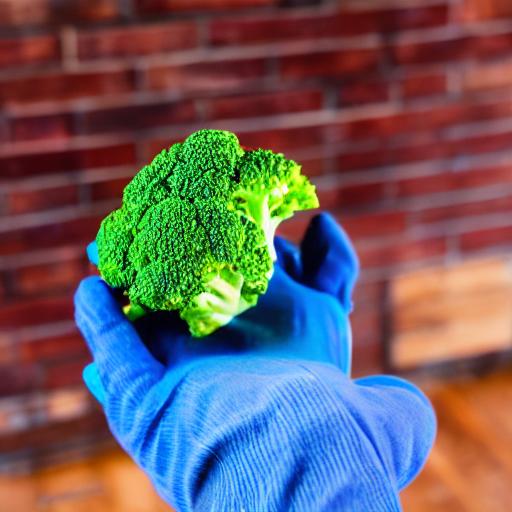}

      \includegraphics[width=1\linewidth]{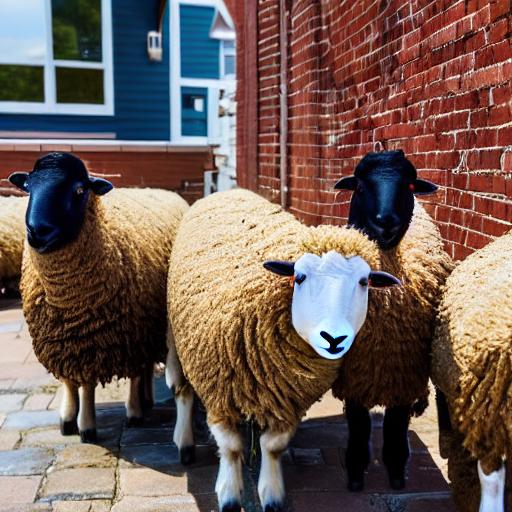}

      \caption{Multi-ControlNet}
    \end{subfigure}
    \begin{subfigure}[t]{0.137\linewidth}
      \captionsetup{justification=centering, labelformat=empty, font=scriptsize}

      \includegraphics[width=1\linewidth]{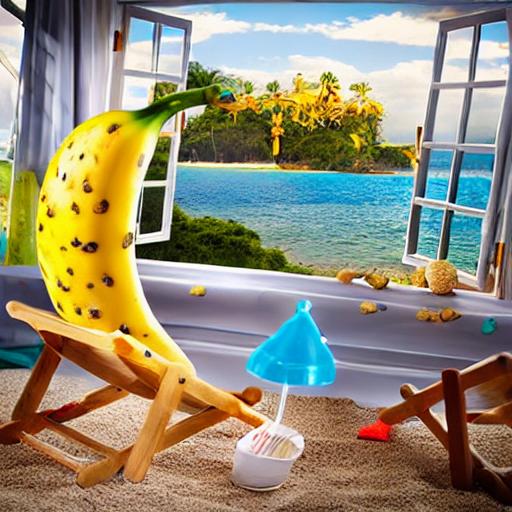}

      \includegraphics[width=1\linewidth]{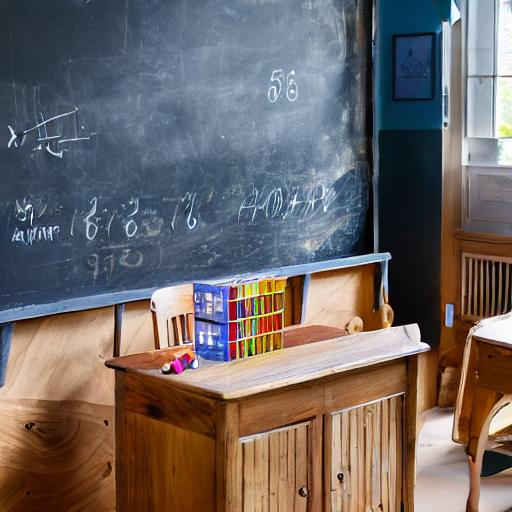}
      \includegraphics[width=1\linewidth]{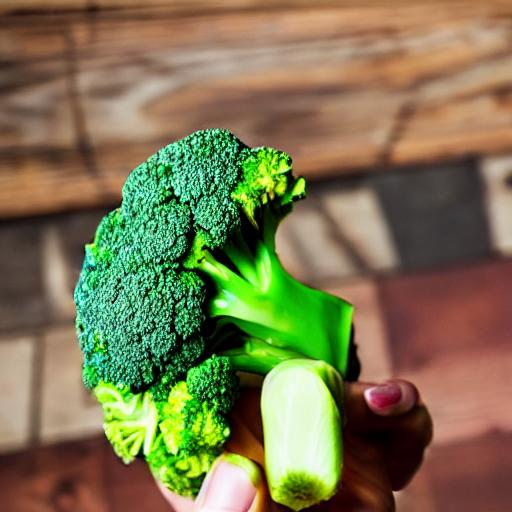}

      \includegraphics[width=1\linewidth]
{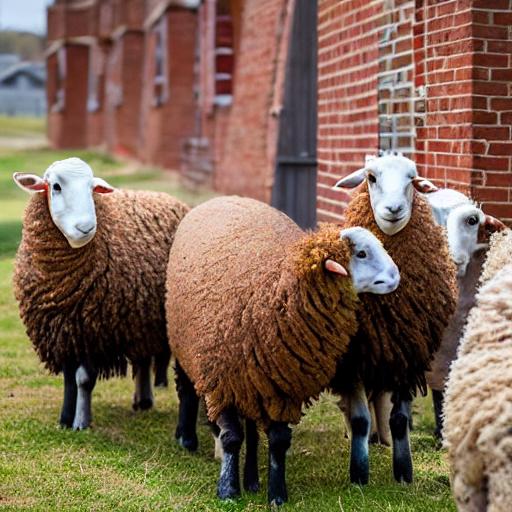}

      \caption{OURS}
    \end{subfigure}
    \vspace{-3mm}    \caption{\textbf{Qualitative comparisons for complimentary conditions in COCO dataset}. For the unimodal models we condition with its corresponding conditions. For multimodal models, we condition using both the Tasks. The text prompts used for conditioning are: from top  (1): \textit{"A banana  sitting on a chair and looking at the beach." (2) "A brocolli in a hand." (3) "A blackboard and table in a classroom" (4) "A couple of sheep standing"}}
    \label{fig:coco}
    \vspace{-2mm}
  \end{figure*}

We present results for multi-modal conditions in \Cref{fig:facesematic} and \ref{fig:coco}. Here, we compare with uni-modal networks SPADE~\cite{park2019SPADE} and PITI~\cite{wang2022pretraining}. For multi-modal conditioning, we implement ControlNet and T2IAdapter with different conditioning scales for the various input tasks. On the left, we display the included modalities. As observed, the single-modality segmentation model SPADE~\cite{park2019SPADE} fails to capture edges clearly, as it is trained solely with segmentation maps. Similarly, PITI often struggles with image quality in most cases, as its training data may not adequately represent complex scenarios like "A banana on a chair". The multi-modal networks T2I-Adapter and ControlNet generally perform well across most cases, yet they may falter in generating intricate details in challenging scenarios. Please refer to \Cref{fig:coco} for examples from the top two rows. On a broader scale, multi-Controlnet performs quite well but struggles to preserve key details when multiple conditions are present at the same location, as depicted in the provided examples. Our proposed method successfully transfers details and performs well in challenging cases while maintaining consistency with the text prompt. Additional illustrative examples are available in the supplementary material.

\begin{table}[tp]
\centering
\vspace{-3mm}
\scalebox{0.8}{
\begin{tabular}{c|c c c|c c c}
\toprule
Method & \multicolumn{3}{c|}{Pose, Seg} & \multicolumn{3}{c}{Pose, Depth} \\
\midrule
& FID $(\downarrow)$ & MSE-P $(\downarrow)$ & MSE-S $(\downarrow)$ & FID $(\downarrow)$ & MSE-P $(\downarrow)$ & MSE-D $(\downarrow)$ \\
\midrule
T2I-Adapter & \cellcolor{orange!50}55.61 & \cellcolor{orange!50}0.0140 & 0.1338 & 51.47 & \cellcolor{orange!50}0.0134 & 0.0603 \\
ControlNet & 63.46 & 0.0142 & \cellcolor{orange!50}0.1276 & \cellcolor{orange!50}48.18 & 0.0137 & \cellcolor{blue!25}0.0357 \\
Ours & \cellcolor{blue!25}46.72 & \cellcolor{blue!25}0.0135 & \cellcolor{blue!25}0.1080 & \cellcolor{blue!25}47.37 & \cellcolor{blue!25}0.0119 & \cellcolor{orange!50}0.0410 \\
\bottomrule
\end{tabular}
}
\caption{Metrics for contradictory conditions on COCO}
\label{tab:compgen7}
\vspace{-12mm}
\end{table}
\subsection{Quantitative Evaluations:}

\noindent\textbf{Quantitative Analysis for Contradictory Conditions on COCO}: Since existing datasets don't encompass contradictory inputs in COCO, and general strategies for deriving spatial conditioning maps do not provide such datasets, we created a synthetic dataset for evaluation purposes. We selected all COCO images containing humans and derived their pose maps. Additionally, we included an extra pose map sourced from another set of images obtained from the internet. Consequently, the pose maps now incorporate additional spatial features, while other spatial features express the scene. The quantitative analysis is presented in \Cref{tab:compgen7}.

\noindent\textbf{Quantitative Analysis for Complementary Conditions on COCO:} For experiments involving segmentation maps, we employ SPADE~\cite{park2019SPADE} as the single-modality model. For experiments involving sketches, we utilize PITI~\cite{wang2022pretraining} as the single-modality model. As for the choice of multi-modal models, we establish the baseline using T2IAdapter~\cite{mou2023t2i} with naive averaging, simultaneously feeding in multi-modality inputs. In experiments combining Depth-Segmentation, we employ NIQE~\cite{mittal2012making} and CLIP consistency~\cite{radford2021learning} as evaluation metrics. Moreover, to ensure conditioning consistency, we derive the conditions from the generated images and evaluate them using Mean Square Error against their ground truth values. Observations reveal that Unimodal networks perform admirably in terms of conditioning consistency. However, they fall short in terms of text consistency and the quality of generated images. On the other hand, multimodal networks exhibit superior performance and yield overall better metrics in terms of consistency. Notably, our approach demonstrates better CLIP scores compared to other evaluation methods, ensuring enhanced consistency across different text prompts.

\begin{table}[tbp]
\centering
\begin{minipage}[b]{0.48\linewidth}
  \centering
  \scalebox{0.6}{
\begin{tabular}{c|c|c c c c c}
\toprule
Method & Modality & FID$(\downarrow)$ & NIQE$(\downarrow)$ & Clip$(\uparrow)$ & Depth$(\downarrow)$ & Seg$(\downarrow)$\\
\midrule
SPADE\cite{park2019SPADE} & \emph{Uni} & 41.05 & 4.95 & 0.2813 & 0.0320 & 0.0923  \\
PITI-Mask\cite{wang2022pretraining} & \emph{Uni} & 35.52 & 6.20 & 0.2816 & 0.0269 & 0.0909 \\
\midrule
T2I-Adapter\cite{mou2023t2i} & \emph{Multimodal}  & 43.24 & 5.34 & \cellcolor{blue!25}0.2955 & 0.0554 & 0.1387  \\
Ours & \emph{Multimodal} & \cellcolor{blue!25}38.84 & \cellcolor{blue!25}5.34 & 0.2954 & \cellcolor{blue!25}0.0296 & \cellcolor{blue!25}0.1107  \\
\bottomrule
\end{tabular}}
\end{minipage}
\hfill
\begin{minipage}[b]{0.48\linewidth}
  \centering
  \scalebox{0.6}{
\begin{tabular}{c|c|c c c c c }
\toprule
Method & Modality & FID$(\downarrow)$ & NIQE$(\downarrow)$ & Clip$(\uparrow)$ & Depth$(\downarrow)$ & HED$(\downarrow)$\\
\midrule
PITI-Sketch\cite{liu2022compositional} & \emph{Uni} & 38.50 & 6.22 & 0.2669 & 0.0298 & 0.0417\\
\midrule
T2I-Adapter\cite{mou2023t2i} & \emph{Multimodal}  & 30.63 & 3.90 & \cellcolor{blue!25}0.2964 & 0.0417 & 0.1055  \\
Ours & \emph{Multimodal} & \cellcolor{blue!25}29.71 & \cellcolor{blue!25}2.86 & 0.2879 & \cellcolor{blue!25}0.0191 & \cellcolor{blue!25}0.0735 \\
\bottomrule
\end{tabular}}
\end{minipage}
\caption{\textbf{Quantitative comparisons for $\{Seg, Depth \rightarrow Image\}$ and$\{Hed, Depth \rightarrow Image\}$} respectively . }
\vspace{-7mm}
\end{table}

\begin{table}[htbp]
\centering

\scalebox{0.7}{
\begin{tabular}{c|c|c c c c c}
\toprule
Method&Modality&FID$(\downarrow)$&NIQE$(\downarrow)$&Clip$(\uparrow)$&Seg$(\downarrow)$& HED$(\downarrow)$\\
\midrule
SPADE\cite{park2019SPADE}&\emph{Uni}  &41.05&4.95&0.2813 &0.0923& 0.0772 \\
PITI-Mask\cite{wang2022pretraining}&\emph{Uni}&35.52  &6.20 &0.2816 &0.0909 &0.0844\\
PITI-Sketch\cite{wang2022pretraining}&\emph{Uni} &38.50 &6.22&0.2669  &0.1044& 0.0417\\
\midrule
T2I-Adapter\cite{mou2023t2i}&\emph{Multimodal} &40.14 &3.94&0.2922 &0.1506&0.1201  \\
Ours&\emph{Multimodal}  &\cellcolor{blue!25}37.21&\cellcolor{blue!25}2.84&\cellcolor{blue!25}0.2941 &\cellcolor{blue!25}0.0985 &\cellcolor{blue!25}0.0774  \\

\bottomrule
\end{tabular}
}
\caption{\textbf{Qualitative comparisons for $\{Hed, Seg \rightarrow Image\}$.} Quantitative results for different models for multimodal generation. Please note that the multimodal models have additional text conditioning capability}
\label{tab:compgen3}
\vspace{-4mm}
\end{table}

\section{Discussions}
\noindent{\textbf{Why is it advantageous to use the proposed variance scheme rather than naively adding the spatial conditions at the control input?} During our experiments, we observed that when both conditions are complementary, i.e., when we provide more than one condition at the same spatial location, naive averaging of the features works well for most cases. This is evident in Figure~\ref{fig:coco}, where we can observe satisfactory performance for the naive averaging scheme. However, when the conditions contradict each other, as shown in Figure~\ref{fig:facesematic}, the naive averaging scheme fails to incorporate details from both modalities into the scene effectively. Our approach offers a clear advantage over existing approaches in this scenario, opening up possibilities for zero-shot generation by deriving details of different objects from different types of input.

\noindent\textbf{Going beyond spatial conditioning:} Recently trained version of Multi-T2I-Adapter has shown the capability to include style conditioning as an additional modality. To show that MaxFusion works utilizing style as a modality as well, we present the corresponding results with different conditioning techniques in Figure~\ref{fig:style}. Additional results are provided in the supplementary material.

\begin{figure}[htbp]
\vspace{-3mm}
\centering
\includegraphics[width=\linewidth]{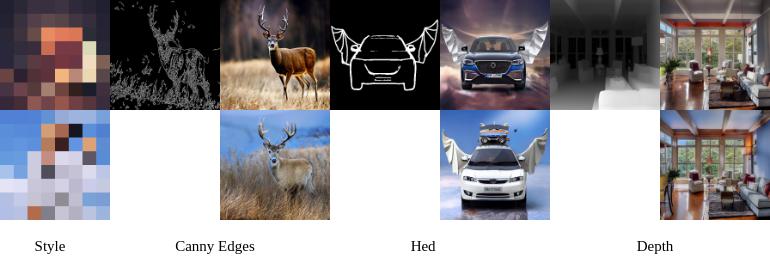}
\vspace{-3mm}
\caption{Qualitative results with style as a modality }
\label{fig:style}
\vspace{-2mm}
\end{figure}

\noindent\textbf{Scalability beyond two modalities}: In order to show that MaxFusion extends beyond two modalities, we show that one can change Maxfusion to combine two modalities first and add the next one on top of it and continue this process until all tasks are added. Let us assume we have three tasks, namely {Task1,Task2,Task3}. One can first consider two at a time and on the obtained intermediate feature, add the third task. The incremental addition scheme is the same for the intra-modal fusion and the inter-modal fusion. For extending to N tasks, one can follow the same procedure, performing an incremental addition of each task to the previous. We show illustrative examples for the same in Figure~\ref{fig:3modality}.

\begin{figure}[htp]
\vspace{-3mm}
\centering
\includegraphics[width=0.9\linewidth]{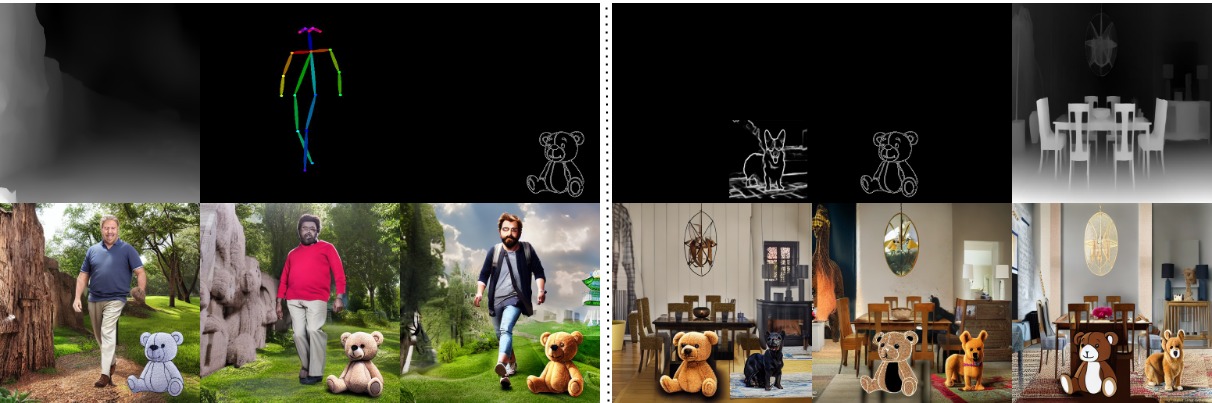}
\vspace{-3mm}
\caption{Qualitative results for extending to 3 modalities. }
\label{fig:3modality}
\vspace{-1mm}
\end{figure}

\begin{table}[htp]
\centering
\vspace{-1mm}
\scalebox{0.65}{
\begin{tabular}{c| c c c c  } 
\toprule
Method &\multicolumn{4}{c}{\{Seg, Depth, HED\}}\\
\midrule
 &FID$(\downarrow)$&MSE-S$(\downarrow)$&MSE-D$(\downarrow)$&MSE-H$(\downarrow)$ \\
\midrule
T2I-Adapter  &51.87&0.1354&0.0448 &0.1046  \\
ControlNet  &{50.47} &{0.1019}&{0.0362} &{0.0758}  \\
Ours  &\cellcolor{blue!25}{47.48} &\cellcolor{blue!25}{0.1000}&\cellcolor{blue!25}{0.0204} &\cellcolor{blue!25}{0.0692} \\
\bottomrule
\end{tabular}
}
\caption{Scalability metrics, scaling to 3 modalities.}
\label{tab:compgen6}
\vspace{-10mm}
\end{table}

\section{Ablation Studies}
\noindent\textbf{Effect of correlation Value for Multi-Modal Generation:} We present the results of varying the correlation threshold from $0$ to $1$ in Figure~\ref{fig:ablation}. When the correlation threshold is $0$, naive averaging of features occurs, leading to reduced conditioning strength of depth and edges and resulting in inconsistent depth effects in the scene. Specifically, at $\delta = 0$, the image lacks 3D depth-based effects in regions where the teddy bear is situated. Conversely, when the correlation threshold $\delta = 1$, more realistic images are generated. Varying the correlation threshold impacts sample quality, with a threshold above $\delta > 0.5$ yielding realistic images. To mitigate very high-frequency out-of-distribution components in the image, we set the value of $\delta$ to $0.7$. We observed that $\delta = 0.75$ works well for most cases.

  \begin{figure*}[htbp]
    \centering
        \begin{subfigure}[t]{0.137\linewidth}
      \captionsetup{justification=centering, labelformat=empty, font=scriptsize}
      \includegraphics[width=1\linewidth]{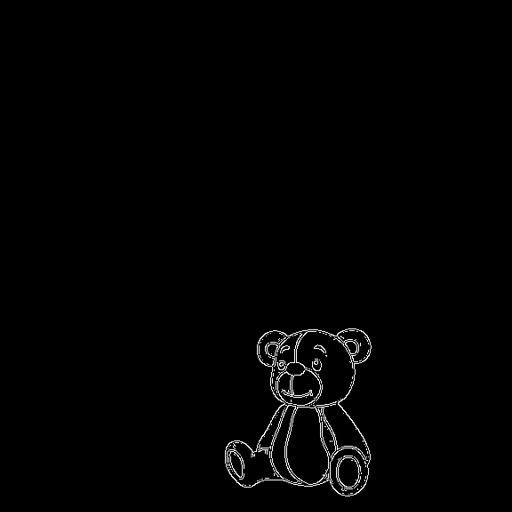}
      \caption{Edge Map}
    \end{subfigure}
    \begin{subfigure}[t]{0.137\linewidth}
      \captionsetup{justification=centering, labelformat=empty, font=scriptsize}
      \includegraphics[width=1\linewidth]{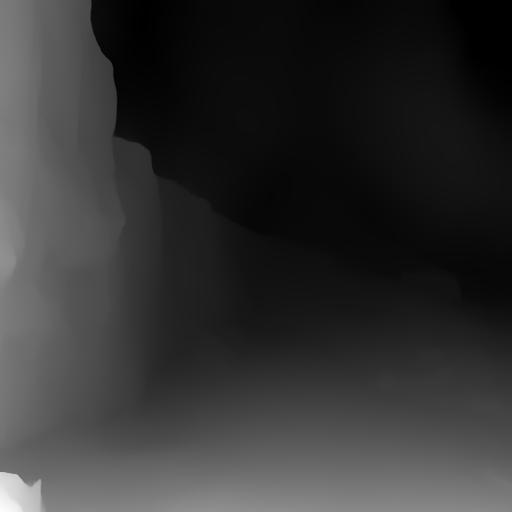}
      \caption{Depth Map}
    \end{subfigure}
    \begin{subfigure}[t]{0.137\linewidth}
      \captionsetup{justification=centering, labelformat=empty, font=scriptsize}
      \includegraphics[width=1\linewidth]{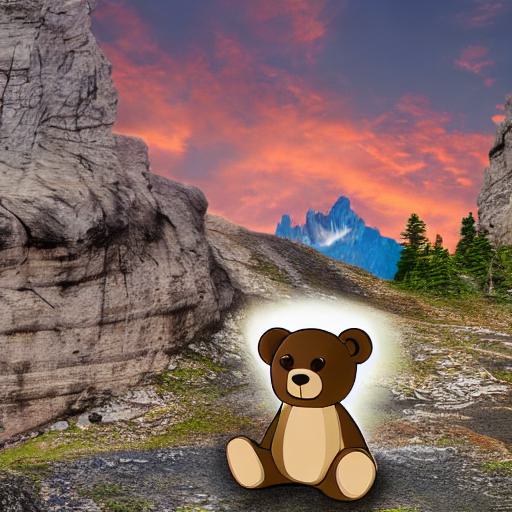}
      \caption{$\delta=0$}
    \end{subfigure}
        \begin{subfigure}[t]{0.137\linewidth}
      \captionsetup{justification=centering, labelformat=empty, font=scriptsize}
      \includegraphics[width=1\linewidth]{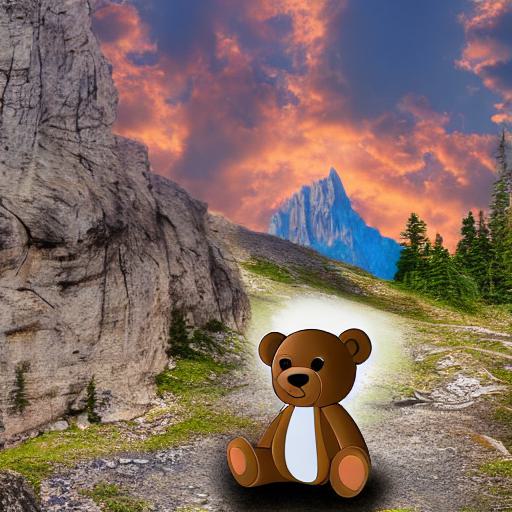}
      \caption{$\delta=0.25$}
    \end{subfigure}
       \begin{subfigure}[t]{0.137\linewidth}
      \captionsetup{justification=centering, labelformat=empty, font=scriptsize}
      \includegraphics[width=1\linewidth]{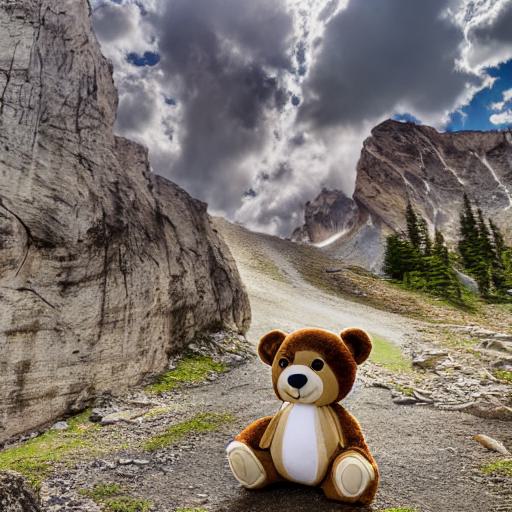}
      \caption{$\delta=0.5$}
    \end{subfigure}
        \begin{subfigure}[t]{0.137\linewidth}
      \captionsetup{justification=centering, labelformat=empty, font=scriptsize}
      \includegraphics[width=1\linewidth]{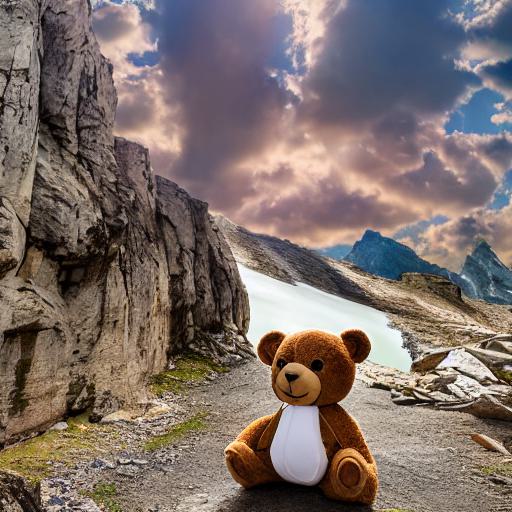}
      \caption{$\delta=0.75$}
    \end{subfigure}   
       \begin{subfigure}[t]{0.137\linewidth}
      \captionsetup{justification=centering, labelformat=empty, font=scriptsize}
      \includegraphics[width=1\linewidth]{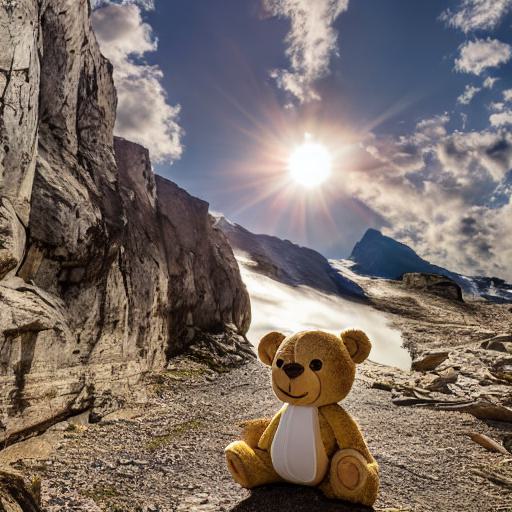}
      \caption{$\delta=1$}
    \end{subfigure}  
    \vspace{-3mm}    \caption{Ablation study corresponding to different correlation thresholds. Please note the level of 3D effect included to the scene as the threshold increases. The prompt used for captioning is \textit{Background:"A mountain". Foreground : "A teddy bear".}}
    \label{fig:ablation}
  \end{figure*}

\noindent\textbf{Effect of Multi-modal Conditioning:} We present the results on single conditioning and multi-conditioning in Figure~\ref{fig:multi}. When conditioned with edges alone, the image appears planar and visually unpleasing. Similarly, when conditioned only on depth, an additional wing is generated at a specific location. In cases like this, more aesthetically pleasing, high-quality images with fine-grained details from 2D edge conditions and 3D depth effects can be generated using multi-modal generation.

  \begin{figure*}[htbp]
    \centering
        \begin{subfigure}[t]{0.19\linewidth}
      \captionsetup{justification=centering, labelformat=empty, font=scriptsize}
      \includegraphics[width=1\linewidth]{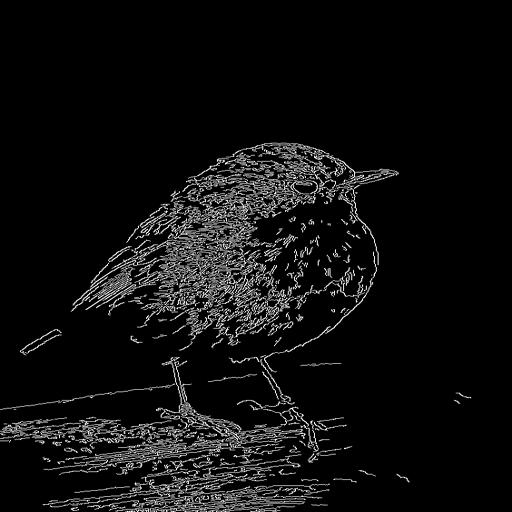}
      \caption{Edge Map}
    \end{subfigure}
    \begin{subfigure}[t]{0.19\linewidth}
      \captionsetup{justification=centering, labelformat=empty, font=scriptsize}
      \includegraphics[width=1\linewidth]{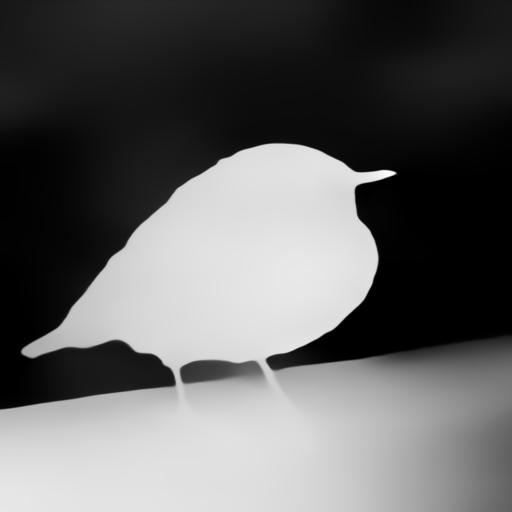}
      \caption{Depth Map}
    \end{subfigure}
    \begin{subfigure}[t]{0.19\linewidth}
      \captionsetup{justification=centering, labelformat=empty, font=scriptsize}
      \includegraphics[width=1\linewidth]{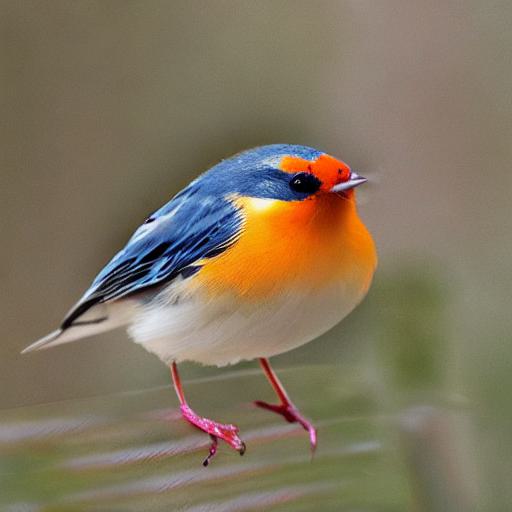}
      \caption{Edge Only}
    \end{subfigure}
        \begin{subfigure}[t]{0.19\linewidth}
      \captionsetup{justification=centering, labelformat=empty, font=scriptsize}
      \includegraphics[width=1\linewidth]{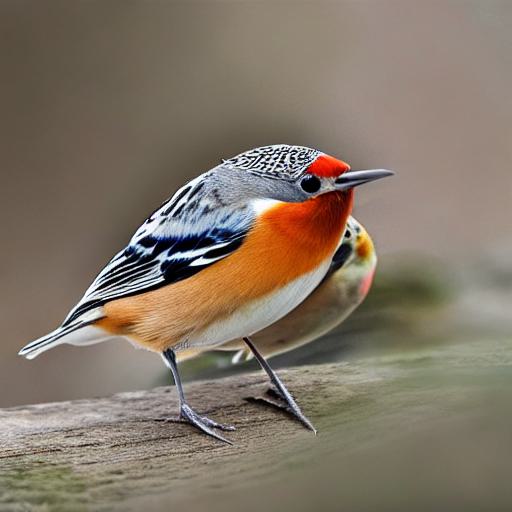}
      \caption{Depth only}
    \end{subfigure}
       \begin{subfigure}[t]{0.19\linewidth}
      \captionsetup{justification=centering, labelformat=empty, font=scriptsize}
      \includegraphics[width=1\linewidth]{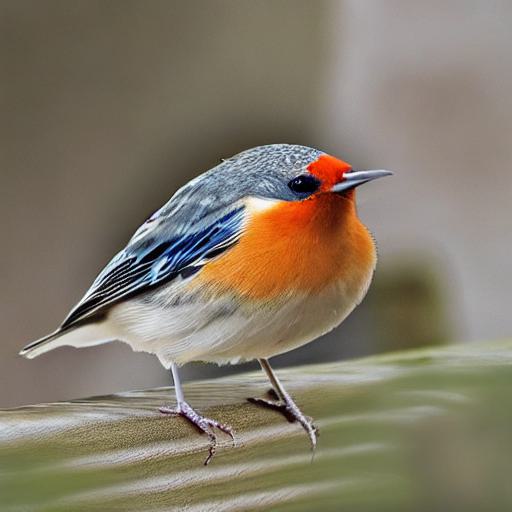}
      \caption{Multimodal}
    \end{subfigure}
        
    \vspace{-3mm}    \caption{Ablation study  drawing contrast between unimodal-models and multimodal models. The prompt used for captioning is \textit{"A bird".}}
    \label{fig:multi}
    \vspace{-5mm}
  \end{figure*}

\section{Limitations, Future Works, and Potential Impacts}

Our model inherently inherits the limitations of Stable Diffusion in generating human hands and faces. Additionally, we observed discrepancies in results when utilizing semantic maps, sometimes differing from the conditioned semantic map. However, this limitation stems from the training approach of ControlNet and T2I-Adapter. As state-of-the-art semantic segmentation networks cannot provide accurate masks for all open-world images, this limitation is inherent in segmentation-conditioned models as well. We provide examples in the paper and supplementary material to elucidate this issue. Furthermore, while our work is capable of generating various spatial inputs, it holds potential societal impacts. Also, one other limitation of our method is that as the number of conditions increases, a trade-off between the conditioning and the sampling fidelity arises. We will attempt to tackle this issue in future works by developing a much more robust algorithm. Like other generative models, our model also has potential societal impacts. The use of such models is at the user's discretion and care must be taken while using the model.

\section{Conclusion}

In this paper, we discover that the expressiveness of conditioning features is present in the variance maps of intermediate layers of diffusion models. Utilizing this as prior information,
we introduced a novel training-free strategy to scale up conditional diffusion models for multimodal generation using off-the-shelf models trained on individual tasks separately. For multimodal fusion, we proposed a new fusion paradigm based on variance across channels for different feature layers. The proposed fusion approach is lightweight and can be seamlessly integrated on top of existing models. This new fusion scheme empowers the model to perform zero-shot generation, which is otherwise not achievable with a single model. We further showed that the proposed method works beyond spatial conditioning and works with style conditioning as well. Moreover, Maxfusion can be seamlessly integrated to more than two conditioning modalities. We conducted experiments with four different spatial conditioning modalities to demonstrate the effectiveness of our method.

\bibliographystyle{splncs04}
\bibliography{references}

\title{Supplementary material: MaxFusion: Plug\&Play  Multi-Modal Generation in  Text-to-Image Diffusion Models
}

\titlerunning{MaxFusion}

\author{Nithin Gopalakrishnan Nair\inst{1}\orcidlink{0000-1111-2222-3333} \and
Jeya Maria Jose Valanarasu\inst{2} \and
Vishal M Patel\inst{1}\orcidlink{0000-0002-5239-692X}}

\authorrunning{Nair et al.}

\institute{Johns Hopkins University$^{1}$, Stanford University$^{2}$ 
\email{\{ngopala2, vpatel36\}@jhu.edu, jmjose@stanford.edu}}

\maketitle


\section{Maxfusion without variance renormalization}

We present a variant of Maxfusion without the variance renormalization procedure mentioned in Section 4.1 in the main paper here. The corresponding results are shown in \Cref{fig:multiablation}
\begin{algorithm}[H]
\caption{MaxFusion for scaling to two modalities}
\label{ref:algo1}
\begin{algorithmic}[1]
\renewcommand{\algorithmicrequire}{\textbf{Input:}}
\renewcommand{\algorithmicensure}{\textbf{Initialize:}}
\Require Model 1 layers $l_1$, Model 2 layers $l_2$ , threshold $\delta$, \\
input condition $c_1$, input condition $c_2$, SD input=\{\}
\For{$l = {1}, \ldots, L$}
   \If{$l=1$}
    \State $f_1= l_1(c_1)$
   \State $f_2= l_2(c_2)$
    \Else
   \If{$\rho(f_1^{(j,k)},f_2^{(j,k)})\geq\delta$}
   \State $f_{eff}^{(j,k)}=\frac{f_1l^{(j,k)}+f_2l^{(j,k)}}{2}$
   \Else
    \State $f_{eff}^{(j,k)} = f_il^{(j,k)}, \text{max}_i(\sigma_i^{(j,k)})$
    \EndIf
    \State  SD decoder $\leftarrow f_{eff}$
\If{$\rho(f_1^{(j,k)},f_2^{(j,k)})\geq\delta$}
   \State $fi^{(j,k)}= f_{eff}^{(j,k)}; i=1,2$
   \Else
    \State $f_{i}^{(j,k)} =  f_{i}^{(j,k)}; i=1,2$
    \EndIf
  \EndIf  
\EndFor
\\
\Return $f_{eff}$
\end{algorithmic}
\label{test algo1}
\end{algorithm}

\section{Extending beyond two tasks}
In order to extend MaxFusion beyond two tasks, one can change MaxFusion to combine two modalities first and add then add the next one on top of it and continue this process until all tasks are added.
For example, let us assume we have three tasks namely \{Task1,Task2,Task3\}
, one can first consider two at a time and on the obtained intermediate feature, add the third task. The incremental addition scheme is the same for the intramodel fusion and the intermodal fusion For extending to N tasks, one can follow the same procedure, performing an incremental addition of each task to the previous. we show illustrative examples of the same in Figures \ref{fig:supp1},\ref{fig:supp2}, \ref{fig:supp3}, \ref{fig:supp4}, \ref{fig:supp5}

\begin{figure*}[htb!]
    \centering
    \setlength{\tabcolsep}{0.5pt}
    {\small
    \renewcommand{\arraystretch}{0.5} 
    \begin{tabular}{c c c c c c c c c c}
    \captionsetup{type=figure, font=scriptsize}
\tabularnewline
  \includegraphics[width=0.18\linewidth]{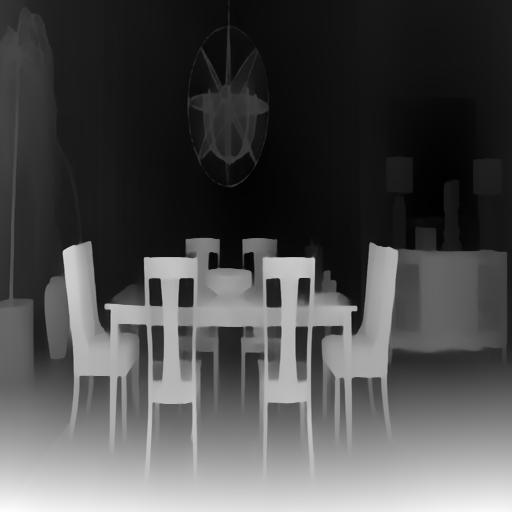}
   \includegraphics[width=0.18\linewidth]{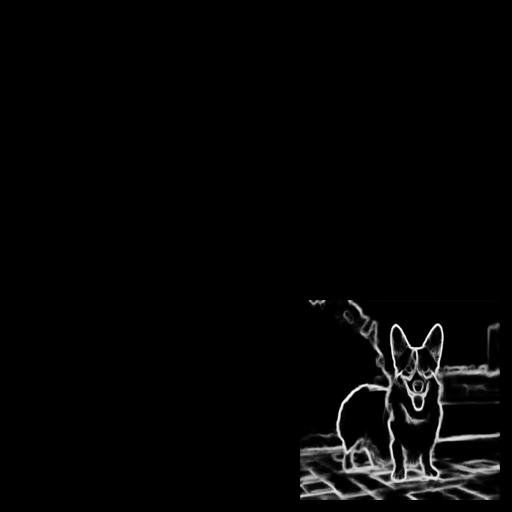}
  \includegraphics[width=0.18\linewidth]{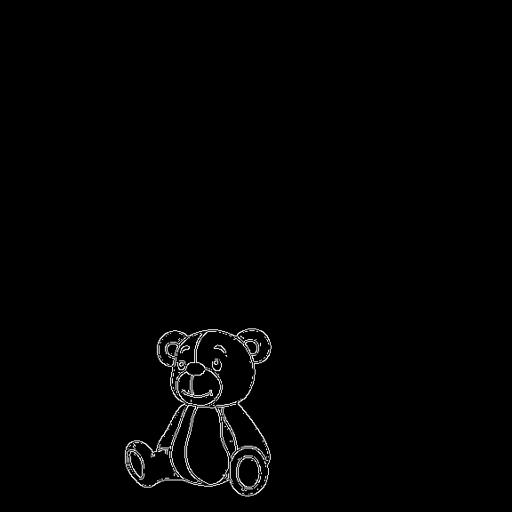}
   \tabularnewline

  \includegraphics[width=0.18\linewidth]{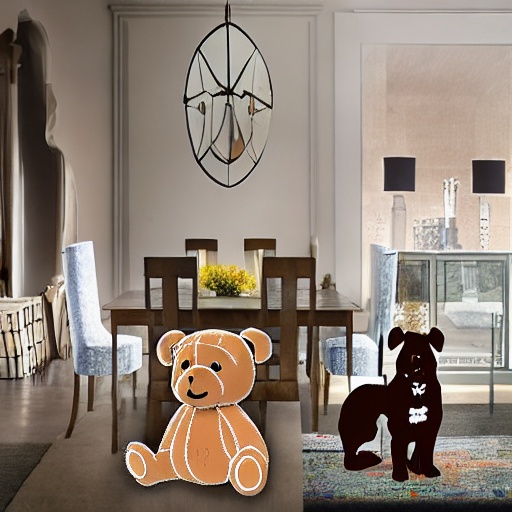}
  \includegraphics[width=0.18\linewidth]{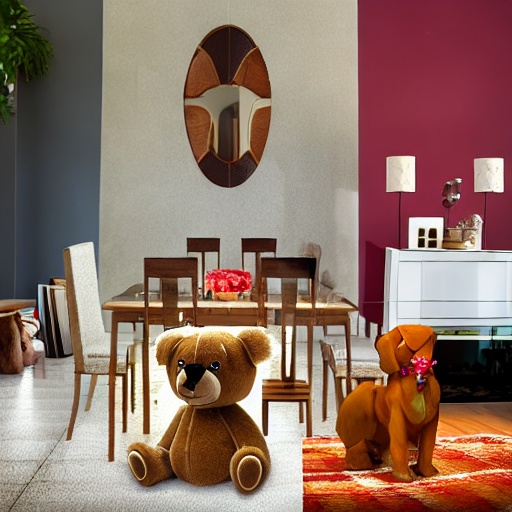}
  \includegraphics[width=0.18\linewidth]{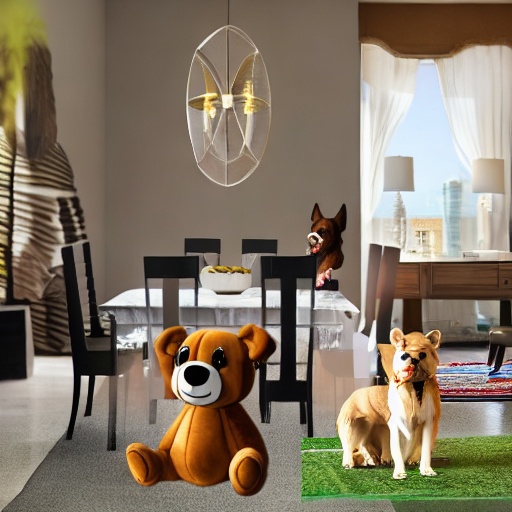}
  \includegraphics[width=0.18\linewidth]{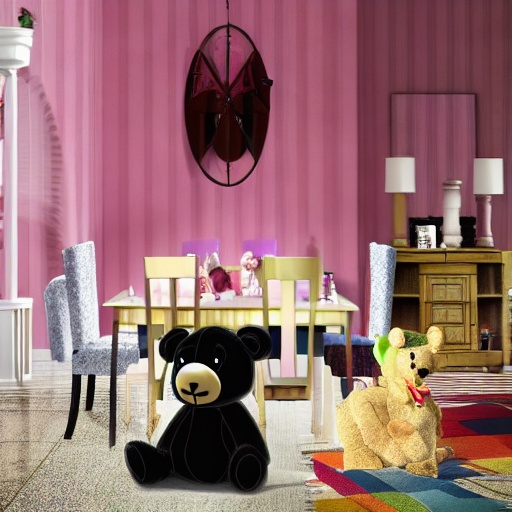}
  \includegraphics[width=0.18\linewidth]{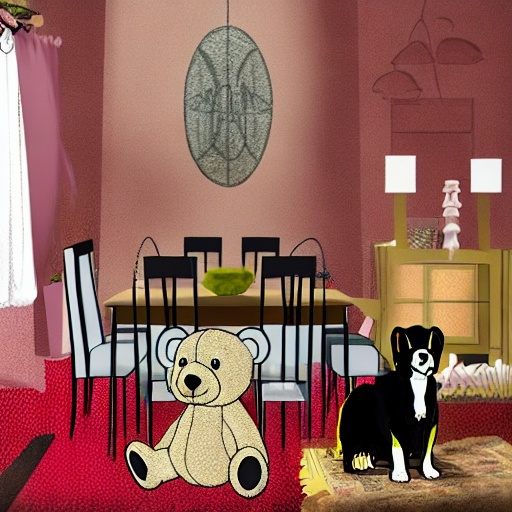}
 \tabularnewline
  \includegraphics[width=0.18\linewidth]{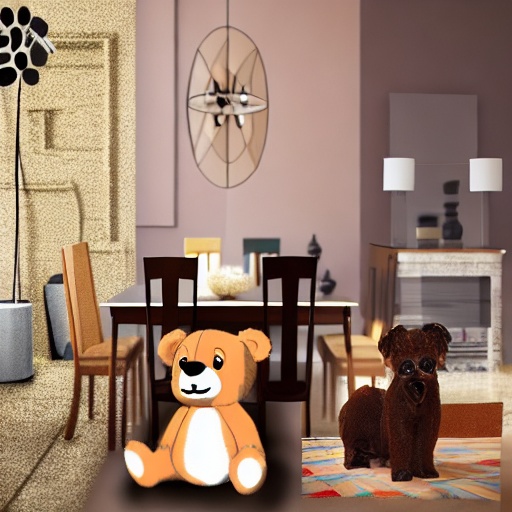}
  \includegraphics[width=0.18\linewidth]{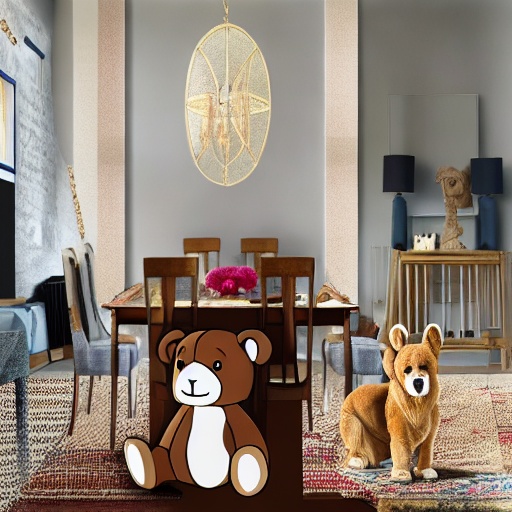}
  \includegraphics[width=0.18\linewidth]{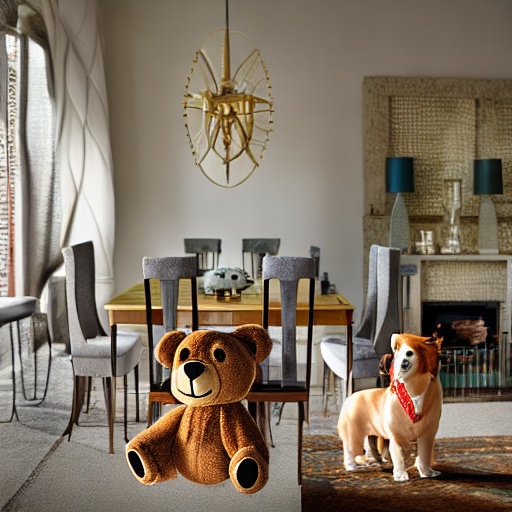}
  \includegraphics[width=0.18\linewidth]{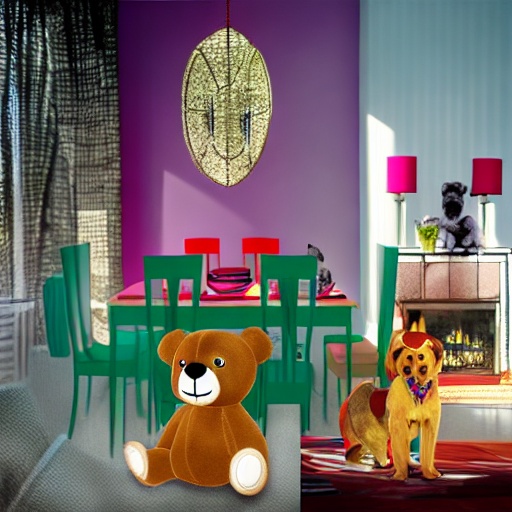}
  \includegraphics[width=0.18\linewidth]{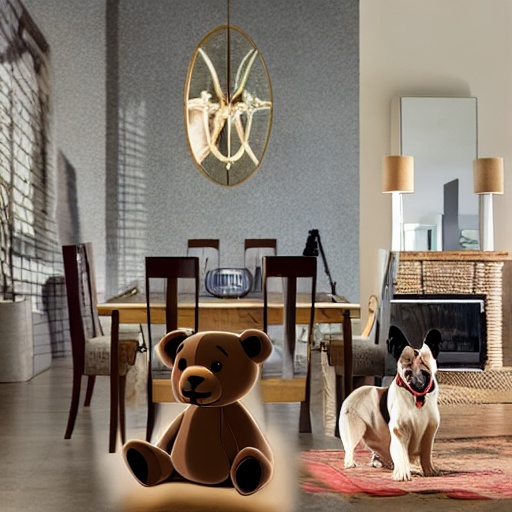}
\tabularnewline
  \includegraphics[width=0.18\linewidth]{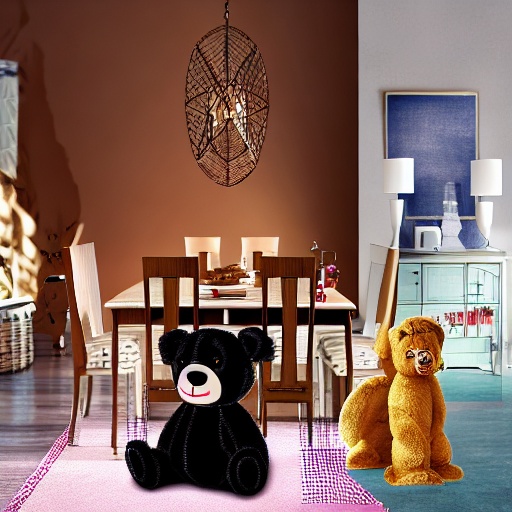}
  \includegraphics[width=0.18\linewidth]{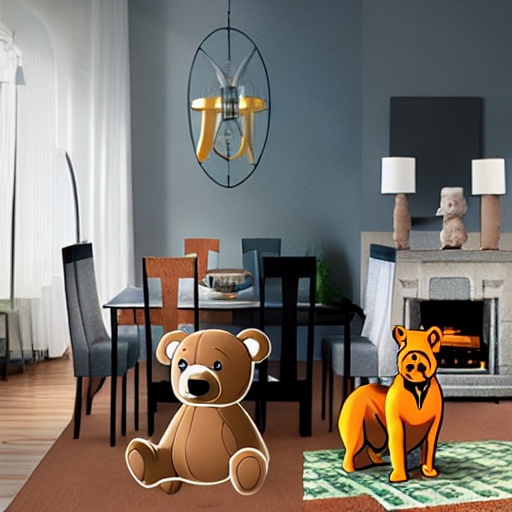}
  \includegraphics[width=0.18\linewidth]{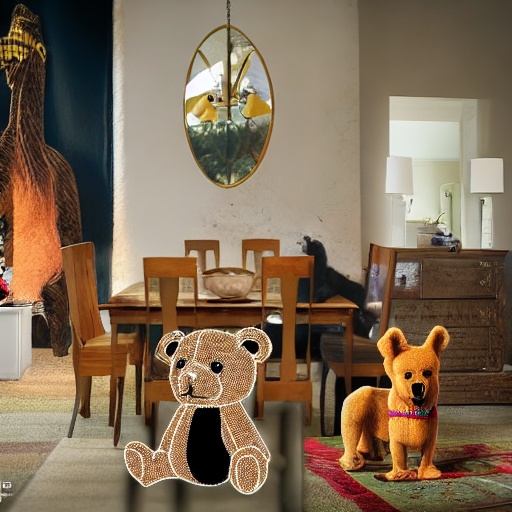}
  \includegraphics[width=0.18\linewidth]{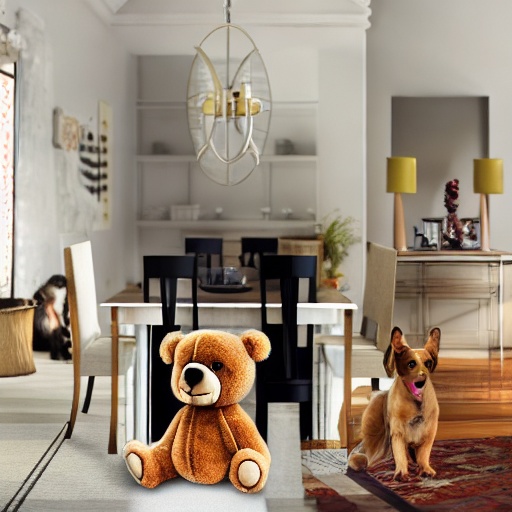}
  \includegraphics[width=0.18\linewidth]{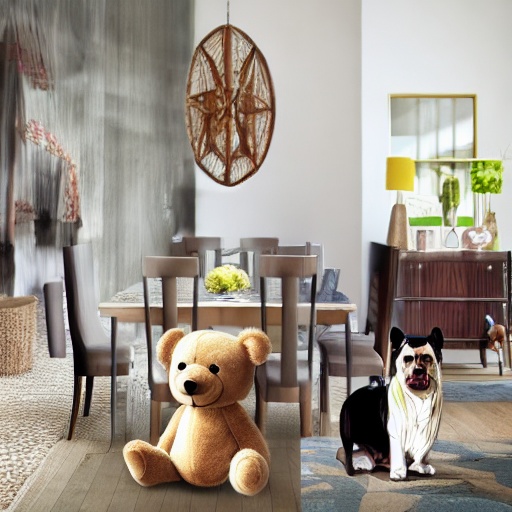}
\tabularnewline
  \includegraphics[width=0.18\linewidth]{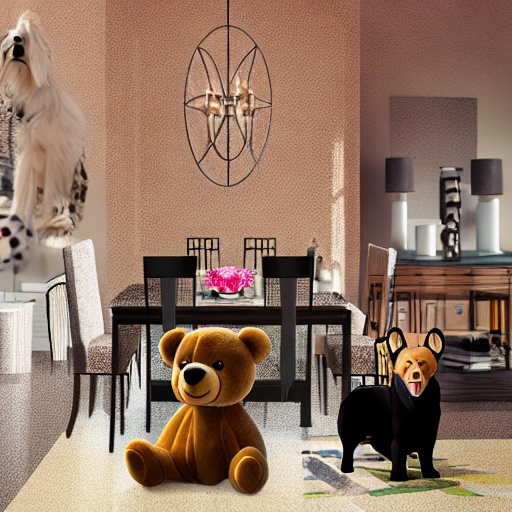}
  \includegraphics[width=0.18\linewidth]{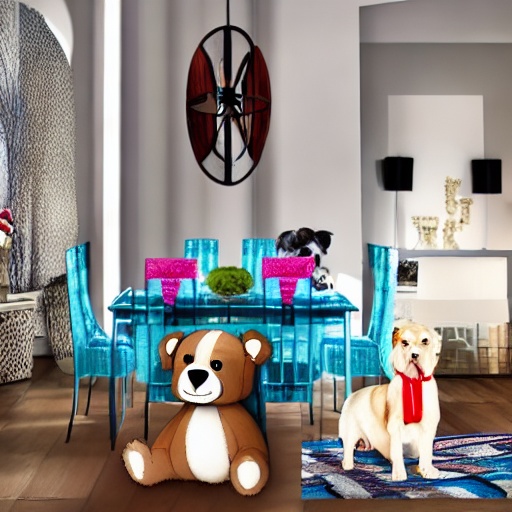}
  \includegraphics[width=0.18\linewidth]{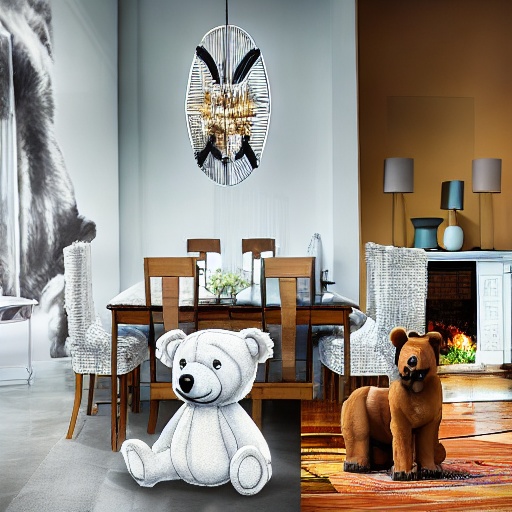}
  \includegraphics[width=0.18\linewidth]{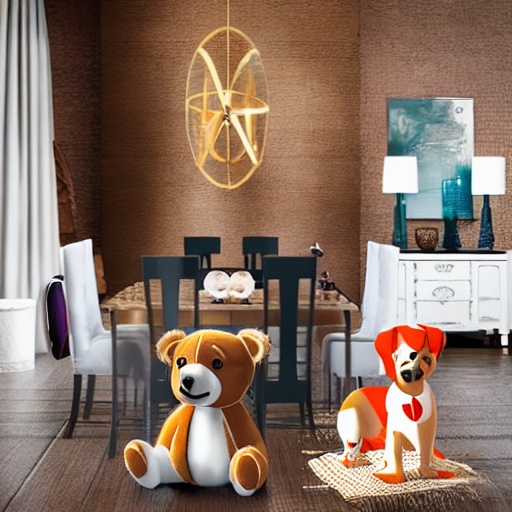}
  \includegraphics[width=0.18\linewidth]{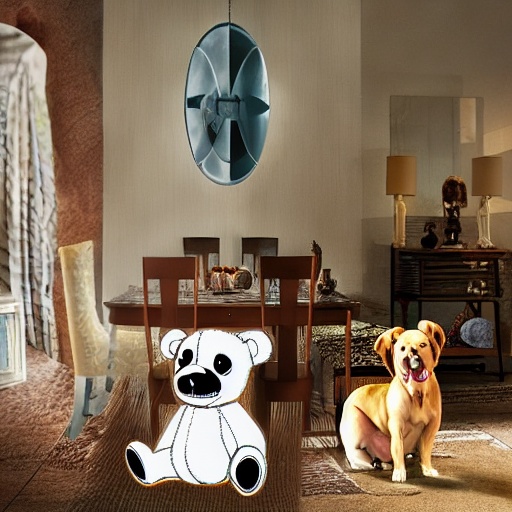}
\tabularnewline
  \includegraphics[width=0.18\linewidth]{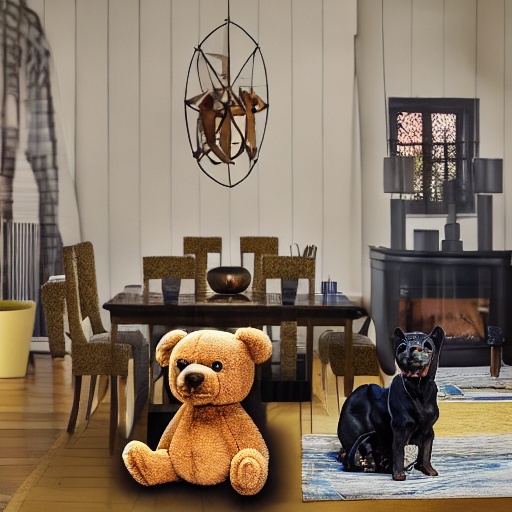}
  \includegraphics[width=0.18\linewidth]{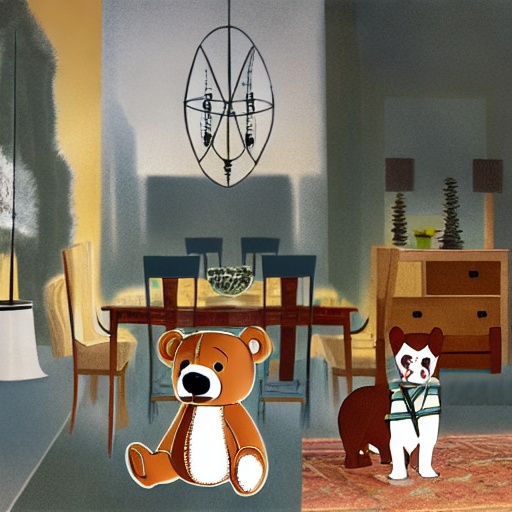}
   \includegraphics[width=0.18\linewidth]{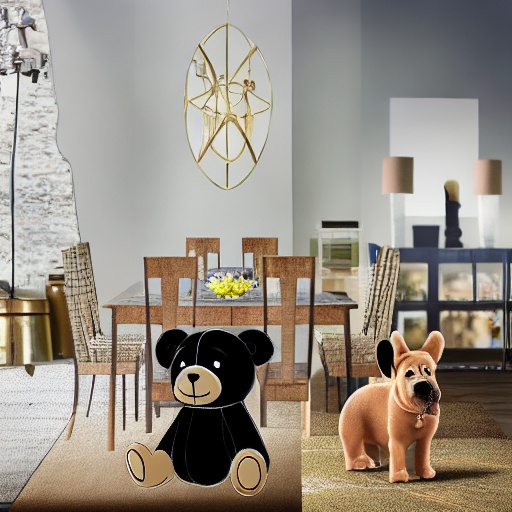}
  \includegraphics[width=0.18\linewidth]{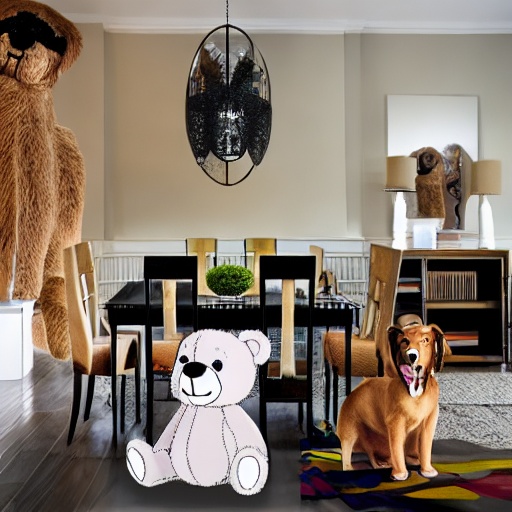}
  \includegraphics[width=0.18\linewidth]{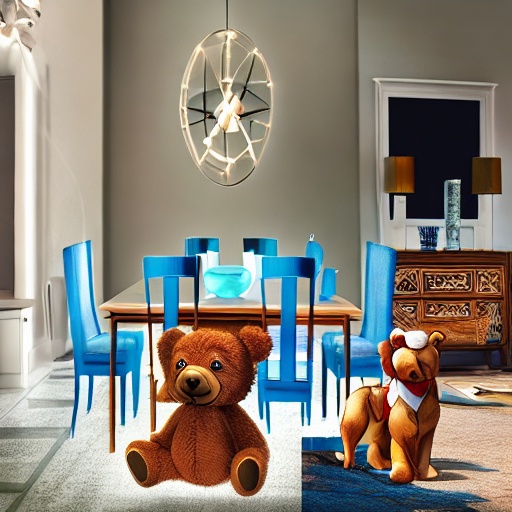}
  \tabularnewline
\vspace{2mm}
\vspace{-2\baselineskip}
\end{tabular}}
\vspace{-0.8cm}
\hspace{20pt}\captionof{figure}{\textbf{Non cherry picked examples for multimodal generation with 3 modalities. $\{Depth,HED,Canny\}$} Text Prompt is "Background: A living room, Foreground: a dog near a teddy bear"}
\label{fig:supp1}
\vspace{-2mm}
\end{figure*}%
\begin{figure*}[tb!]
    \centering
    \setlength{\tabcolsep}{0.5pt}
    {\small
    \renewcommand{\arraystretch}{0.5} 
    \begin{tabular}{c c c c c c c c c c}
    \captionsetup{type=figure, font=scriptsize}
\tabularnewline
  \includegraphics[width=0.18\linewidth,height=0.18\linewidth]{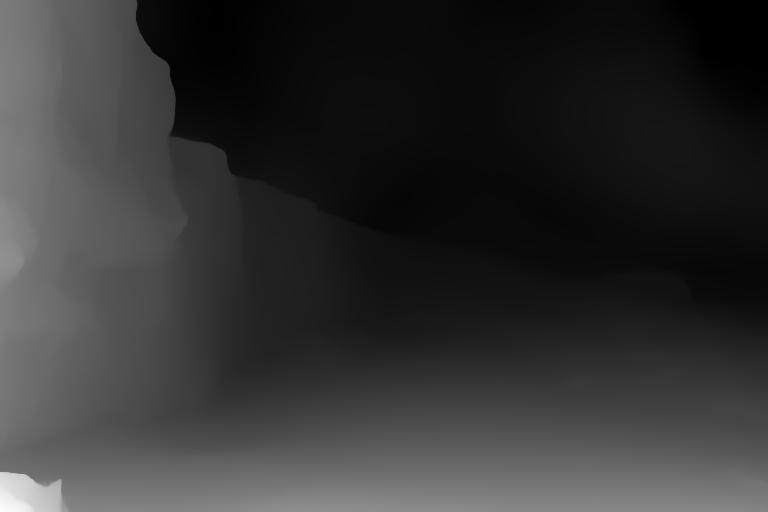}
   \includegraphics[width=0.18\linewidth,height=0.18\linewidth]{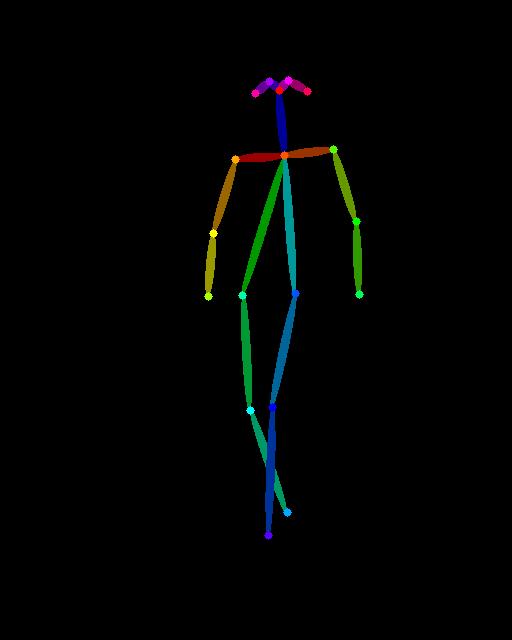}
  \includegraphics[width=0.18\linewidth,height=0.18\linewidth]{3modalitiesjpeg/dogbearroom.jpg}
   \tabularnewline

  \includegraphics[width=0.18\linewidth]{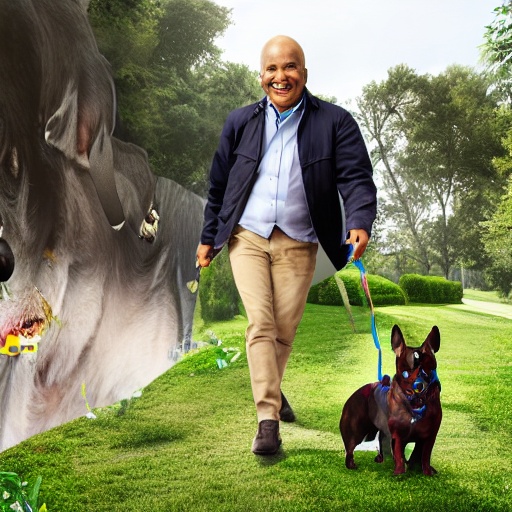}
  \includegraphics[width=0.18\linewidth]{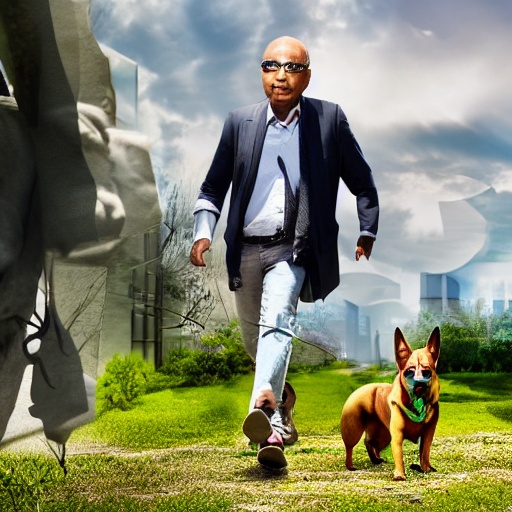}
  \includegraphics[width=0.18\linewidth]{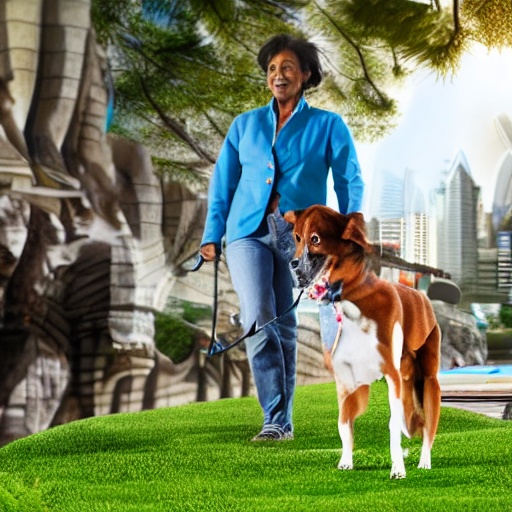}
  \includegraphics[width=0.18\linewidth]{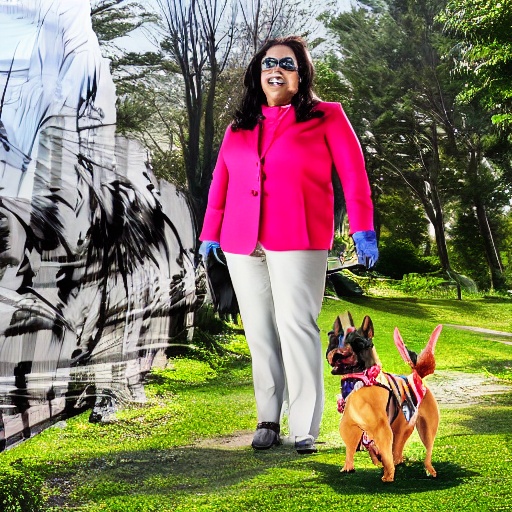}
  \includegraphics[width=0.18\linewidth]{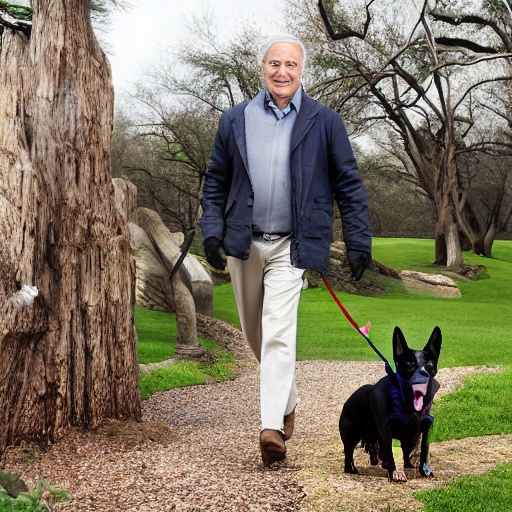}
 \tabularnewline
  \includegraphics[width=0.18\linewidth]{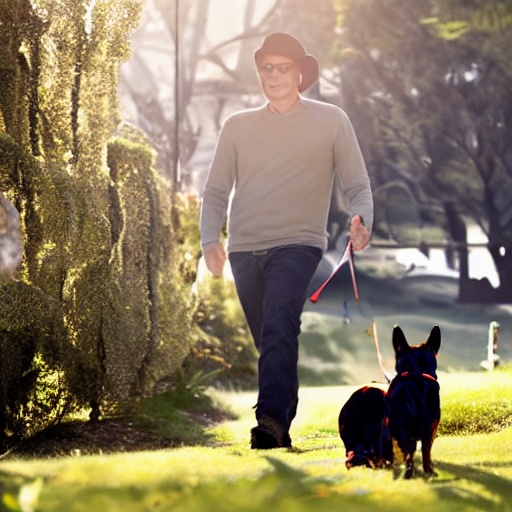}
  \includegraphics[width=0.18\linewidth]{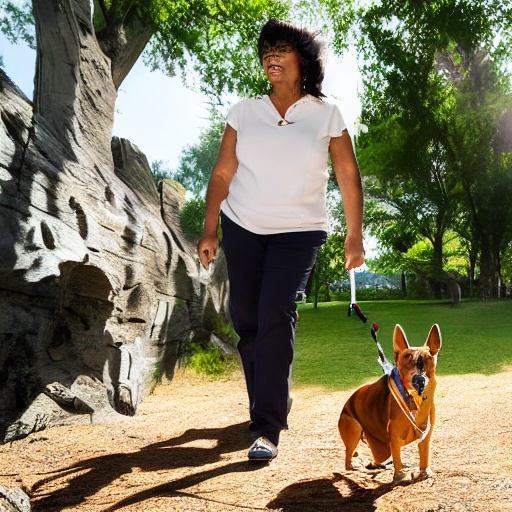}
  \includegraphics[width=0.18\linewidth]{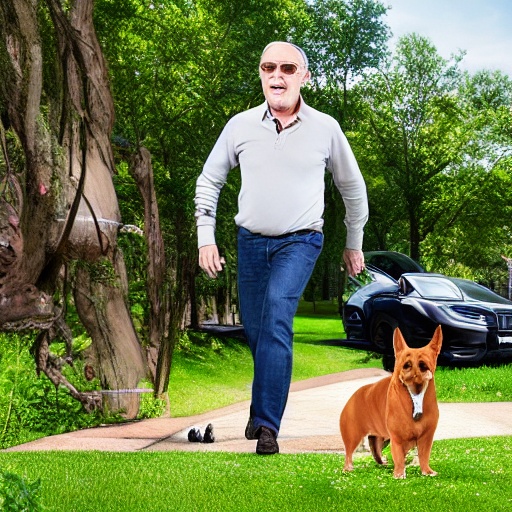}
  \includegraphics[width=0.18\linewidth]{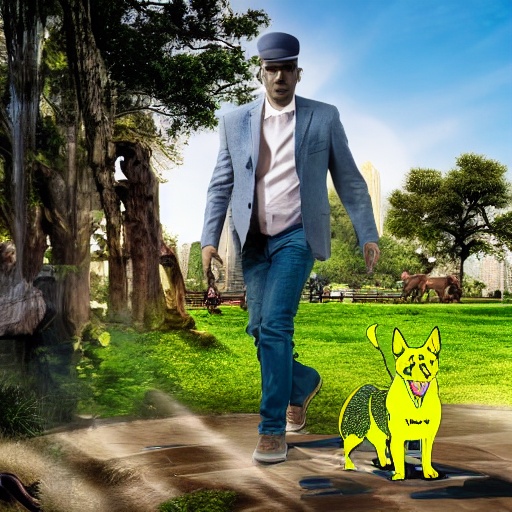}
  \includegraphics[width=0.18\linewidth]{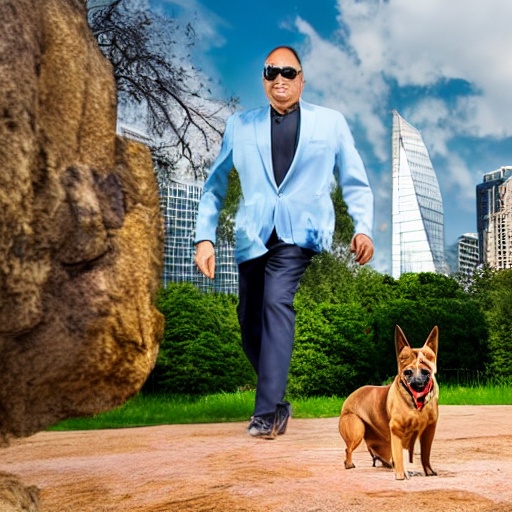}
\tabularnewline
  \includegraphics[width=0.18\linewidth]{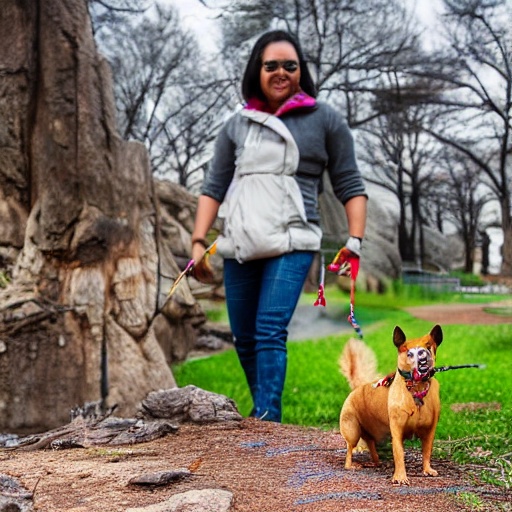}
  \includegraphics[width=0.18\linewidth]{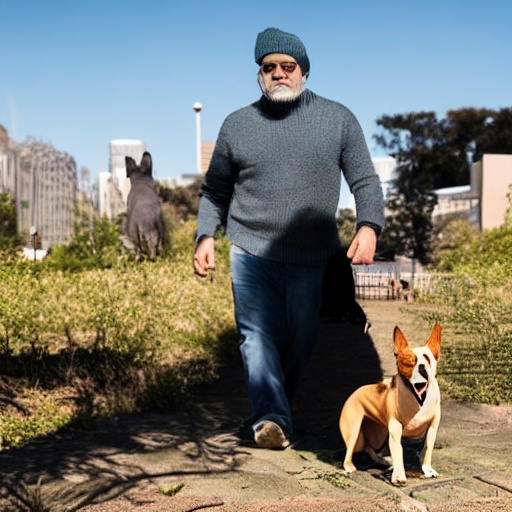}
  \includegraphics[width=0.18\linewidth]{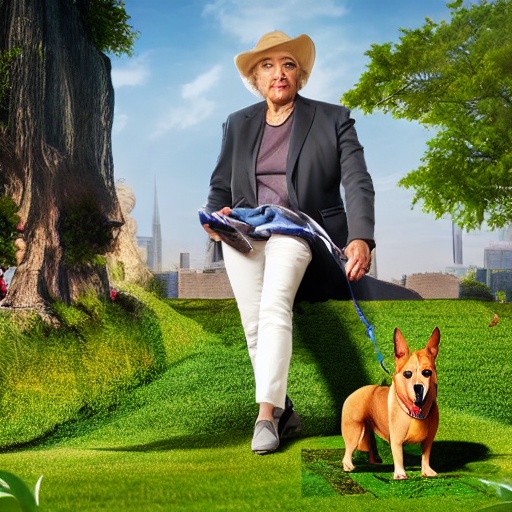}
  \includegraphics[width=0.18\linewidth]{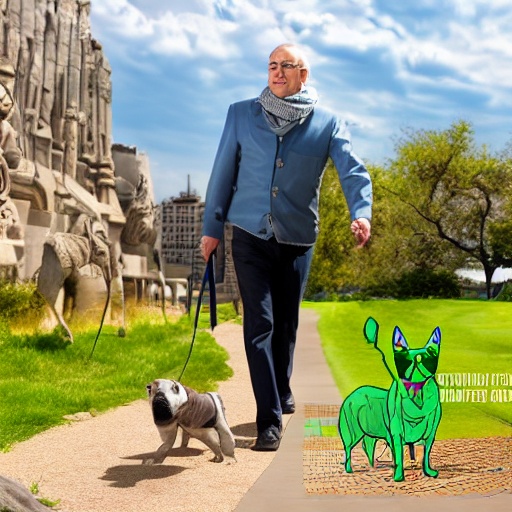}
  \includegraphics[width=0.18\linewidth]{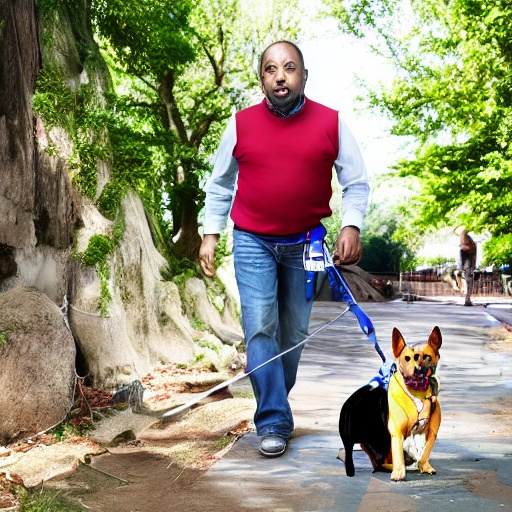}
\tabularnewline
  \includegraphics[width=0.18\linewidth]{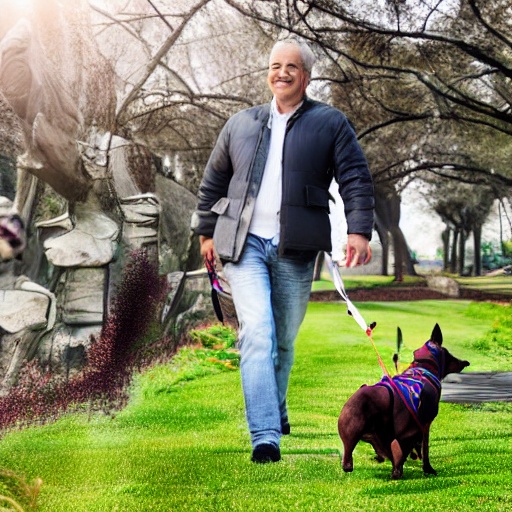}
  \includegraphics[width=0.18\linewidth]{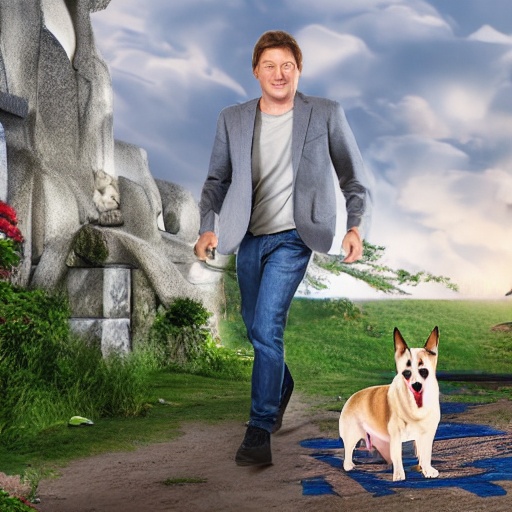}
  \includegraphics[width=0.18\linewidth]{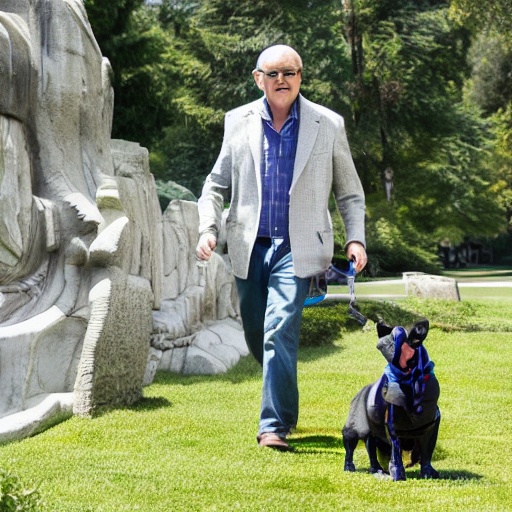}
  \includegraphics[width=0.18\linewidth]{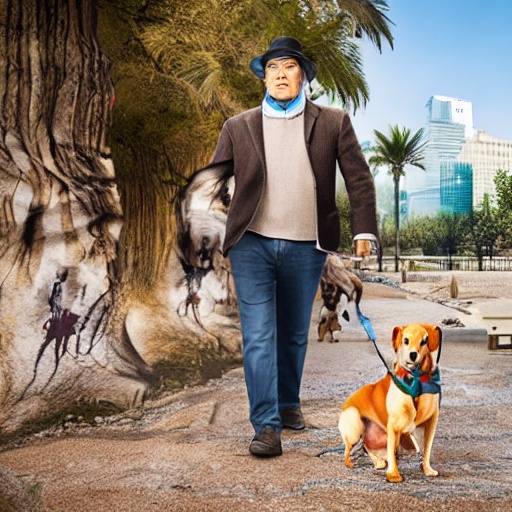}
  \includegraphics[width=0.18\linewidth]{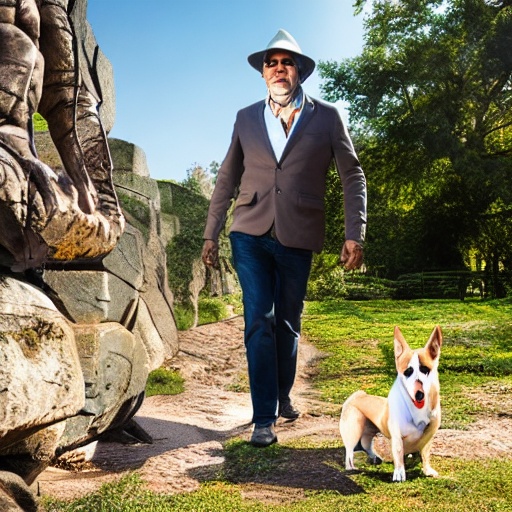}
\tabularnewline
  \includegraphics[width=0.18\linewidth]{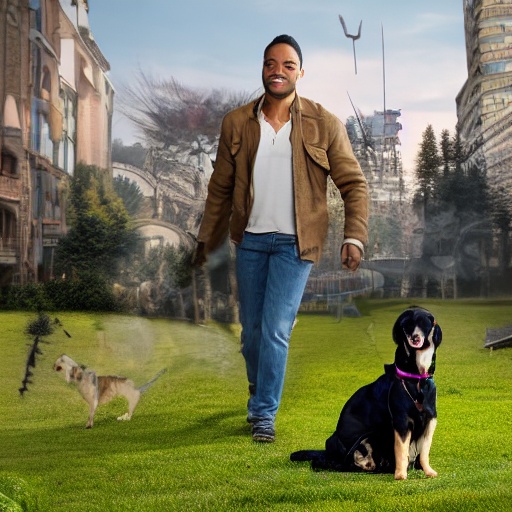}
  \includegraphics[width=0.18\linewidth]{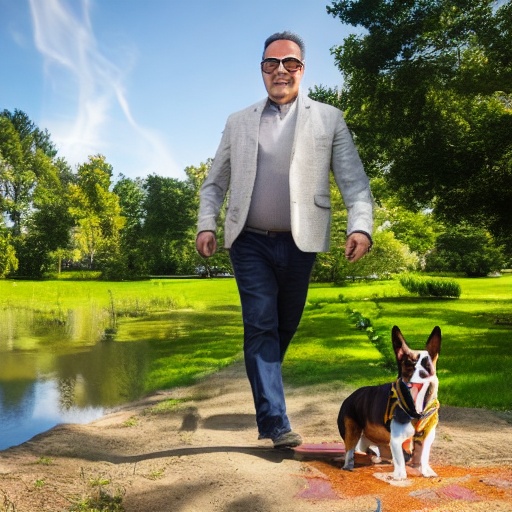}
   \includegraphics[width=0.18\linewidth]{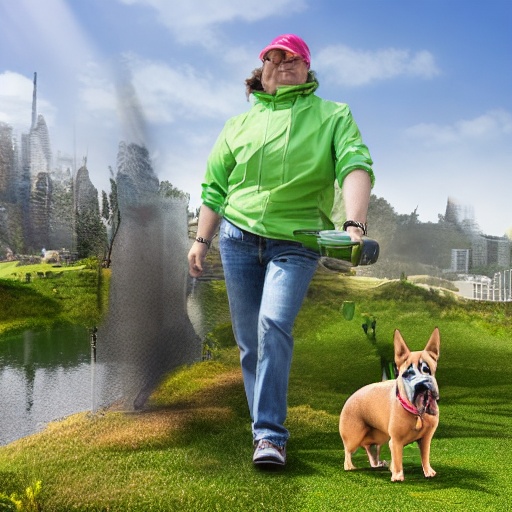}
  \includegraphics[width=0.18\linewidth]{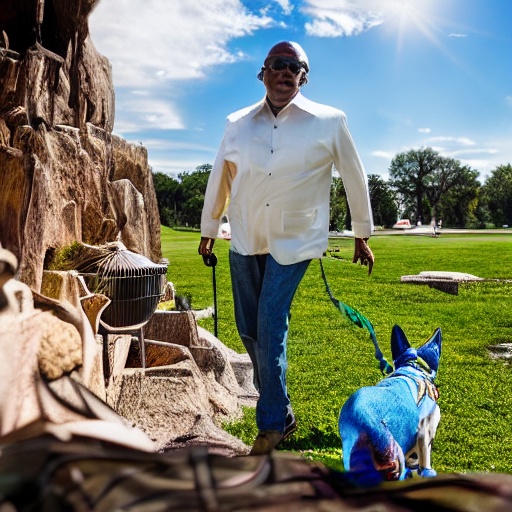}
  \includegraphics[width=0.18\linewidth]{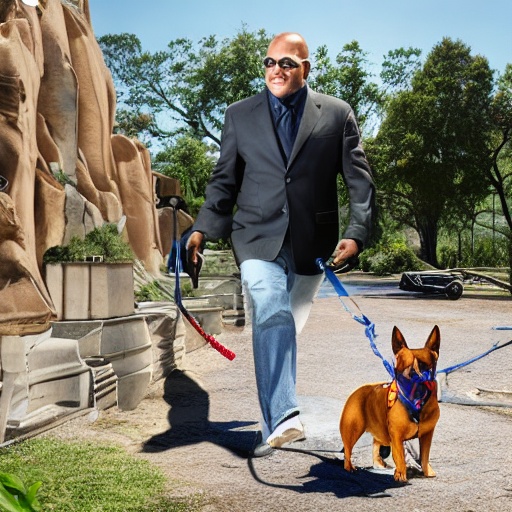}
  \tabularnewline
\vspace{2mm}
\vspace{-2\baselineskip}
\end{tabular}}
\vspace{-0.8cm}
\hspace{20pt}\captionof{figure}{\textbf{Non cherry picked examples for multimodal generation with 3 modalities. $\{Depth,HED,Pose\}$} Text Prompt is "Background: A park, Foreground: a dog near a person"}
\label{fig:supp2}
\vspace{-2mm}
\end{figure*}%
\begin{figure*}[tb!]
    \centering
    \setlength{\tabcolsep}{0.5pt}
    {\small
    \renewcommand{\arraystretch}{0.5} 
    \begin{tabular}{c c c c c c c c c c}
    \captionsetup{type=figure, font=scriptsize}
\tabularnewline
  \includegraphics[width=0.18\linewidth,height=0.18\linewidth]{3modalitiesjpeg/depth_mountain.jpg}
   \includegraphics[width=0.18\linewidth,height=0.18\linewidth]{3modalitiesjpeg/posefull.jpg}
  \includegraphics[width=0.18\linewidth,height=0.18\linewidth]{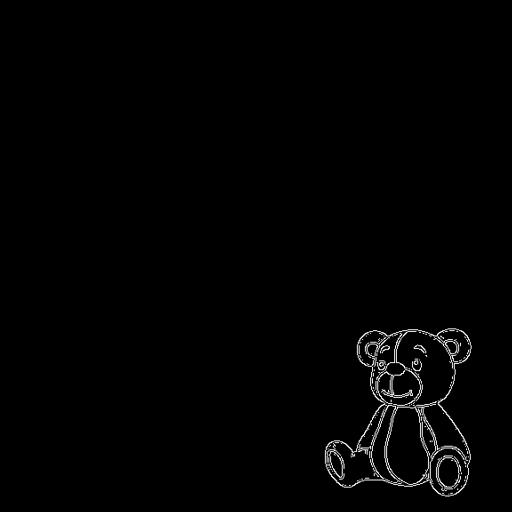}
   \tabularnewline

  \includegraphics[width=0.18\linewidth]{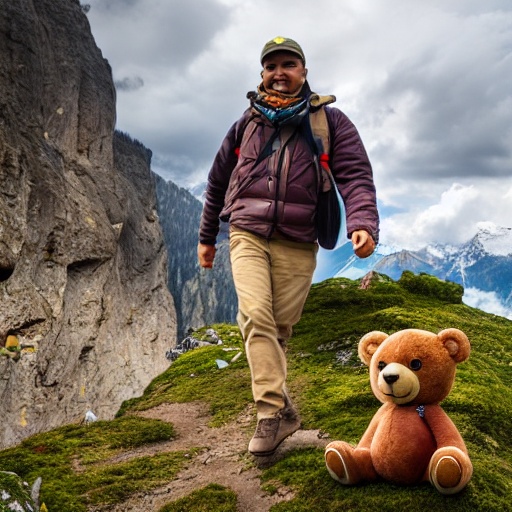}
  \includegraphics[width=0.18\linewidth]{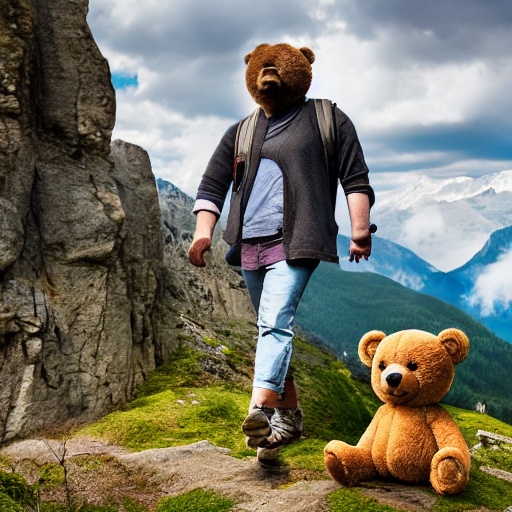}
  \includegraphics[width=0.18\linewidth]{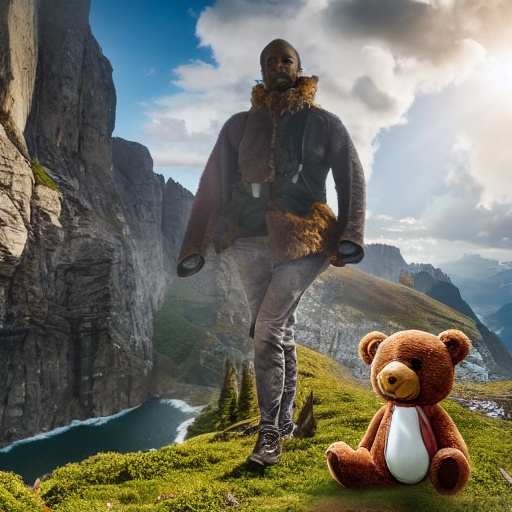}
  \includegraphics[width=0.18\linewidth]{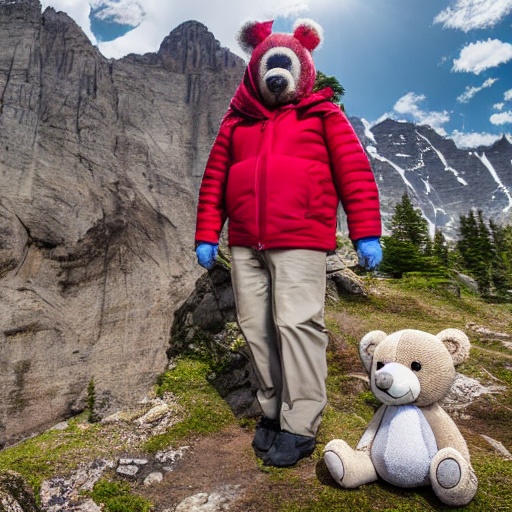}
  \includegraphics[width=0.18\linewidth]{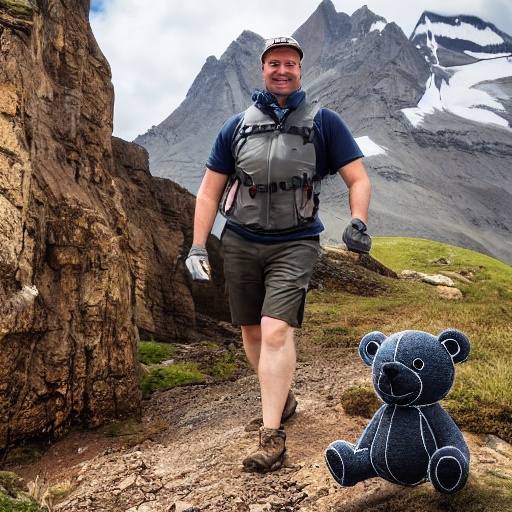}
 \tabularnewline
  \includegraphics[width=0.18\linewidth]{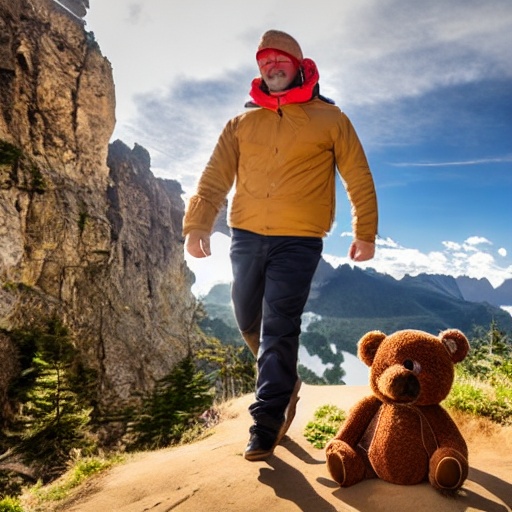}
  \includegraphics[width=0.18\linewidth]{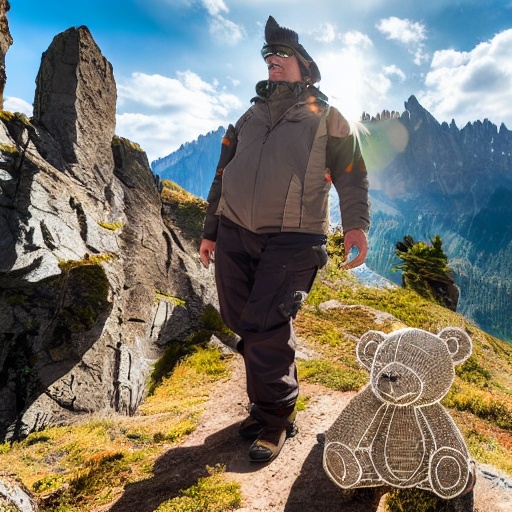}
  \includegraphics[width=0.18\linewidth]{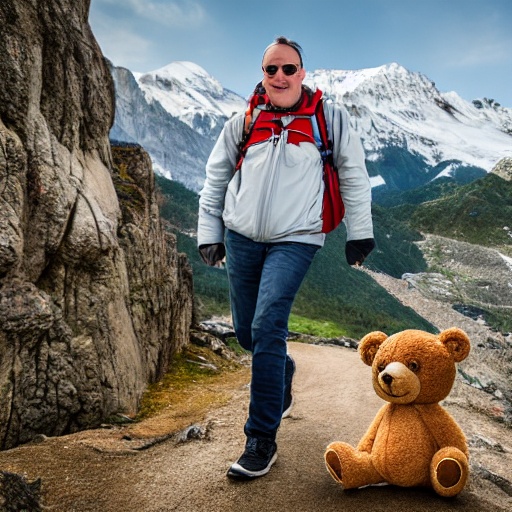}
  \includegraphics[width=0.18\linewidth]{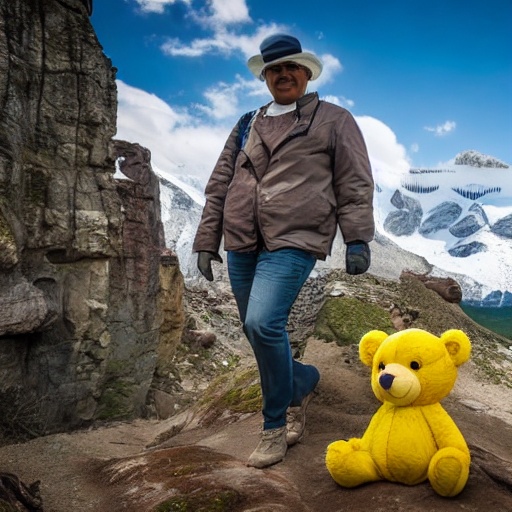}
  \includegraphics[width=0.18\linewidth]{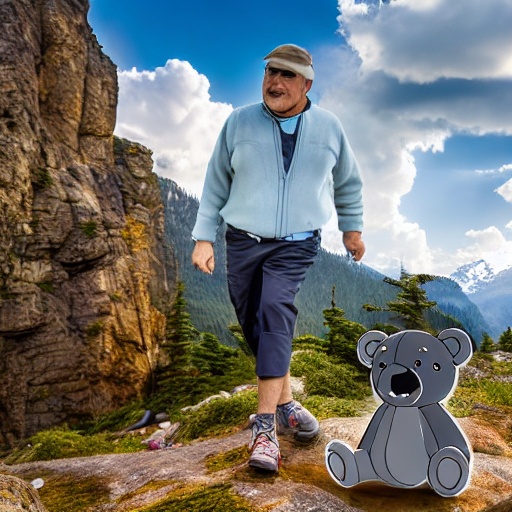}
\tabularnewline
  \includegraphics[width=0.18\linewidth]{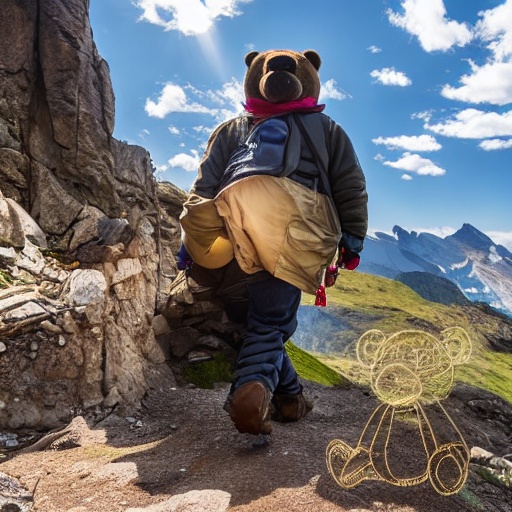}
  \includegraphics[width=0.18\linewidth]{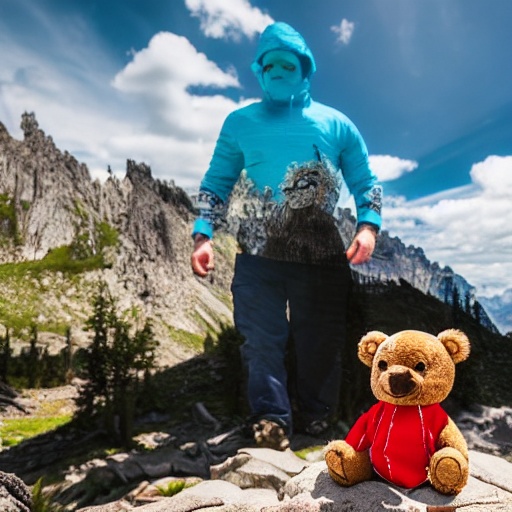}
  \includegraphics[width=0.18\linewidth]{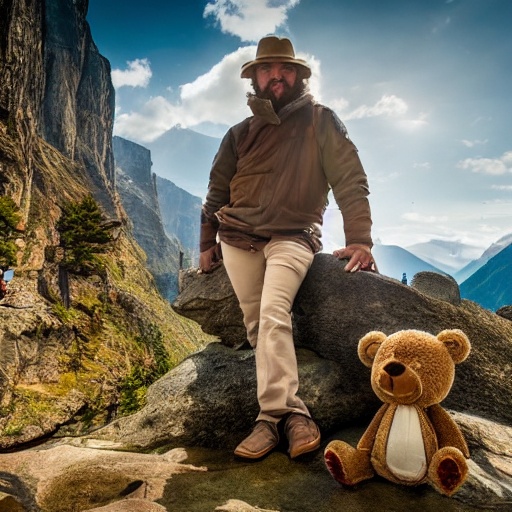}
  \includegraphics[width=0.18\linewidth]{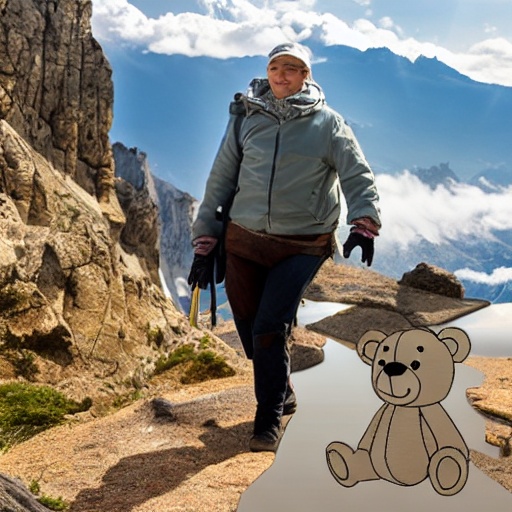}
  \includegraphics[width=0.18\linewidth]{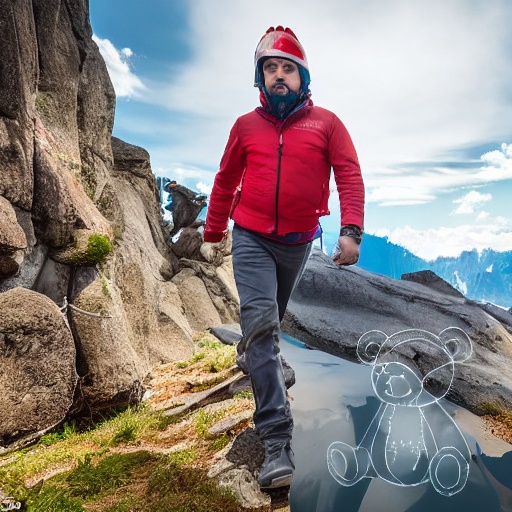}
\tabularnewline
  \includegraphics[width=0.18\linewidth]{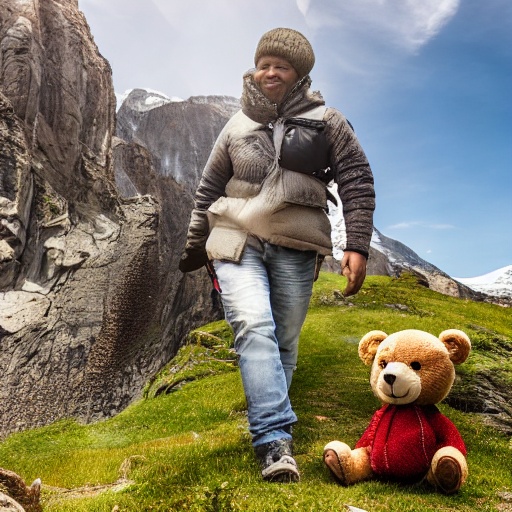}
  \includegraphics[width=0.18\linewidth]{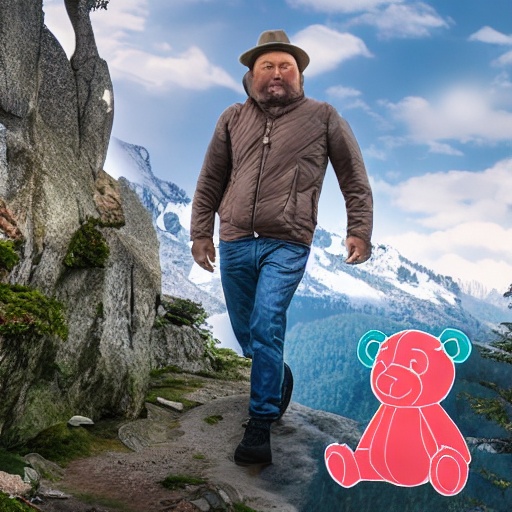}
  \includegraphics[width=0.18\linewidth]{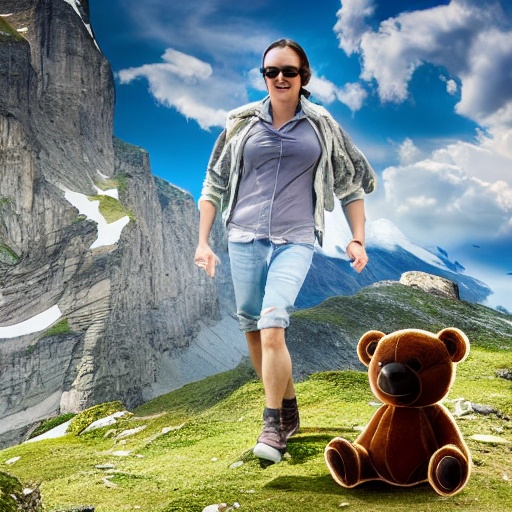}
  \includegraphics[width=0.18\linewidth]{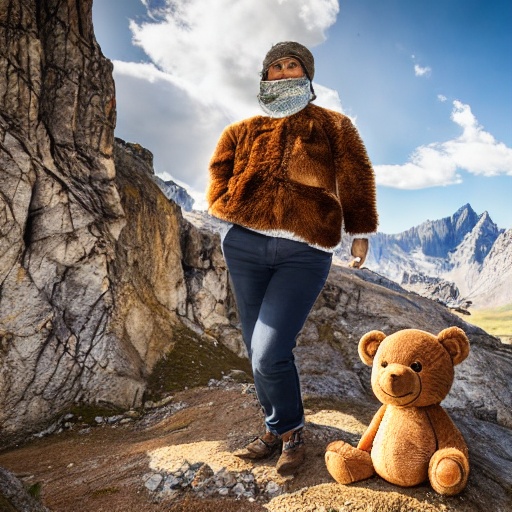}
  \includegraphics[width=0.18\linewidth]{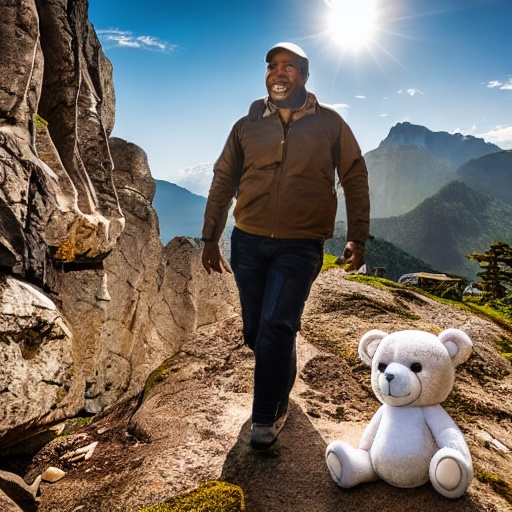}
\tabularnewline
  \includegraphics[width=0.18\linewidth]{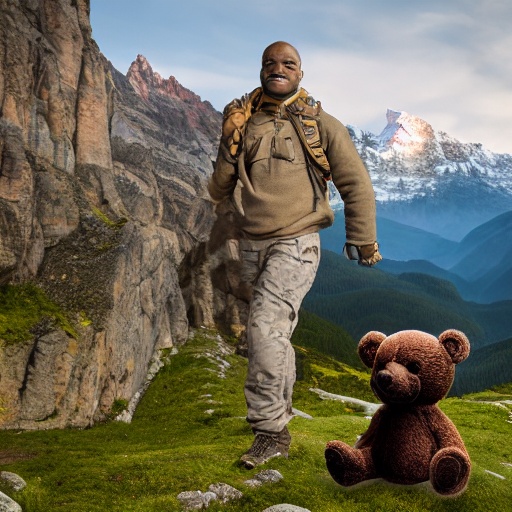}
  \includegraphics[width=0.18\linewidth]{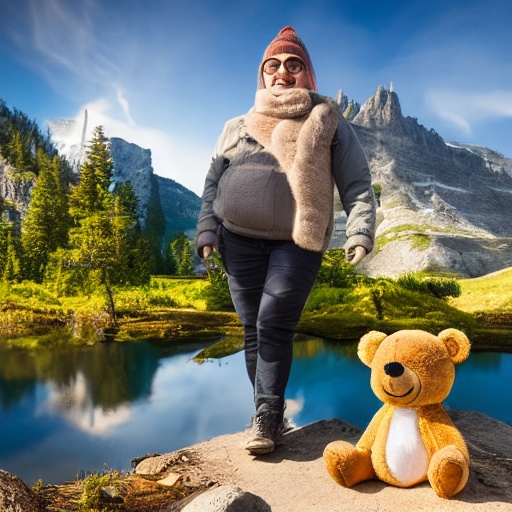}
   \includegraphics[width=0.18\linewidth]{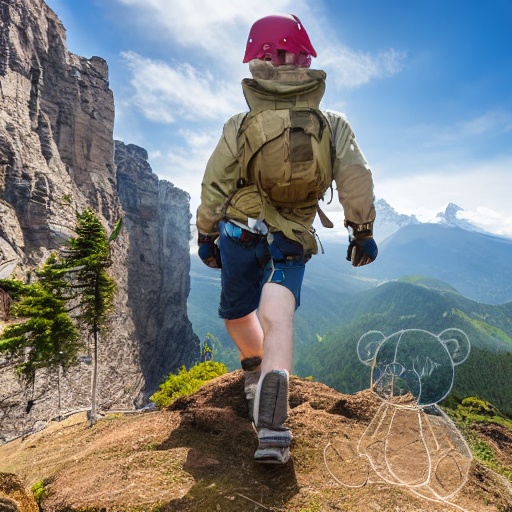}
  \includegraphics[width=0.18\linewidth]{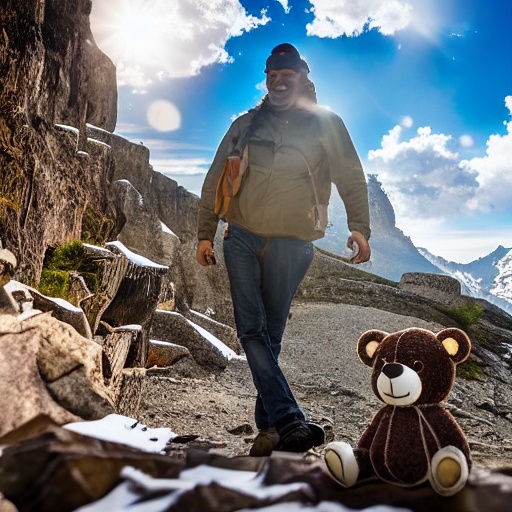}
  \includegraphics[width=0.18\linewidth]{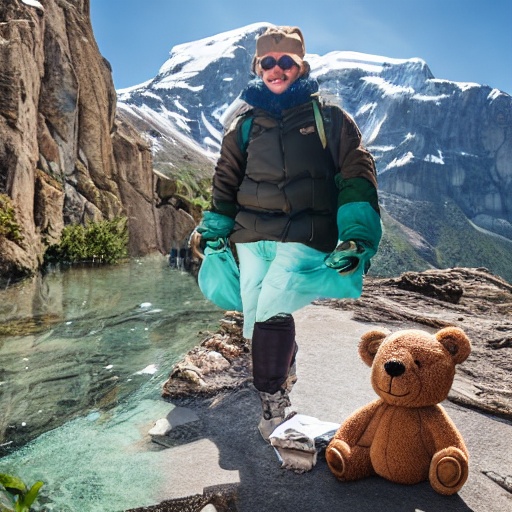}
  \tabularnewline
\vspace{2mm}
\vspace{-2\baselineskip}
\end{tabular}}
\vspace{-0.8cm}
\hspace{20pt}\captionof{figure}{\textbf{Non cherry picked examples for multimodal generation with 3 modalities. $\{HED,Depth,Pose\}$} Text Prompt is "Background: A mountain, Foreground: a teddy bear near a person"}
\label{fig:supp4}
\vspace{-2mm}
\end{figure*}%
\begin{figure*}[tb!]
    \centering
    \setlength{\tabcolsep}{0.5pt}
    {\small
    \renewcommand{\arraystretch}{0.5} 
    \begin{tabular}{c c c c c c c c c c}
    \captionsetup{type=figure, font=scriptsize}
\tabularnewline
  \includegraphics[width=0.18\linewidth,height=0.18\linewidth]{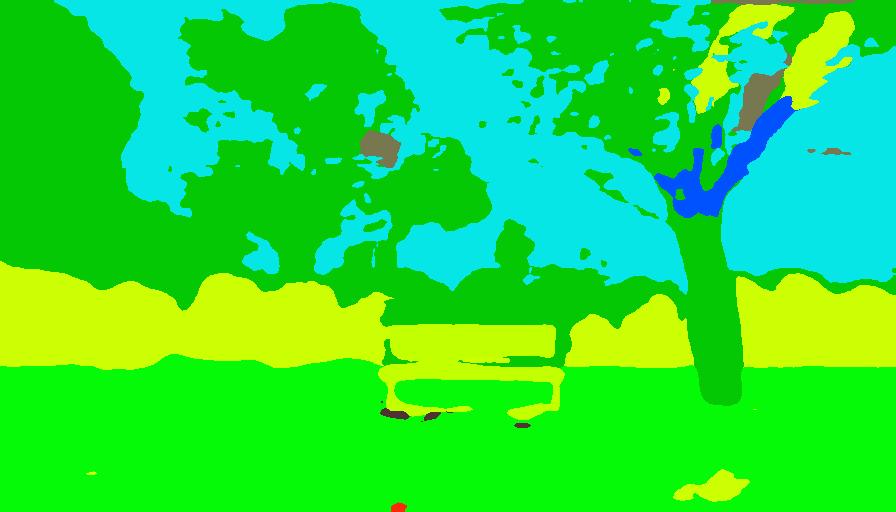}
   \includegraphics[width=0.18\linewidth,height=0.18\linewidth]{3modalitiesjpeg/posefull.jpg}
  \includegraphics[width=0.18\linewidth,height=0.18\linewidth]{3modalitiesjpeg/dogbearroom.jpg}
   \tabularnewline

  \includegraphics[width=0.18\linewidth]{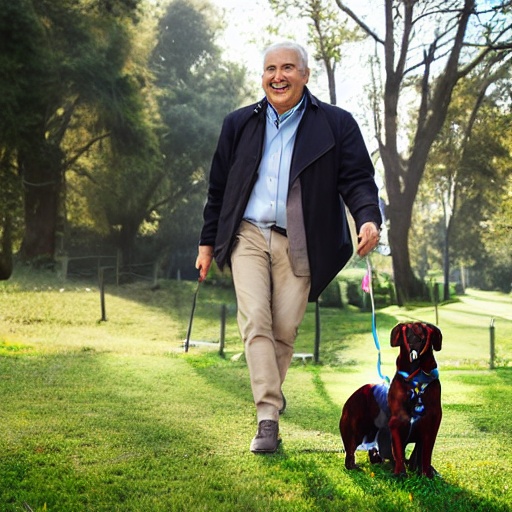}
  \includegraphics[width=0.18\linewidth]{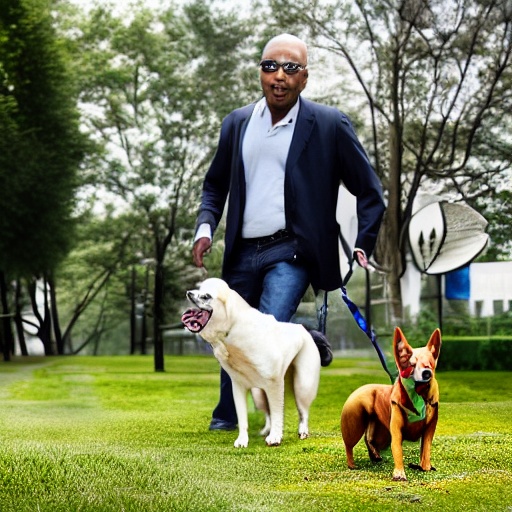}
  \includegraphics[width=0.18\linewidth]{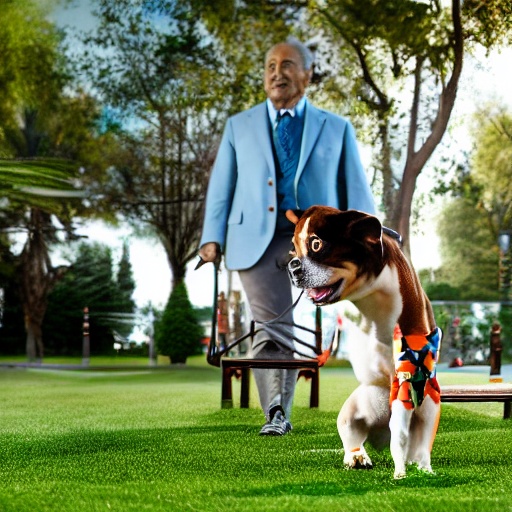}
  \includegraphics[width=0.18\linewidth]{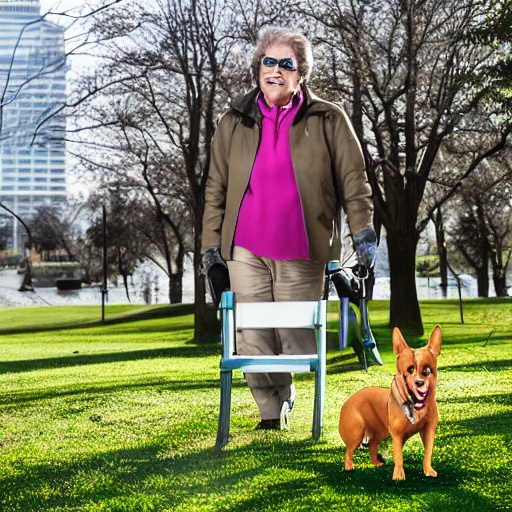}
  \includegraphics[width=0.18\linewidth]{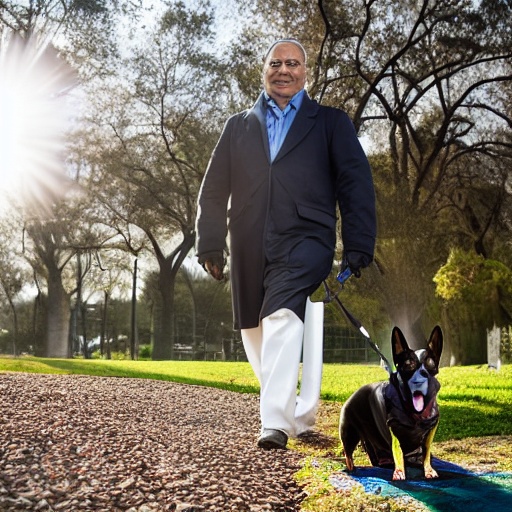}
 \tabularnewline
  \includegraphics[width=0.18\linewidth]{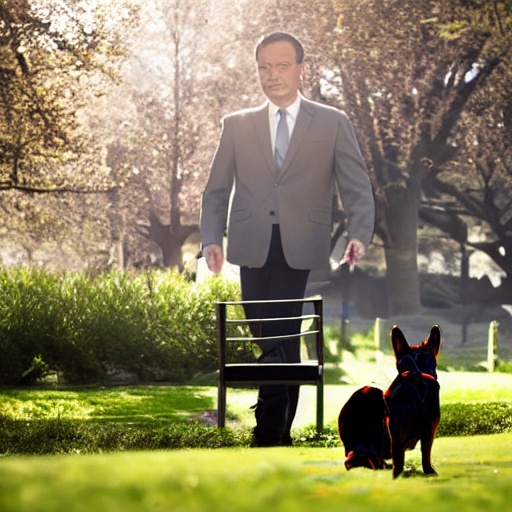}
  \includegraphics[width=0.18\linewidth]{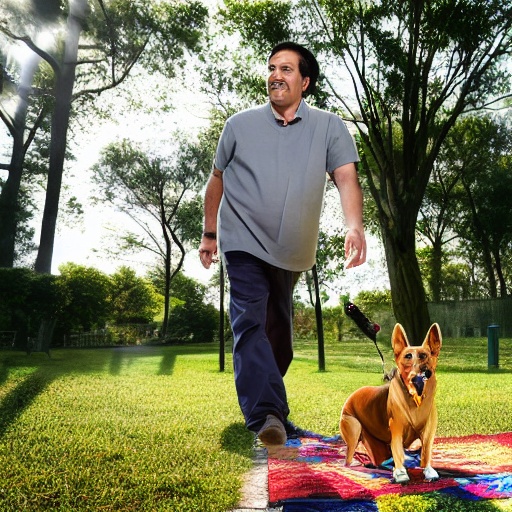}
  \includegraphics[width=0.18\linewidth]{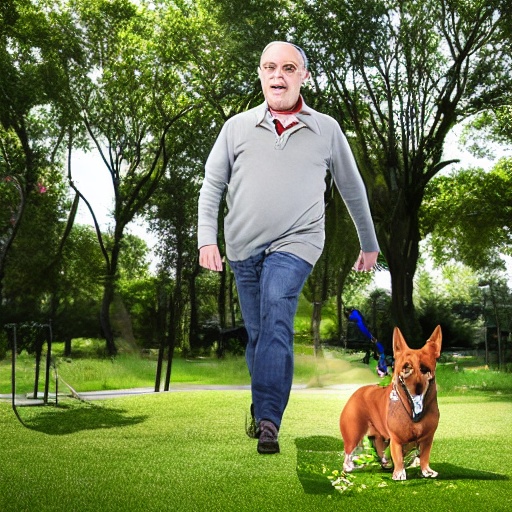}
  \includegraphics[width=0.18\linewidth]{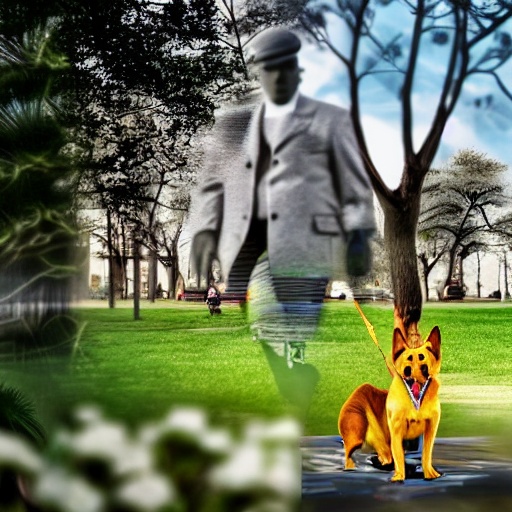}
  \includegraphics[width=0.18\linewidth]{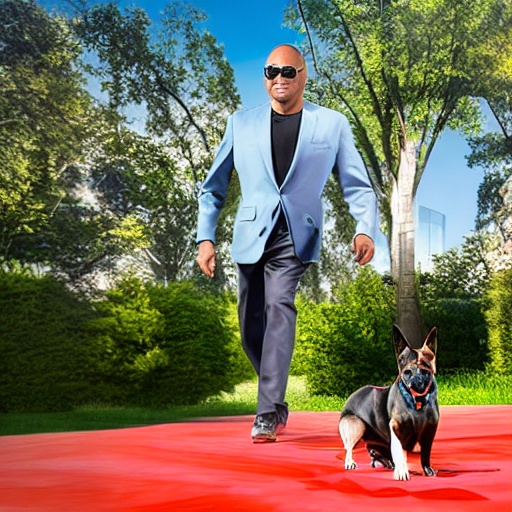}
\tabularnewline
  \includegraphics[width=0.18\linewidth]{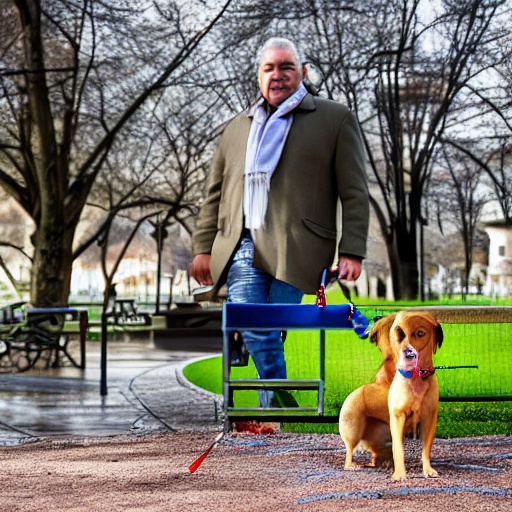}
  \includegraphics[width=0.18\linewidth]{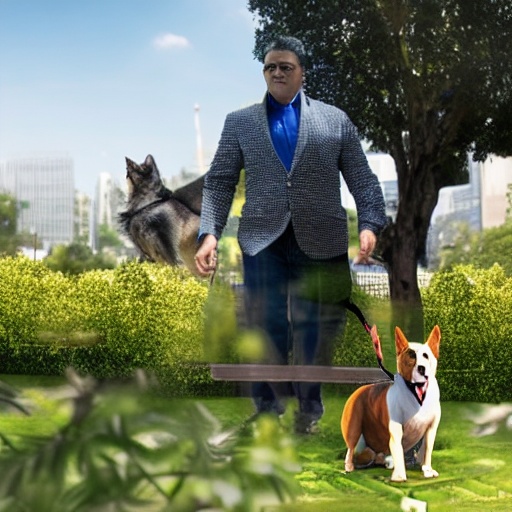}
  \includegraphics[width=0.18\linewidth]{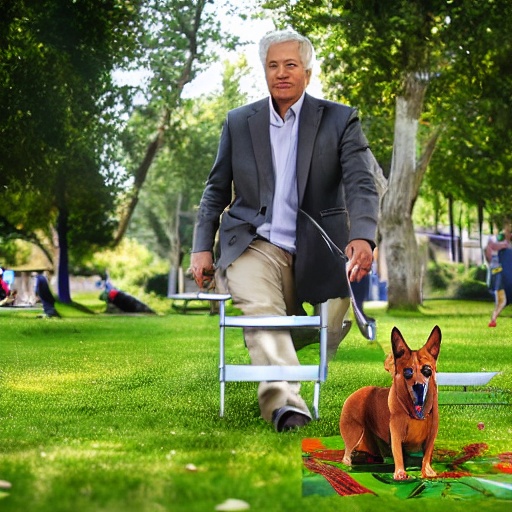}
  \includegraphics[width=0.18\linewidth]{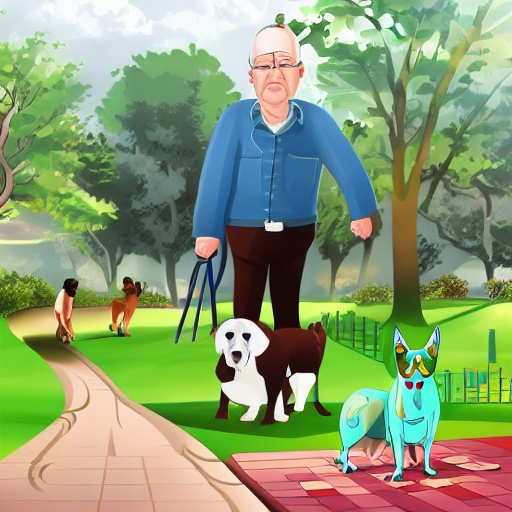}
  \includegraphics[width=0.18\linewidth]{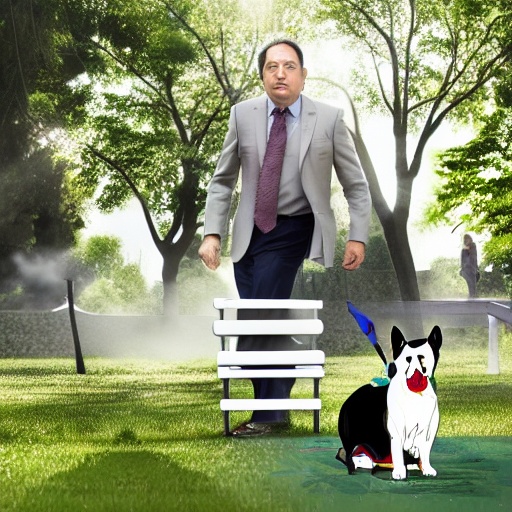}
\tabularnewline
  \includegraphics[width=0.18\linewidth]{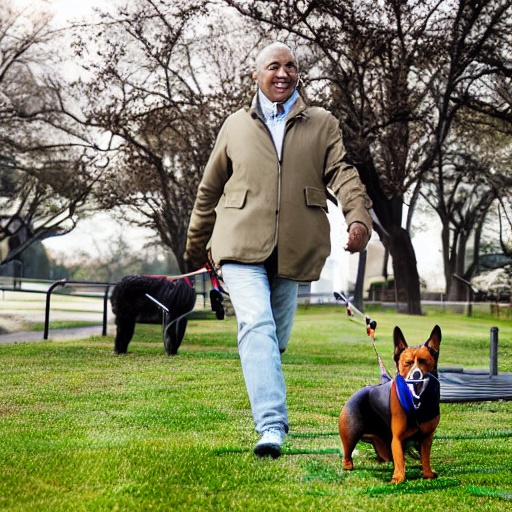}
  \includegraphics[width=0.18\linewidth]{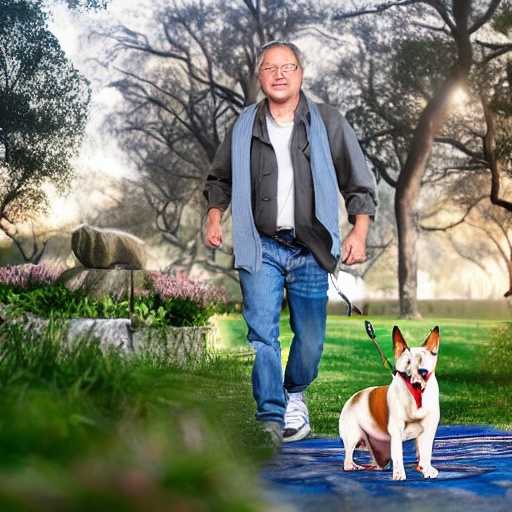}
  \includegraphics[width=0.18\linewidth]{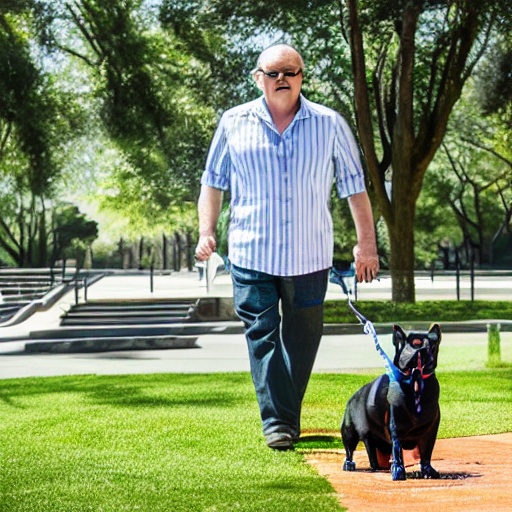}
  \includegraphics[width=0.18\linewidth]{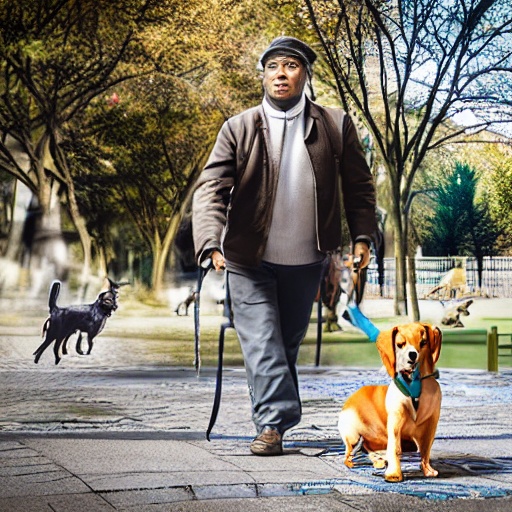}
  \includegraphics[width=0.18\linewidth]{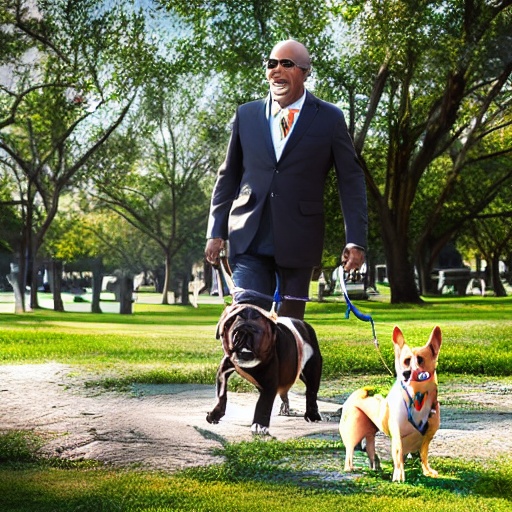}
\tabularnewline
  \includegraphics[width=0.18\linewidth]{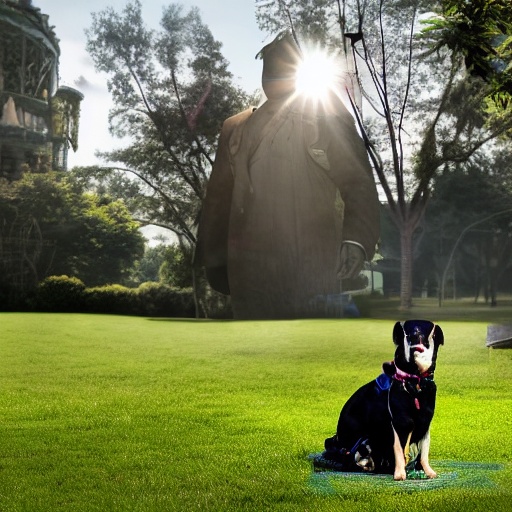}
  \includegraphics[width=0.18\linewidth]{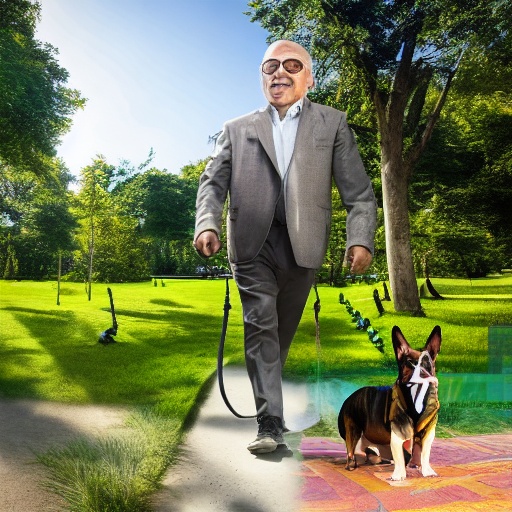}
   \includegraphics[width=0.18\linewidth]{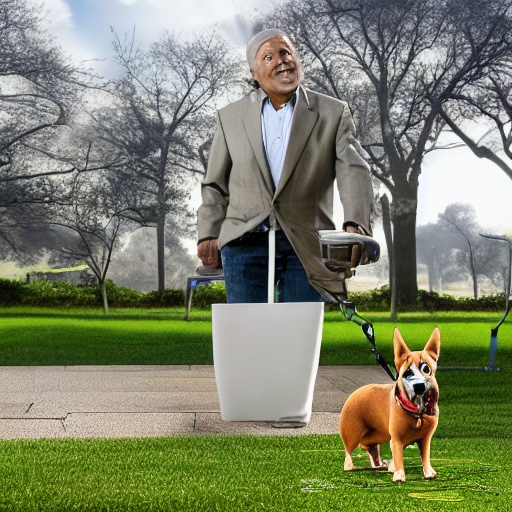}
  \includegraphics[width=0.18\linewidth]{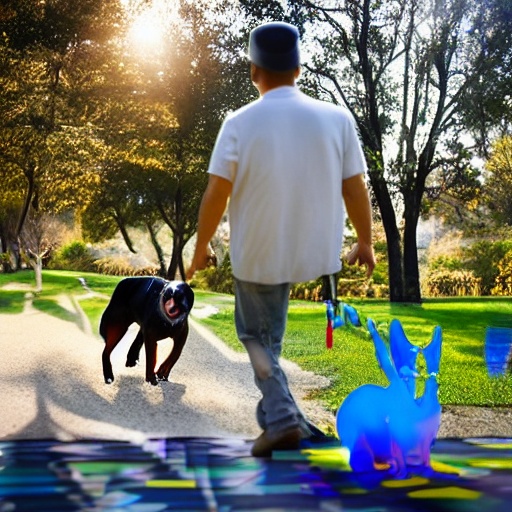}
  \includegraphics[width=0.18\linewidth]{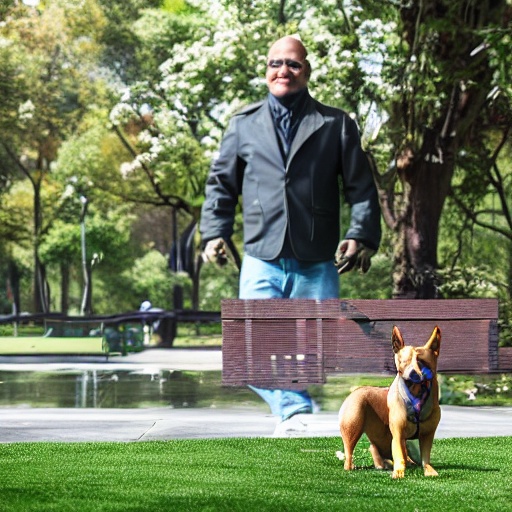}
  \tabularnewline
\vspace{2mm}
\vspace{-2\baselineskip}
\end{tabular}}
\vspace{-0.8cm}
\hspace{20pt}\captionof{figure}{\textbf{Non cherry picked examples for multimodal generation with 3 modalities. $\{SEG,HED,Pose\}$} Text Prompt is "Background: A park, Foreground: a dog near a person"}
\label{fig:supp3}
\vspace{-2mm}
\end{figure*}%
\begin{figure*}[tb!]
    \centering
    \setlength{\tabcolsep}{0.5pt}
    {\small
    \renewcommand{\arraystretch}{0.5} 
    \begin{tabular}{c c c c c c c c c c}
    \captionsetup{type=figure, font=scriptsize}
\tabularnewline
  \includegraphics[width=0.18\linewidth,height=0.18\linewidth]{3modalitiesjpeg/depth_mountain.jpg}
   \includegraphics[width=0.18\linewidth,height=0.18\linewidth]{3modalitiesjpeg/posefull.jpg}
  \includegraphics[width=0.18\linewidth,height=0.18\linewidth]{3modalitiesjpeg/bearend.jpg}
   \tabularnewline

  \includegraphics[width=0.18\linewidth]{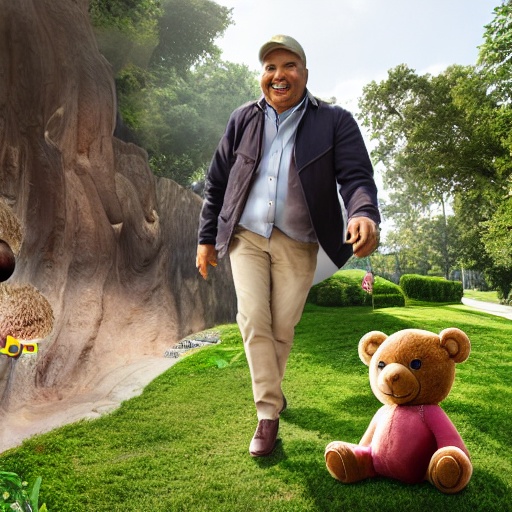}
  \includegraphics[width=0.18\linewidth]{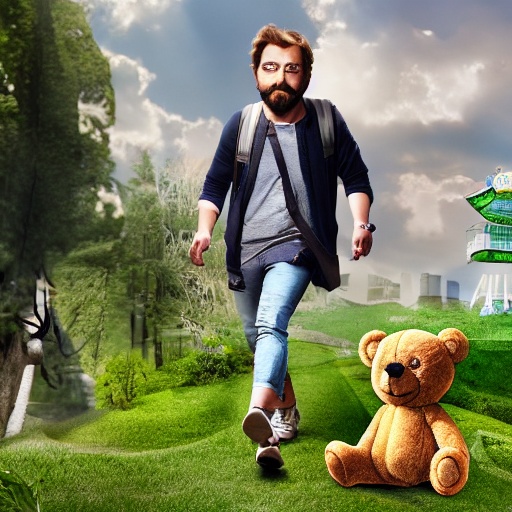}
  \includegraphics[width=0.18\linewidth]{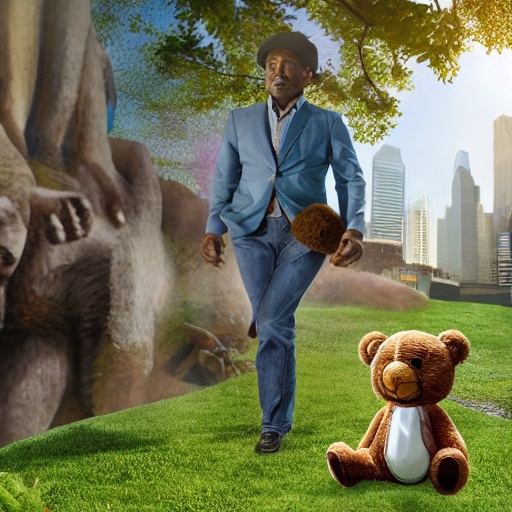}
  \includegraphics[width=0.18\linewidth]{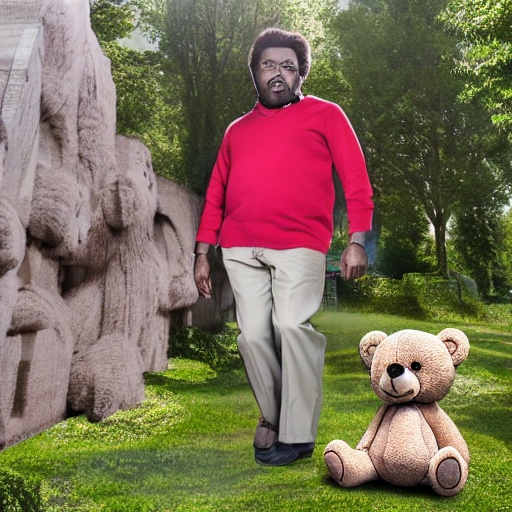}
  \includegraphics[width=0.18\linewidth]{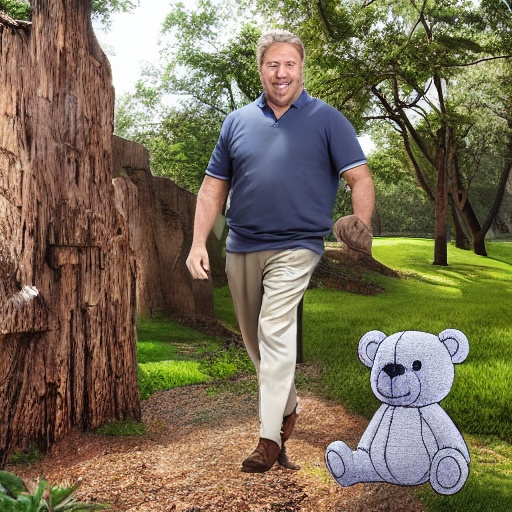}
 \tabularnewline
  \includegraphics[width=0.18\linewidth]{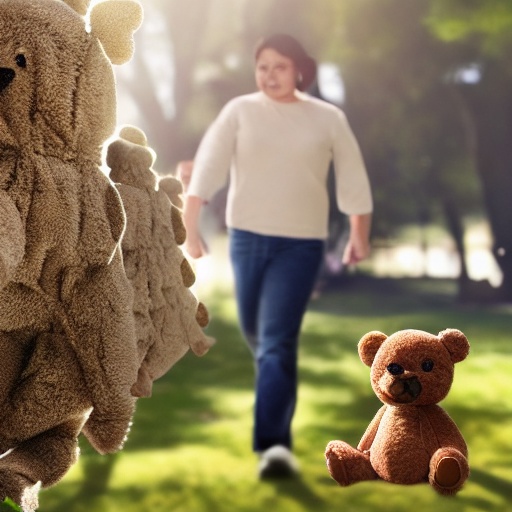}
  \includegraphics[width=0.18\linewidth]{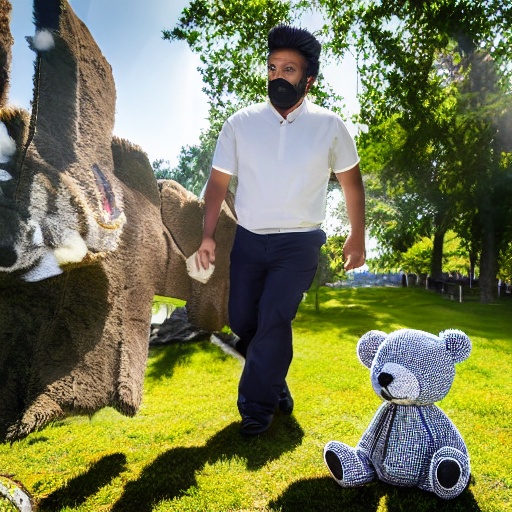}
  \includegraphics[width=0.18\linewidth]{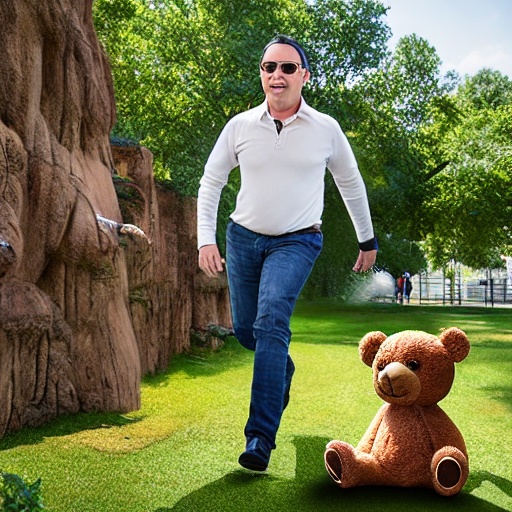}
  \includegraphics[width=0.18\linewidth]{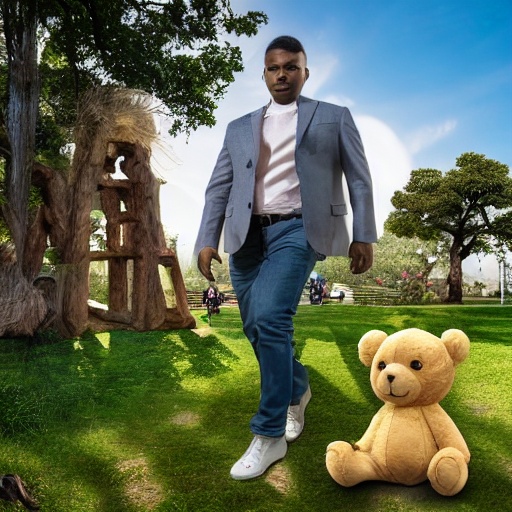}
  \includegraphics[width=0.18\linewidth]{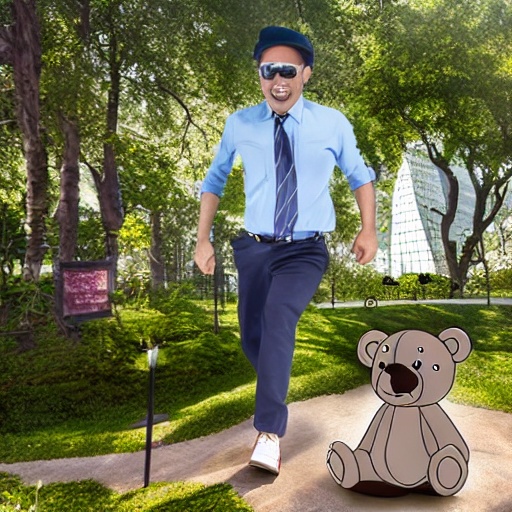}
\tabularnewline
  \includegraphics[width=0.18\linewidth]{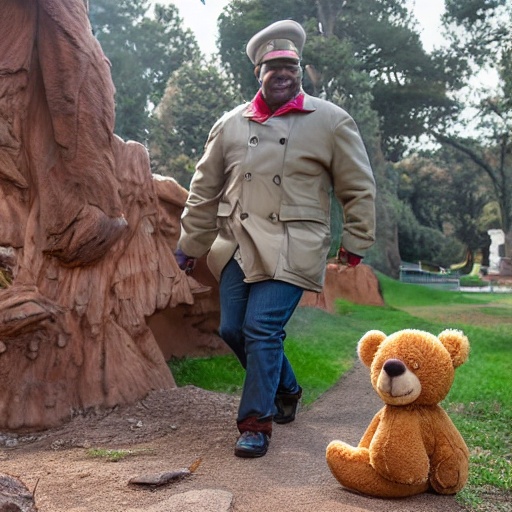}
  \includegraphics[width=0.18\linewidth]{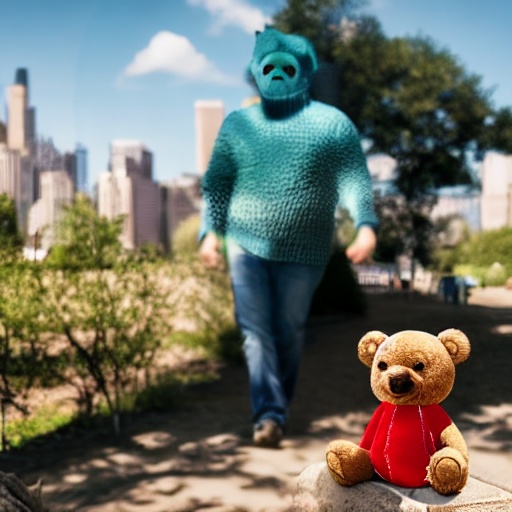}
  \includegraphics[width=0.18\linewidth]{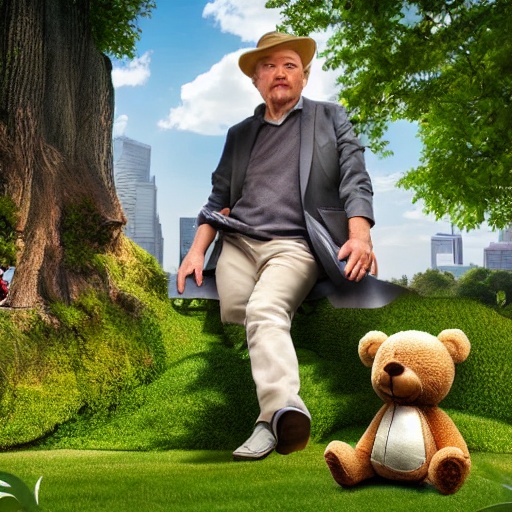}
  \includegraphics[width=0.18\linewidth]{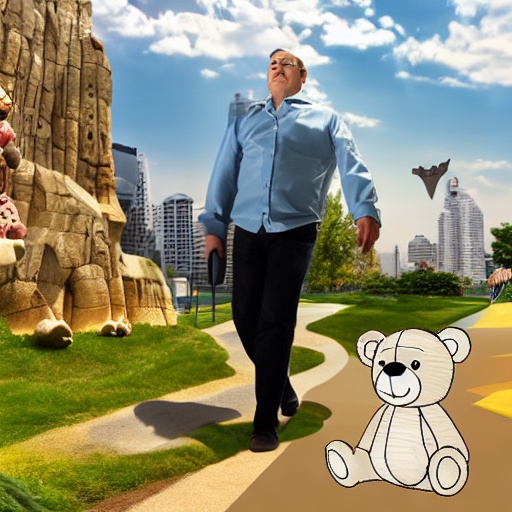}
  \includegraphics[width=0.18\linewidth]{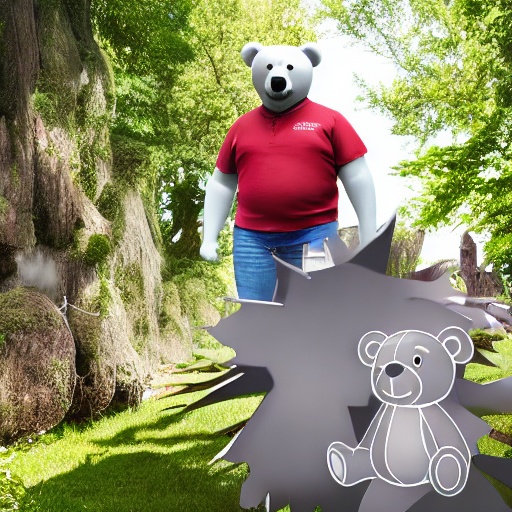}
\tabularnewline
  \includegraphics[width=0.18\linewidth]{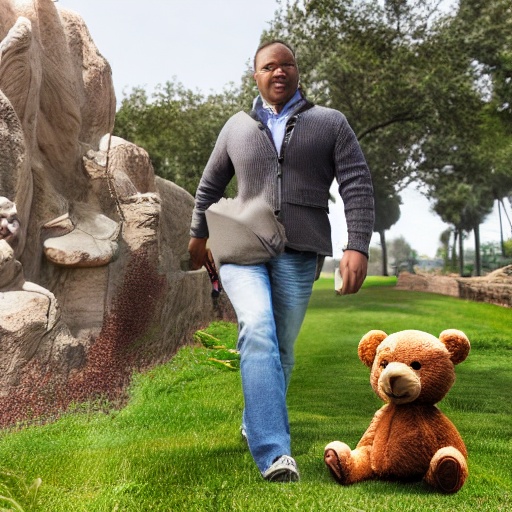}
  \includegraphics[width=0.18\linewidth]{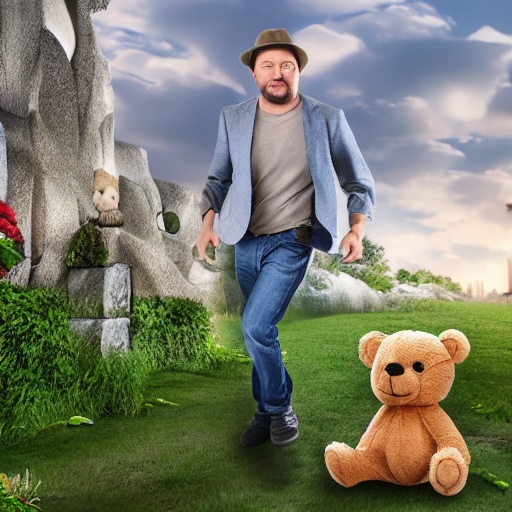}
  \includegraphics[width=0.18\linewidth]{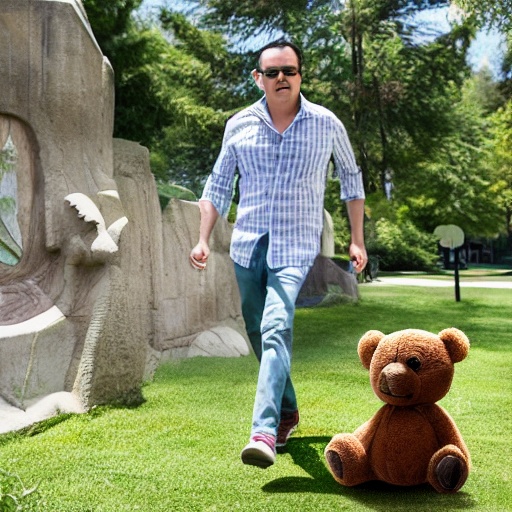}
  \includegraphics[width=0.18\linewidth]{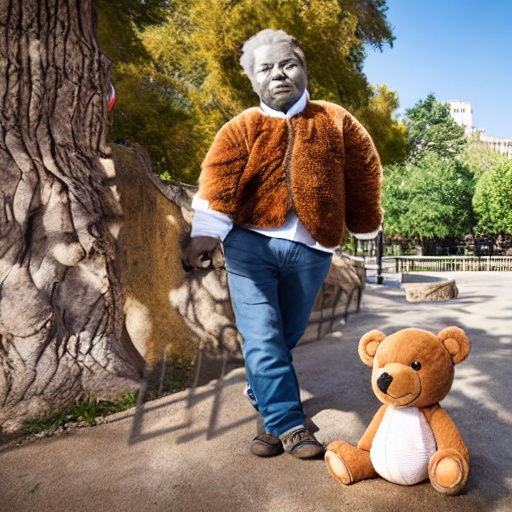}
  \includegraphics[width=0.18\linewidth]{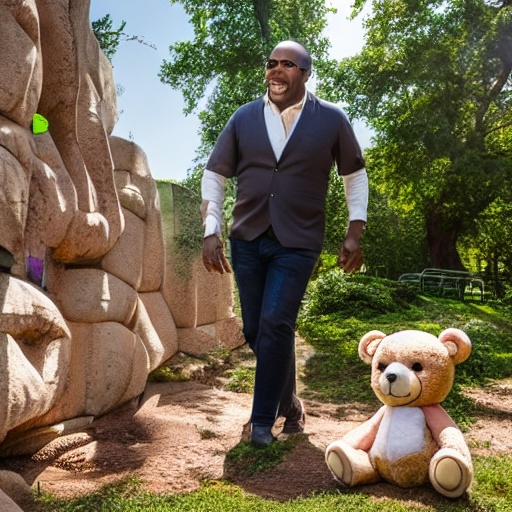}
\tabularnewline
  \includegraphics[width=0.18\linewidth]{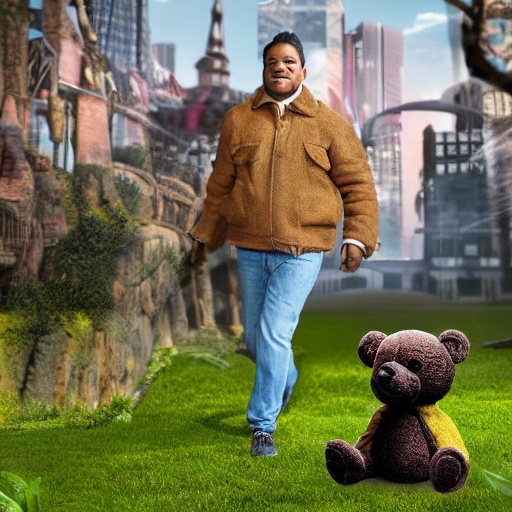}
  \includegraphics[width=0.18\linewidth]{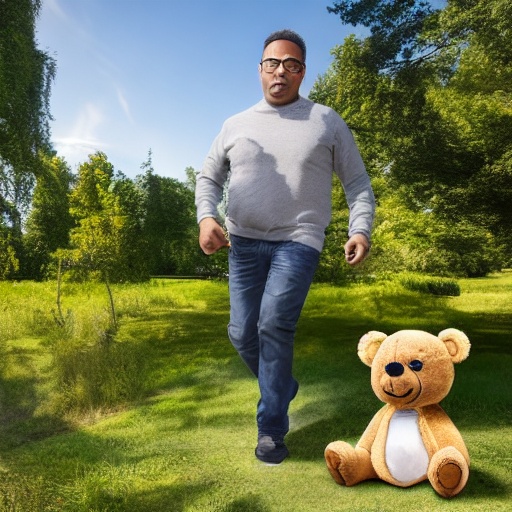}
   \includegraphics[width=0.18\linewidth]{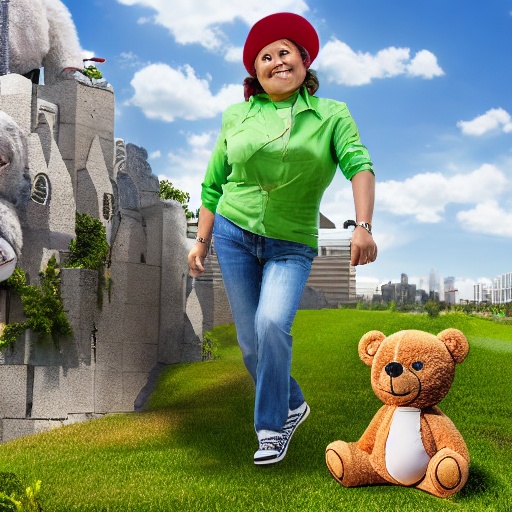}
  \includegraphics[width=0.18\linewidth]{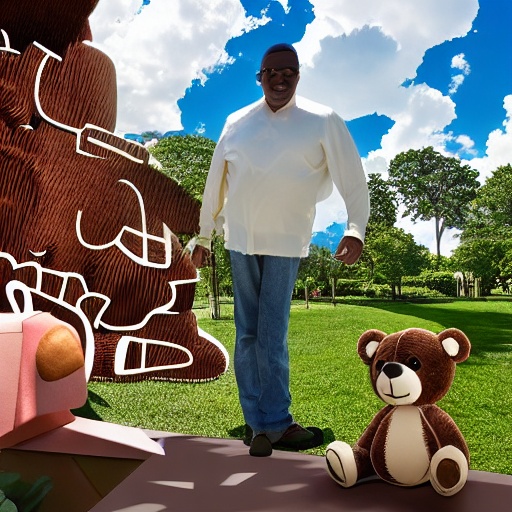}
  \includegraphics[width=0.18\linewidth]{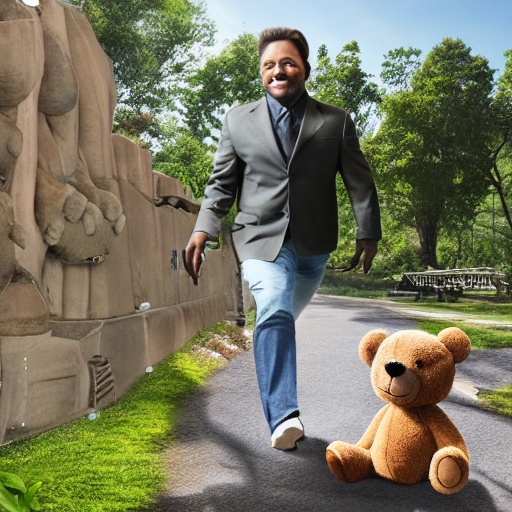}
  \tabularnewline
\vspace{2mm}
\vspace{-2\baselineskip}
\end{tabular}}
\vspace{-0.8cm}
\hspace{20pt}\captionof{figure}{\textbf{Non cherry picked examples for multimodal generation with 3 modalities. $\{HED,Depth,Pose\}$} Text Prompt is "Background: A park, Foreground: a teddy bear near a person"}
\label{fig:supp5}
\vspace{-2mm}
\end{figure*}%

  \begin{figure*}[t!]
    \centering
        \begin{subfigure}[t]{0.24\linewidth}
      \captionsetup{justification=centering, labelformat=empty, font=scriptsize}
      \includegraphics[width=1\linewidth]{intro/park_seg4.jpg}
      \caption{Seg Map}
    \end{subfigure}
    \begin{subfigure}[t]{0.24\linewidth}
      \captionsetup{justification=centering, labelformat=empty, font=scriptsize}
      \includegraphics[width=1\linewidth]{intro/corgi_hed.jpg}
      \caption{HED Map}
    \end{subfigure}
    \begin{subfigure}[t]{0.24\linewidth}
      \captionsetup{justification=centering, labelformat=empty, font=scriptsize}
      \includegraphics[width=1\linewidth]{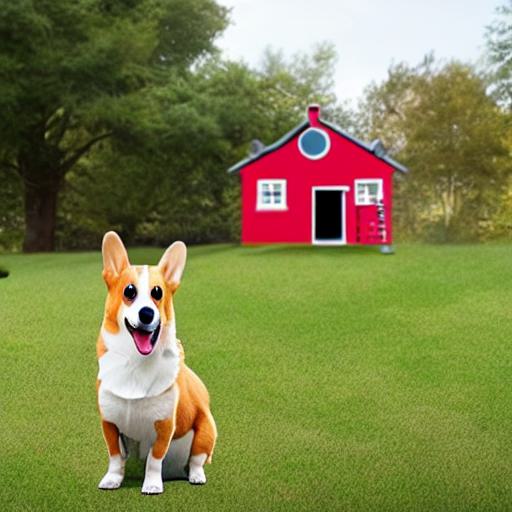}
      \caption{Variant Algorithm}
    \end{subfigure}
       \begin{subfigure}[t]{0.24\linewidth}
      \captionsetup{justification=centering, labelformat=empty, font=scriptsize}
      \includegraphics[width=1\linewidth]{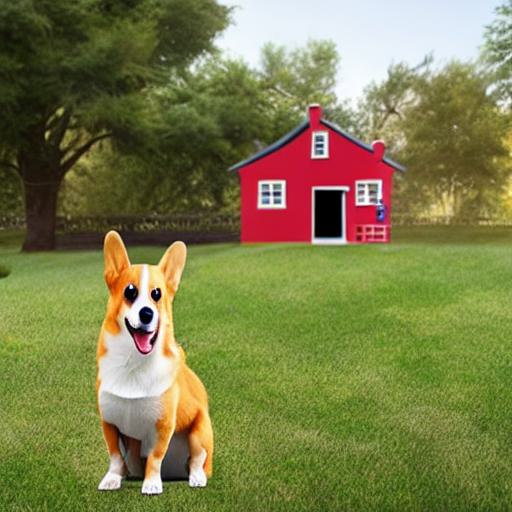}
      \caption{Original}
    \end{subfigure}
    \tabularnewline
        \begin{subfigure}[t]{0.24\linewidth}
      \captionsetup{justification=centering, labelformat=empty, font=scriptsize}
      \includegraphics[width=1\linewidth]{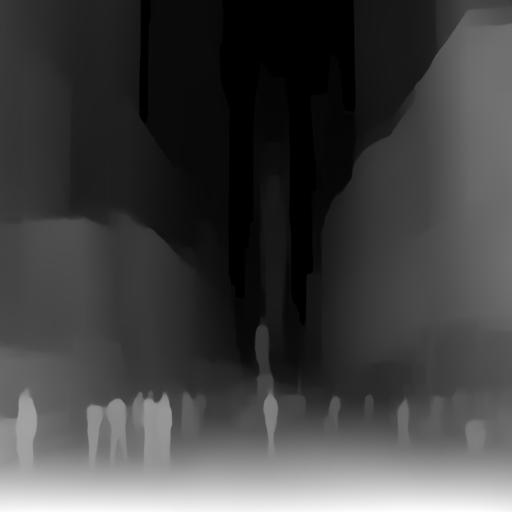}
      \caption{Depth Map}
    \end{subfigure}
    \begin{subfigure}[t]{0.24\linewidth}
      \captionsetup{justification=centering, labelformat=empty, font=scriptsize}
      \includegraphics[width=1\linewidth]{intro/luffy_pose.jpg}
      \caption{Pose Map}
    \end{subfigure}
    \begin{subfigure}[t]{0.24\linewidth}
      \captionsetup{justification=centering, labelformat=empty, font=scriptsize}
      \includegraphics[width=1\linewidth]{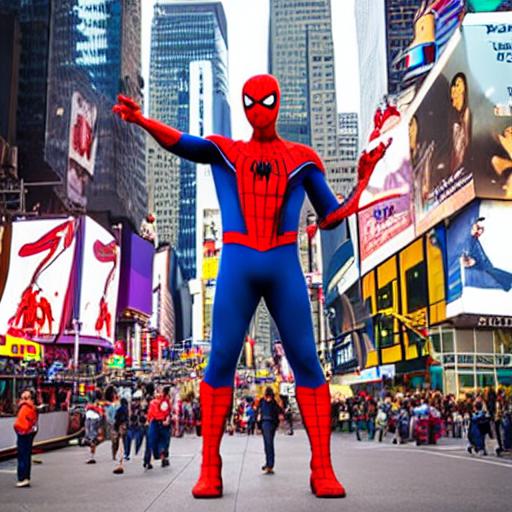}
      \caption{Variant Algorithm}
    \end{subfigure}
       \begin{subfigure}[t]{0.24\linewidth}
      \captionsetup{justification=centering, labelformat=empty, font=scriptsize}
      \includegraphics[width=1\linewidth]{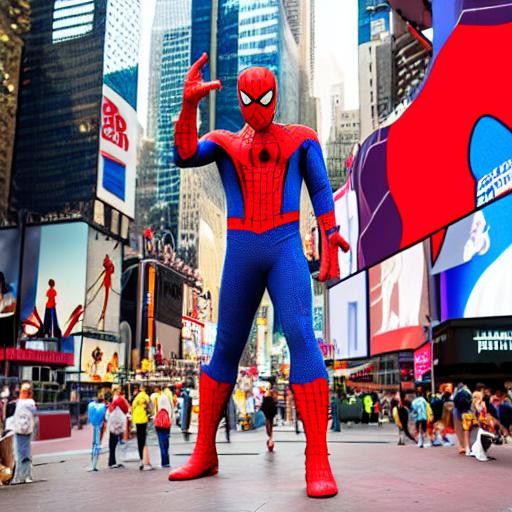}
      \caption{Original}
    \end{subfigure}
    \tabularnewline
        \begin{subfigure}[t]{0.24\linewidth}
      \captionsetup{justification=centering, labelformat=empty, font=scriptsize}
      \includegraphics[width=1\linewidth]{intro/hero_hed.jpg}
      \caption{HED Map}
    \end{subfigure}
    \begin{subfigure}[t]{0.24\linewidth}
      \captionsetup{justification=centering, labelformat=empty, font=scriptsize}
      \includegraphics[width=1\linewidth]{intro/alley_depth.jpg}
      \caption{Depth Map}
    \end{subfigure}
    \begin{subfigure}[t]{0.24\linewidth}
      \captionsetup{justification=centering, labelformat=empty, font=scriptsize}
      \includegraphics[width=1\linewidth]{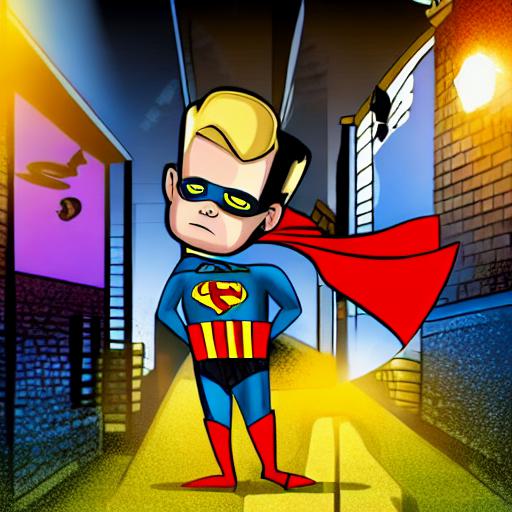}
      \caption{Variant Algorithm}
    \end{subfigure}
       \begin{subfigure}[t]{0.24\linewidth}
      \captionsetup{justification=centering, labelformat=empty, font=scriptsize}
      \includegraphics[width=1\linewidth]{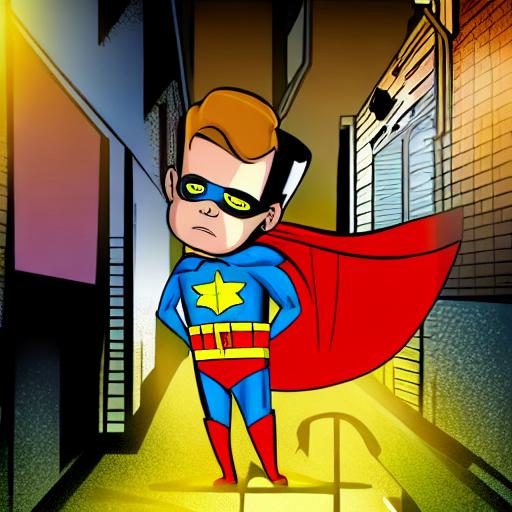}
      \caption{Original}
    \end{subfigure}
    \tabularnewline
        \begin{subfigure}[t]{0.24\linewidth}
      \captionsetup{justification=centering, labelformat=empty, font=scriptsize}
      \includegraphics[width=1\linewidth]{intro/bag.jpg}
      \caption{Seg Map}
    \end{subfigure}
    \begin{subfigure}[t]{0.24\linewidth}
      \captionsetup{justification=centering, labelformat=empty, font=scriptsize}
      \includegraphics[width=1\linewidth]{intro/posecanny.jpg}
      \caption{HED Map}
    \end{subfigure}
    \begin{subfigure}[t]{0.24\linewidth}
      \captionsetup{justification=centering, labelformat=empty, font=scriptsize}
      \includegraphics[width=1\linewidth]{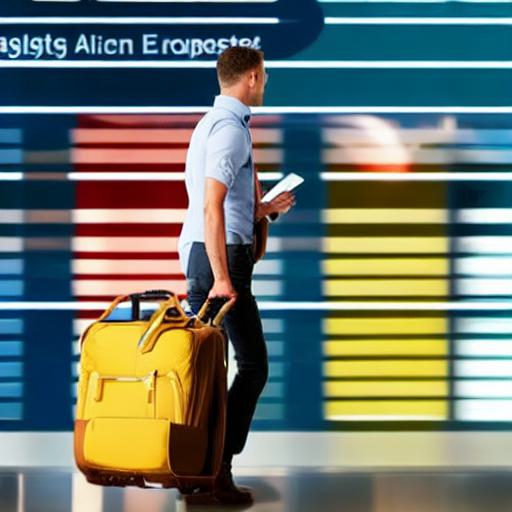}
      \caption{Variant Algorithm}
    \end{subfigure}
       \begin{subfigure}[t]{0.24\linewidth}
      \captionsetup{justification=centering, labelformat=empty, font=scriptsize}
      \includegraphics[width=1\linewidth]{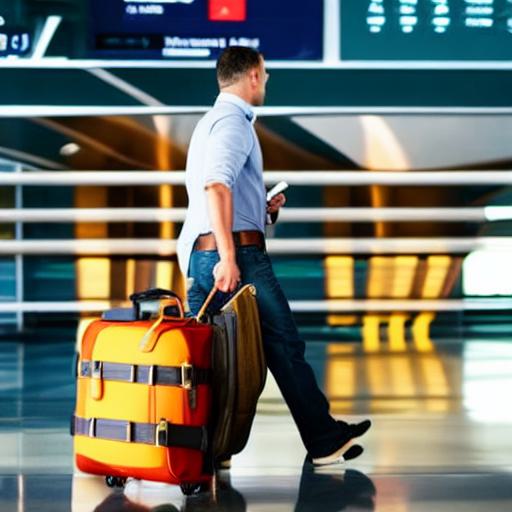}
      \caption{Original}
    \end{subfigure}
    \vspace{-3mm}    \caption{Ablation study corresponding to Variant algorithm vs the Original algorithm for different modalities}
    \label{fig:multiablation}
    \vspace{-5mm}
  \end{figure*}


\end{document}